\documentclass[11pt]{article}

\PassOptionsToPackage{hyphens}{url}
\usepackage[final]{acl}

\usepackage{times}
\usepackage{latexsym}

\usepackage[T1]{fontenc}

\usepackage[utf8]{inputenc}

\usepackage{microtype}

\usepackage{inconsolata}

\usepackage{graphicx}

\usepackage{url}
\usepackage{booktabs}
\usepackage{amsfonts}
\usepackage{amsmath}
\usepackage{nicefrac}
\usepackage{xcolor}
\usepackage[most]{tcolorbox}

\title{Can LLMs Take the Pulse of the Economy? A Real-Time Evaluation of LLM Nowcasts on Macroeconomic Indicators}

\author{
  \textbf{Xinyue Zhao\textsuperscript{1}}\thanks{\ Equal contribution.},
  \textbf{Ruiyi Zhang\textsuperscript{2}}\footnotemark[1],
  \textbf{Liqin Ye\textsuperscript{1}},
  \textbf{Rui Cao\textsuperscript{1}},
  \textbf{Pengtao Xie\textsuperscript{2}},
  \textbf{Sudheer Chava\textsuperscript{1}}\thanks{\ Corresponding author.}
\\
\\
  \textsuperscript{1}Georgia Institute of Technology,
  \textsuperscript{2}University of California San Diego
\\
  \small{\texttt{\{xinyue.zhao, sudheer.chava\}@scheller.gatech.edu}}
}

\begin{document}

\maketitle

\begin{abstract}
Nowcasting headline macroeconomic indicators, i.e., estimating an indicator's value for the current reference period before its official release, is critical for monetary policy and financial markets, and central banks devote dedicated teams of expert economists to producing such estimates. Large language model (LLM) agents are a promising candidate for this task, combining broad world knowledge with real-time web search and supporting queries at higher frequency than institutional nowcasts. Evaluating their nowcasting capability is, however, challenging: headline indicators such as GDP and CPI are widely reported and likely memorized during pretraining, so any evaluation on historical releases is vulnerable to data contamination. To address this, we introduce LiveMacroEval, a live, contamination-resistant benchmark in which LLM agents produce hourly nowcasts for sixteen major U.S.\ macroeconomic indicators over a pre-release window closing at each official release. Nowcast quality is assessed through a LiveMacro Score against announcement-window equity returns and a LiveBetting Score from simulated Polymarket-style trading, with Federal Reserve regional-bank nowcasts, the Bloomberg ECOS professional consensus, and an auto-ARIMA baseline as comparators. Over six months with four state-of-the-art LLM agents configured with web search, aggregate nowcast accuracy is broadly comparable to the institutional and professional benchmarks, with performance varying widely across individual indicators. This highlights LLM agents' potential as real-time estimators of macroeconomic conditions.
\end{abstract}

\section{Introduction}

Headline macroeconomic indicators such as GDP growth, inflation, and retail activity guide monetary policy and asset pricing, yet government statistical agencies publish them only weeks to months after the reference period~\citep{bok2018nowcasting}. \emph{Nowcasting}, the task of estimating an indicator's value for the current or just-completed period before its scheduled official release~\citep{giannone2008nowcasting,banbura2013nowcasting}, is therefore a central concern of empirical macroeconomics, with major central banks devoting dedicated teams of expert economists to producing such estimates~\citep{atlfed_gdpnow,nyfed_nowcast,angelini2011euro}. Large language model (LLM) agents are a promising candidate technology for this task: nowcasting requires continuous access to up-to-date information (structured statistical releases, central-bank communications, the financial-news flow) and the ability to reason over that information to revise a forecast, and modern LLM agents combine tool use such as web search with general-purpose reasoning over the retrieved content~\citep{yao2023react,schick2023toolformer}. 
They can also be queried at higher frequency than institutional nowcasts, which update only weekly or upon major source-data releases.

Evaluating LLM nowcasting capability is, however, challenging. Headline indicators such as GDP and CPI are among the most widely reported numbers and their historical values are likely memorized during pretraining, so directly evaluating LLM agents on historical releases is exposed to training-data contamination~\citep{golchin2024timetravel,gao2025rewindtime}. A reliable assessment requires evaluation in a strictly live setting, where each target indicator has not yet been officially released at prediction time. Recent live benchmarks for general-domain prediction~\citep{zeng2026futurex,karger2025forecastbench,white2025livebench,halawi2024approaching} establish this pretraining-contamination-free methodology, but do not target macroeconomic indicator nowcasting.

\begin{figure*}[t]
    \centering
    \includegraphics[width=\linewidth]{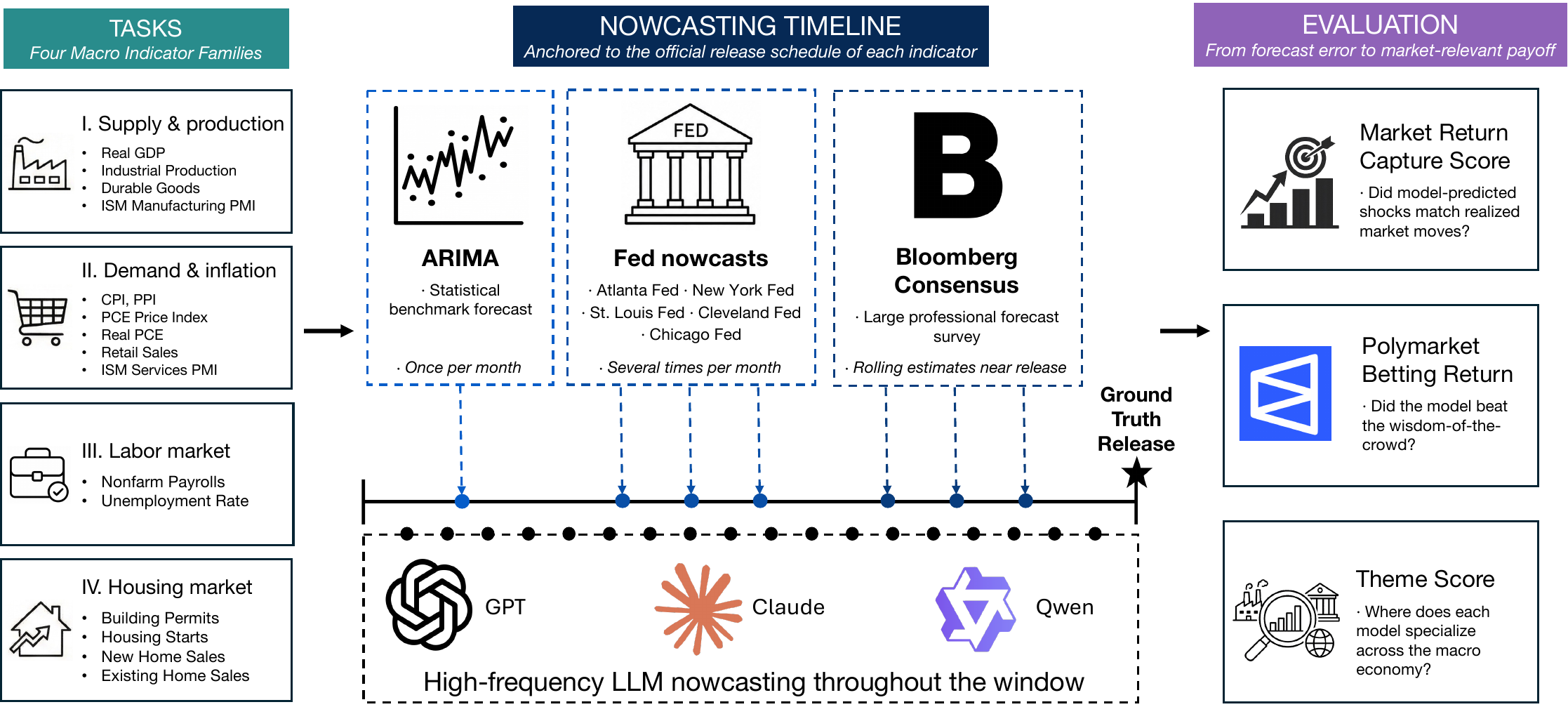}
    \caption{Overview of the LiveMacroEval pipeline. For each of sixteen U.S.\ headline macroeconomic indicators, LLM agents produce high-frequency nowcasts in the live pre-release window using web search. Each nowcast is then scored against the official release under two evaluation metrics: the LiveMacro Score and the LiveBetting Score. For reference, the same metrics are computed for an econometric baseline and for the top expert nowcasts produced by Federal Reserve regional banks and the Bloomberg ECOS professional consensus.}
    \label{fig:overview}
\end{figure*}

To address this gap, we first introduce LiveMacroEval\footnote{Code and data: \url{https://github.com/LiveMacroEval/LiveMacroEval}. Website: \url{https://livemacroeval.github.io/}.}, a live, contamination-resistant benchmark for LLM agents on macroeconomic nowcasting, illustrated in Figure~\ref{fig:overview}. LiveMacroEval covers sixteen major U.S.\ indicators spanning four thematic blocks (supply and production, demand and inflation, the labor market, and housing) that jointly summarize the macroeconomy. For each indicator and reference period, LLM agents produce hourly nowcasts over a weeks-to-months window that closes at the scheduled official release. We also propose two novel metrics for nowcast quality: a LiveMacro Score based on the aggregate explanatory effect of the LLM agent's predicted macro-indicator values on the equity market, motivated by the econ-finance result that macro-indicator surprises causally move equity prices~\citep{andersen2003micro,boyd2005stock}, and a LiveBetting Score that reports the cumulative return from hourly trading of Polymarket-style prediction-market contracts. LiveMacroEval provides two types of baselines for comparison: economist nowcasts (five Federal Reserve regional-bank nowcasts from Atlanta, New York, St.~Louis, Cleveland, and Chicago, together with the Bloomberg ECOS professional consensus survey) and an econometric time-series baseline (univariate auto-ARIMA).

We conduct a six-month live evaluation of four web-search-enabled LLM agents (GPT-5, Claude-sonnet-4.5, Qwen3-235B, Qwen3-80B) using LiveMacroEval. We observe that the top LLM performer (GPT-5) ties with the state-of-the-art economist consensus (Bloomberg ECOS) on the aggregate LiveMacro Score, and that the theme-level decomposition reveals notable heterogeneity. On the LiveBetting Score, several LLM agents are competitive with the corresponding institutional Federal Reserve nowcast for GDP and the unemployment rate. The LLM agents' nowcast revisions further align with economically meaningful information events within the evaluation window. Together, our results indicate the strong potential of LLM agents for macroeconomic nowcasting. More broadly, LiveMacroEval offers a concrete template for designing live, contamination-resistant evaluation protocols for LLM agents on economically impactful real-world tasks.

\section{Related Work}\label{sec:related-work}

Our work is related to four lines of work. (1) Nowcasting, which is the standard task in empirical macroeconomics for real-time estimation of headline indicators before their official release \citep{giannone2008nowcasting,kuzin2011midas,bok2018nowcasting,babii2022machine}. (2) LLMs for economic and financial forecasting, which study how prompted frontier models can produce point estimates of macroeconomic and financial variables \citep{carriero2024macroeconomic,fariaecastro2025artificial,iadisernia2025prompting,ganum2026how,lopezlira2023can}. (3) LLM memorization, which means that widely reported facts and benchmark instances seen during pretraining can be reproduced at inference time \citep{golchin2024timetravel,gao2025rewindtime}. (4) Live future-prediction benchmarks, which evaluate LLMs on events unresolved at query time to obtain a pretraining-contamination-free assessment \citep{white2025livebench,halawi2024approaching,karger2025forecastbench,zeng2026futurex}. An extended discussion of all four threads is provided in Appendix~\ref{app:related-work-extended}.

\section{LiveMacroEval}

\subsection{Overview}

LiveMacroEval is a live, contamination-resistant framework for evaluating LLM agents on macroeconomic nowcasting: real-time estimation of a headline indicator before its scheduled official release \citep{giannone2008nowcasting,banbura2013nowcasting}. As illustrated in Figure~\ref{fig:overview}, the task targets sixteen U.S.\ indicators across four thematic blocks (Section~\ref{sec:tasks-themes}). For each prediction window (weeks to months, closing at the official release), LLM agents nowcast continuously (Section~\ref{sec:agent-nowcasts}) while LiveMacroEval collects Bloomberg ECOS and Federal Reserve nowcasts, auto-computes an ARIMA baseline (Section~\ref{sec:comparators}), and ingests the released value as ground truth. LiveMacroEval further introduces two novel metrics, the LiveMacro Score and the LiveBetting Score, which assess how each model's nowcasts translate into equity returns and prediction-market trading returns (Section~\ref{sec:metric}). Together, these components address the paper's central question: can LLM agents outperform traditional econometric models and top human experts at real-time macroeconomic nowcasting?

\subsection{Definition of Nowcast}\label{sec:prelim}

A nowcast is a running estimate of a headline macroeconomic indicator for the current or just-completed reference period, revised as new information arrives and finalized at the agency's official release. Nowcasts are a primary input to monetary and fiscal policymaking under multi-week release lags and repeated revisions \citep{bok2018nowcasting}, and major central banks therefore produce dedicated nowcast products \citep{nyfed_nowcast,atlfed_gdpnow,angelini2011euro}. Methodologically, nowcasting sits at the frontier of empirical macroeconomic forecasting, with mixed-frequency frameworks (dynamic factor models, MIDAS) and machine-learning extensions exploiting the asynchronous data calendar \citep{marcellino2010factormidas,babii2022machine,huber2023nowcasting}. We provide an extended discussion in Appendix~\ref{app:nowcasting-background}.

\subsection{Macroeconomic Indicators}\label{sec:tasks-themes}

LiveMacroEval covers sixteen major U.S.\ macroeconomic indicators organized into four thematic blocks, following the standard empirical macroeconomic forecasting decomposition \citep{stockwatson2002diffusion}. These indicators are tracked by professional economists and major institutions, including the Bloomberg ECOS consensus survey \citep{bloomberg2024ecos} and the FRED-MD and FRED-QD research databases \citep{mccracken2016fredmd,mccracken2021fredqd}. We retrieve released values from issuing agencies and release timestamps from the Bloomberg Economic Calendar \citep{bloomberg_econ_cal}.

\paragraph{Supply and Production.}
The first block contains four indicators measuring what the real economy produces. Real Gross Domestic Product is the broadest quarterly measure of U.S.\ output. Industrial Production Index tracks physical output in manufacturing and utilities. Manufacturers' New Orders: Durable Goods is a leading indicator of future production. ISM Manufacturing PMI is the most watched manufacturer survey and a forward-looking cycle signal \citep{giannone2008nowcasting}.

\paragraph{Demand and Inflation.}
The second block contains six indicators describing household spending at the prices households pay \citep{faustwright2013forecasting,andersen2003micro}. The Consumer Price Index is the most visible measure of consumer prices. The PCE Price Index is the formal benchmark for the Federal Reserve's 2\% inflation target. The Producer Price Index captures upstream cost pressures. Real Personal Consumption Expenditures is the inflation-adjusted measure of household spending. Retail Sales, Advance Monthly is the earliest monthly read on nominal household spending at retail and food-service. ISM Services PMI is the most watched survey of non-manufacturing firms.

\paragraph{Labor Market.}
The third block contains two employment indicators, a primary input to Federal Reserve policy decisions \citep{andersen2003micro}. Nonfarm Payrolls reports the monthly net change in paid employment, typically the most closely watched U.S.\ data release. The Unemployment Rate measures the share of the labor force without a job and is a core input to Taylor-rule-style policy analyses.

\paragraph{Housing.}
The final block contains four indicators describing the housing market, the largest source of household debt and the most informative segment for monetary-policy transmission \citep{leamer2007housing}. Building Permits is a forward-looking indicator of the residential pipeline. Housing Starts is a coincident indicator of residential investment. New Home Sales is a timely read on primary-market demand. Existing Home Sales accounts for the bulk of U.S.\ housing transactions.

Extended descriptions of each indicator are in Appendix~\ref{app:indicator-details}.

\subsection{LLM Agent Nowcasts}\label{sec:agent-nowcasts}

Two features make LLM agents natural candidates for nowcasting: the task requires synthesizing heterogeneous, mixed-frequency information (statistical releases, market signals, and unstructured text such as central-bank communications and the news flow), and it rewards frequent updating, both of which align with what tool-using LLM agents are designed to do. In LiveMacroEval, each agent is prompted with a strict definition of the target task and variable for the given indicator, including a fixed title, unit, reporting transformation, required output format, choice of summary measure (e.g., headline versus core), and choice of growth transformation (e.g., year-over-year versus month-over-month). The full prompt is shown in Appendix~\ref{app:prompt}. Each LLM agent is allowed to issue multiple web-search calls during a single nowcast call, and the agent itself decides when and what to search.

The core of the evaluation is the agent's prediction window: for each target indicator, it opens in the final week of the reference month and closes at the scheduled official release, a span of weeks to months that varies across indicators (e.g., the agent begins forecasting March CPI at the end of March and continues at hourly frequency through the release on April~10). All nowcasts within this window are generated in a strictly real-time workflow with no access to post-release information, so the target value cannot have entered the model's training corpus or its retrieved web content, making the evaluation pretraining-contamination-free by construction. As a sanity check, letting the agent continue nowcasting for five days after the release shows the cluster snapping onto the released value (Appendix~\ref{app:postrelease}), confirming the agent can retrieve official numbers once they exist and underscoring the necessity of evaluating strictly before release.

\subsection{Comparative Method Nowcasts}\label{sec:comparators}

To benchmark the LLMs, LiveMacroEval collects two types of comparative nowcasts on the same prediction window: nowcasts from human economists and an econometric time-series baseline.

\paragraph{Economist nowcasts.} LiveMacroEval contains two types of economist nowcasts: U.S.\ Federal Reserve economist nowcasts on GDP, CPI, and the unemployment rate, and Bloomberg ECOS economist nowcasts on all sixteen target indicators. On the Federal Reserve side, the Atlanta Fed GDPNow \citep{atlfed_gdpnow}, the New York Fed Staff Nowcast \citep{nyfed_nowcast}, and the St.~Louis Fed Real GDP Nowcast \citep{stlfed_realgdp_nowcast} all target real GDP. The Cleveland Fed Inflation Nowcasting targets CPI \citep{clevelandfed_inflation_nowcast}. The Chicago Fed Unemployment Rate Nowcast (CHURN) targets the unemployment rate \citep{chicagofed_churn}. 

On the Bloomberg side, the ECOS survey \citep{bloomberg2024ecos}, distributed via the paid Bloomberg Terminal, aggregates estimates from a panel of professional economists in the final week before each release. We take the cross-sectional median as the consensus, matching Bloomberg's convention. The Bloomberg consensus is the canonical benchmark for real-time macroeconomic expectations in the asset-pricing and monetary-policy literature \citep{coibion2015information,faustwright2013forecasting} and serves both as a direct comparison nowcast and a reference input for the market-based metrics in the next subsection. Appendix~\ref{app:contamination-check} addresses the concern that LLM agents may copy the Bloomberg consensus or the Federal Reserve nowcasts, reporting three direct tests that find no evidence of copying. Appendix~\ref{app:reasoning-traces} inspects the rationales from GPT-5's reasoning output. The two benchmarks differ in the information they can draw on. Bloomberg's contributors are traders, portfolio managers, think tanks, and academics, so their edge is whatever information their institutions can bring to bear, while four of the five Federal Reserve nowcasts run on essentially public government data, with only the Chicago Fed's CHURN adding private feeds. Appendix~\ref{app:benchmark-information} documents the inputs behind each benchmark.

\paragraph{Time-series baseline.} For each indicator, we fit a univariate auto-ARIMA baseline at the start of each forecasting window, a standard fast benchmark for univariate macroeconomic forecasting \citep{hyndman2008automatic}. Full details are provided in Appendix~\ref{app:arima-baseline}.

\subsection{Evaluation Metrics}\label{sec:metric}

Standard prediction-error metrics (MSE, MAE) on the released level ignore the differing economic relevance of indicators, conflate prediction quality with each indicator's intrinsic scale, and collapse the full pre-release trajectory into a single horizon-specific number. We address these limitations with two metrics rooted in economics and finance theory: the LiveMacro Score and the LiveBetting Score. As a familiar point of reference, we report each indicator's real-time squared relative error against the official release in Appendix~\ref{app:nowcast-error}.

\paragraph{LiveMacro Score.} The motivation comes from financial-economics research: the surprise component of a macro release, defined as the deviation of the released value from the prevailing professional consensus, causally affects equity prices, whereas the released level itself is already impounded through the consensus \citep{andersen2003micro,boyd2005stock}. Concretely, consider a single release event for an indicator $i$. Let $X_i$ be its released value, $X^c_i$ the prevailing Bloomberg consensus, and $\sigma_i$ a historical scale for that indicator's surprises. The standardized surprise is $S_i = (X_i - X^c_i)/\sigma_i$. We select the log return on E-mini S\&P~500 futures over a tight $[-5, +30]$ minute window around the release as our measure of equity price change $r$, following the standard event-study convention in finance to isolate the causal effect of the macro surprise from confounding news \citep{andersen2007realtime,gurkaynak2020missing}. The window is short and its timing is known in advance, so little other major news is expected to arrive inside it and the return can be read as a response to the released surprise alone. This is the identifying assumption asset-pricing research uses to isolate the effect of a shock \citep{kuttner2001monetary,nakamura2018high,bauer2023alternative,andersen2003micro}. The causal relationship can then be captured by the linear regression
\begin{equation}
r = \beta_i S_i + u,
\end{equation}
where $\beta_i$ is the causal effect of a one-unit surprise in indicator $i$ on $r$. Following the high-frequency event-study convention \citep{gurkaynak2020missing}, the regression carries no constant: under the identifying assumption that the announcement-window return responds only to the released surprise, a zero surprise carries no expected return. We estimate $\beta_i$ once on a long historical sample and freeze it before any live outcome is scored. Appendix~\ref{app:capture-score-details} reports a robustness check that re-estimates $\{\beta_i\}$ with the COVID-19 window excluded and re-scores the identical releases. Because $\beta_i$ encodes each indicator's long-run causal weight on equity returns, downstream evaluation that builds on $\{\beta_i\}$ automatically rewards skill on high-impact releases (e.g., CPI, nonfarm payrolls, GDP) and discounts smaller-impact ones. This is what lets the score collapse sixteen per-indicator accuracies into one comparable number, so that which agent nowcasts the economy better overall has a well-defined answer. Weighting and aggregating macro surprises by their measured market impact is long-standing practice in economics \citep{balduzzi2001economic,faust2007highfrequency,scotti2016surprise}, and we follow it rather than weighting the sixteen indicators equally. Appendix~\ref{app:conventional-aggregators} shows why an equal-weight error sum is not ideal. The sum is dominated by a few events, and it weights the most market-moving series the same as the least.

Given the estimates $\{\beta_i\}$, we evaluate the quality of an agent's macro-indicator nowcasts by the combined return-prediction error from plugging them into the calibrated equity-return formula (the LiveMacro Score). For each release event of indicator $i$, we plug the agent's implied surprise $\hat S_i = (\hat X_i - X^c_i)/\sigma_i$, where $\hat X_i$ is the agent's latest pre-release nowcast, into the calibrated formula to obtain a model-implied equity return $\hat r_i = \beta_i \hat S_i$ for that event, and the corresponding realized return $r_i$. The more accurately the nowcasts track the released values, the more accurately $\hat r_i$ tracks $r_i$. The LiveMacro Score then aggregates over all indicators $i$ in the live evaluation sample, comparing $\hat r_i$ against $r_i$ event-by-event and using a consensus-only baseline $r^c_i$ as the reference. With $\mathrm{SS}_{\mathrm{model}} = \sum_i (r_i - \hat r_i)^2$ and $\mathrm{SS}_{\mathrm{cons}} = \sum_i (r_i - r^c_i)^2$, the score is
\begin{equation}
\mathrm{LiveMacro}(r, \hat r, r^c) = \frac{\mathrm{SS}_{\mathrm{cons}} - \mathrm{SS}_{\mathrm{model}}}{\mathrm{SS}_{\mathrm{cons}} + \mathrm{SS}_{\mathrm{model}}},
\end{equation}
a bounded variant of the out-of-sample $R^2$ \citep{campbell2008predicting}. Treating the Bloomberg consensus itself as a model, its implied surprise is identically zero, so the consensus prediction reduces to $r^c_i \equiv 0$ for every event. A score of $+1$ denotes nowcasts that perfectly anticipate announcement-window returns, $0$ matches the consensus baseline, and values approaching $-1$ indicate predictions arbitrarily worse than the consensus. Calculating the score within a single thematic block yields a theme-specific score. The general formulation, which handles simultaneous releases, and detailed data usage are provided in Appendix~\ref{app:capture-score-details}.

\paragraph{LiveBetting Score.} This metric reports the cumulative return from trading Polymarket-style prediction-market contracts on each model's nowcast. The motivation is twofold: prediction markets have been advanced in the recent macroeconomics literature as a high-frequency, continuously updated, and distributionally rich benchmark for real-time macroeconomic expectations \citep{diercks2026kalshi}, and the resulting return is a unit-free quantity that can be aggregated directly across heterogeneous macroeconomic variables without concerns about error-scale comparability. 

For each target indicator, the Polymarket macro market is a set of mutually exclusive numeric buckets, each trading as a binary contract that pays \$1 if the released value falls in that bucket. At every hour strictly before the release, we place a \$1 bet on the bucket containing the agent's most recent nowcast, buying $1/p$ shares at the prevailing market price $p$. Once the release determines the winning bucket, each bet either pays out at its inverse entry price or expires worthless. Here $p$ is the crowd's pool-implied probability for that bucket, so the bucket's liquidity is priced into the return by construction: when the crowd is already betting correctly the winning return is low, and when the crowd is wrong while the agent is right the return is high. The design therefore rewards the agent precisely for beating the crowd. We report the average net return across all hourly bets placed during the prediction window. Formal notation and the closed-form expression are reported in Appendix~\ref{app:polymarket-return-details}.

\section{Results}

\begin{figure}[t]
\centering
\includegraphics[width=\linewidth]{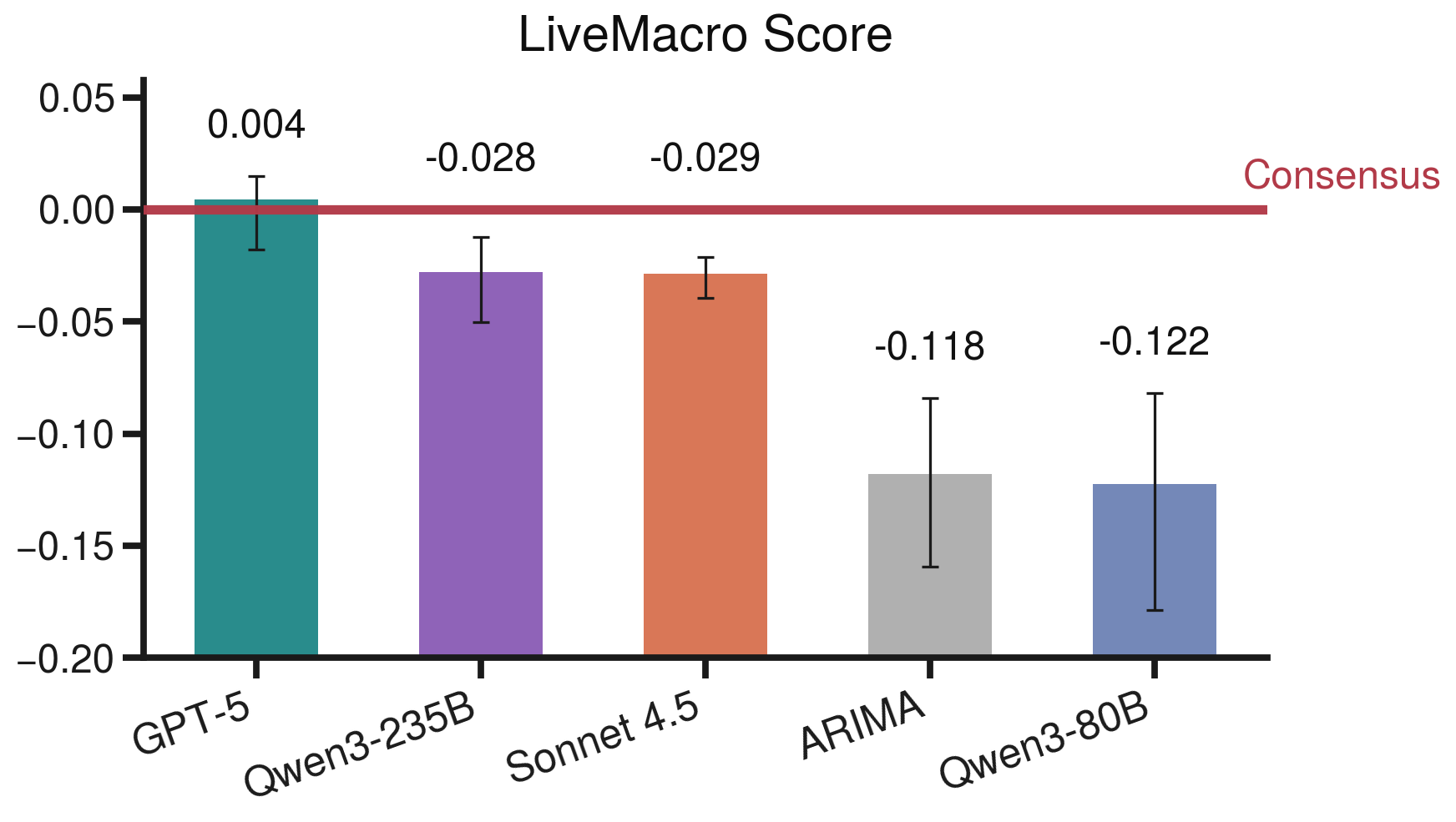}
\caption{LiveMacro Score by model. Six-month live evaluation, target periods November~2025 through March~2026. Positive values indicate the model exceeds the Bloomberg professional consensus on equity-market relevance. Error bars are 90\% parametric-bootstrap confidence intervals. Full construction and sampling-based intervals are in Appendix~\ref{app:capture-score-details}.}
\label{fig:bdrc}
\end{figure}

\subsection{Experimental Setup}

We report results over a six-month live evaluation with target periods from November~2025 through March~2026, corresponding to official release dates from December~1, 2025 through May~5, 2026, yielding hundreds of thousands of nowcasts across the sixteen indicators. The LLM agents evaluated are GPT-5 \citep{openai2025gpt5}, Claude-sonnet-4.5 \citep{anthropic2025sonnet45}, Qwen3-235B, and Qwen3-80B \citep{yang2025qwen3}. The LiveMacroEval framework itself is not specific to this evaluation set or to this set of agents.

\begin{figure*}[t]
\centering
\includegraphics[width=0.8\linewidth]{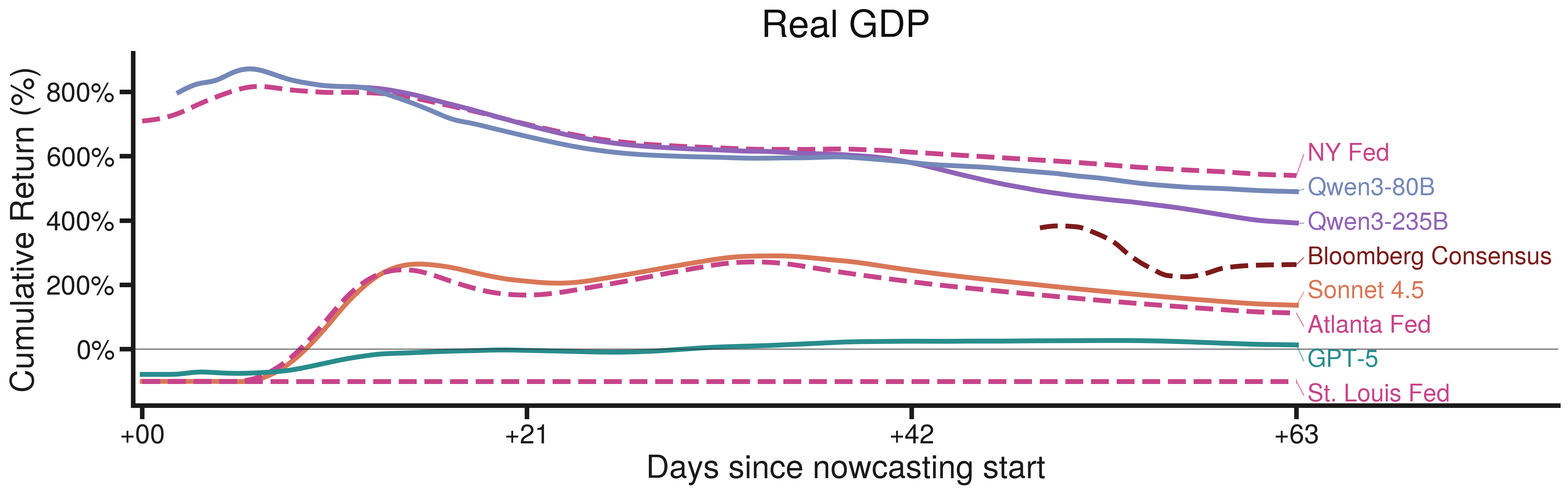}\\[4pt]
\includegraphics[width=0.8\linewidth]{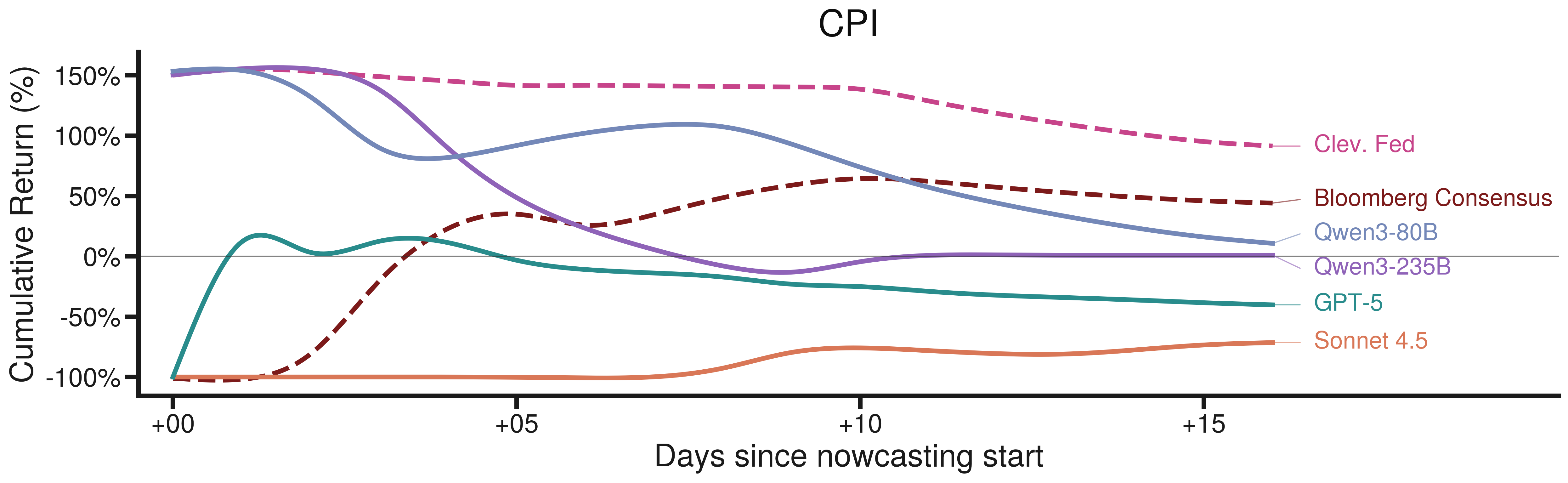}\\[4pt]
\includegraphics[width=0.8\linewidth]{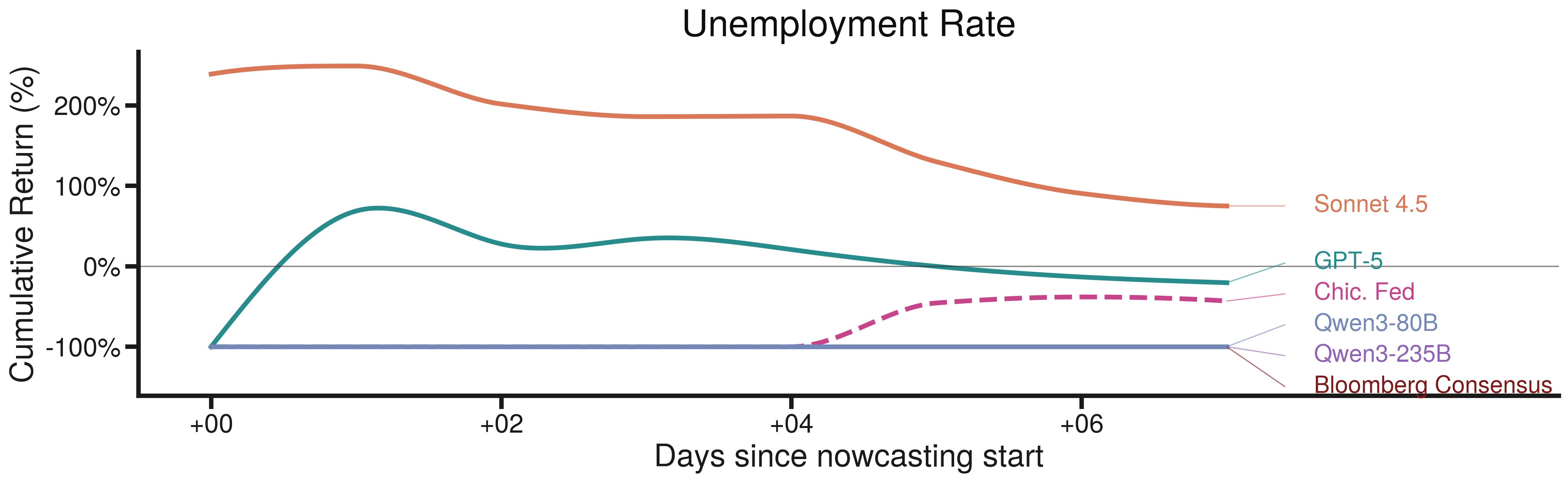}
\caption{Cumulative LiveBetting returns by indicator. Real GDP (top), CPI (middle), and unemployment rate (bottom), overlaying the four LLM agents, the Bloomberg consensus, and the available Federal Reserve nowcasts.}
\label{fig:polymarket-cum-claude}
\end{figure*}

\subsection{LiveMacro Score}\label{sec:results-livemacro}

Figure~\ref{fig:bdrc} reports the LiveMacro Score for each model, aggregating the explanatory effect of each model's implied surprises across all sixteen indicators jointly, with each indicator weighted by its historical causal effect on equity returns. GPT-5 leads the panel at $+0.004$, statistically indistinguishable from the strong Bloomberg consensus and thus a tie. The remaining LLM agents sit modestly below the consensus but still above the auto-ARIMA baseline, except Qwen3-80B, which is indistinguishable from it. Tying the strong Bloomberg consensus on scheduled releases is itself a non-trivial outcome, as the consensus is the strongest publicly available pre-release expectation for U.S.\ headline macro indicators and is well known to be extraordinarily difficult to match \citep{faustwright2013forecasting}. This provides initial, if modest, evidence that a contemporary LLM agent can extract macro-relevant information at a level broadly comparable to the professional consensus, suggesting LLM agents are a candidate technology worth studying for real-time macroeconomic nowcasting. This comparability is an aggregate statement. Decomposing the score by indicator (Appendix~\ref{app:livemacro-results-details}) shows that the wins and shortfalls are concentrated rather than spread evenly, with the average agent adding value on the activity and housing-supply series while CPI, retail sales, and the PCE price index account for $87\%$ of its shortfall. Section~\ref{sec:results-livemacro-theme} gives the theme-level view. Per-model breakdowns and confidence interval calculations are provided in Appendix~\ref{app:livemacro-results-details}. A directional variant of the score isolating sign-prediction skill is in Appendix~\ref{app:livemacro-directional}.

\begin{figure*}[t]
\centering
\includegraphics[width=0.5\linewidth]{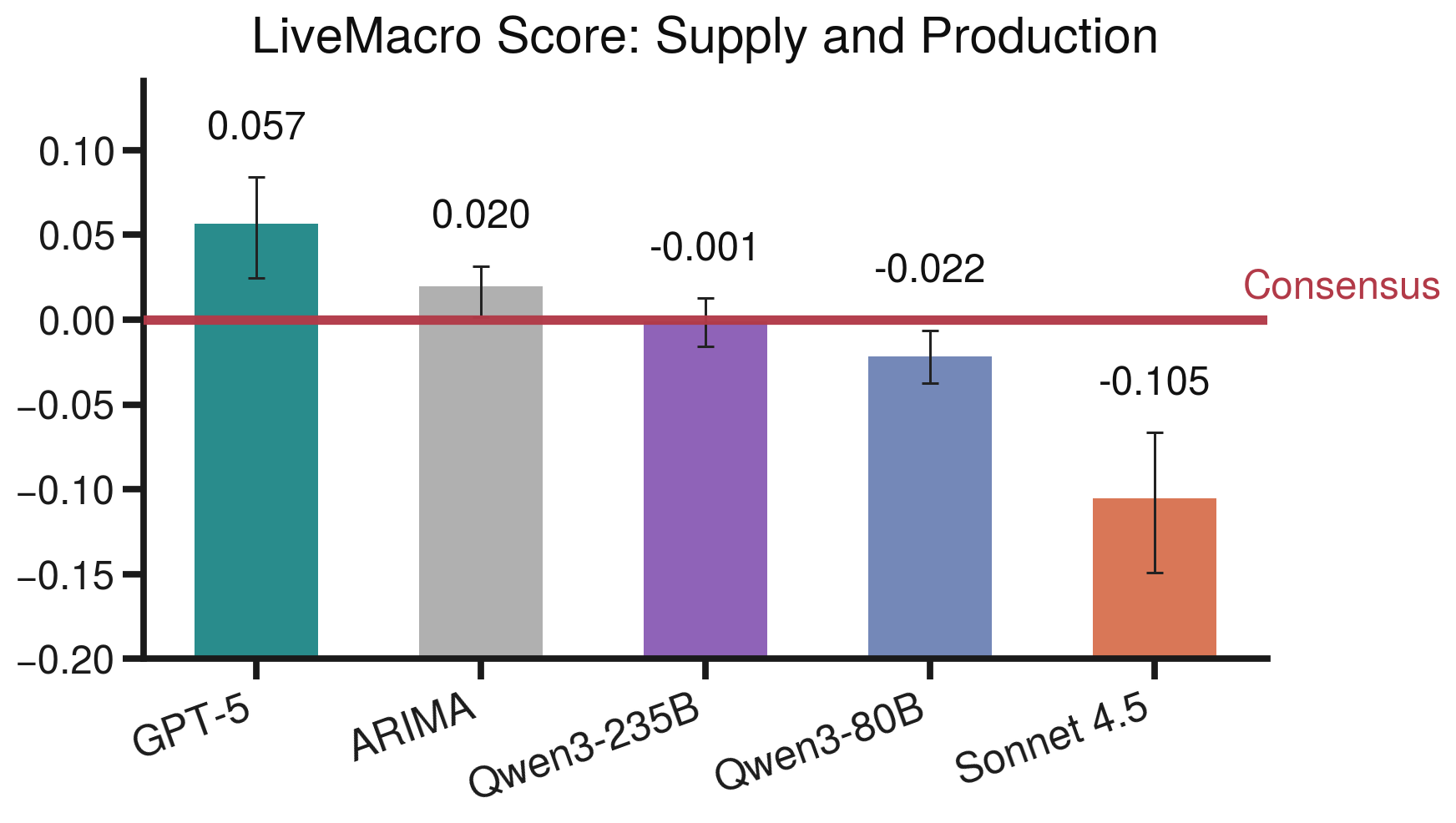}\hfill
\includegraphics[width=0.5\linewidth]{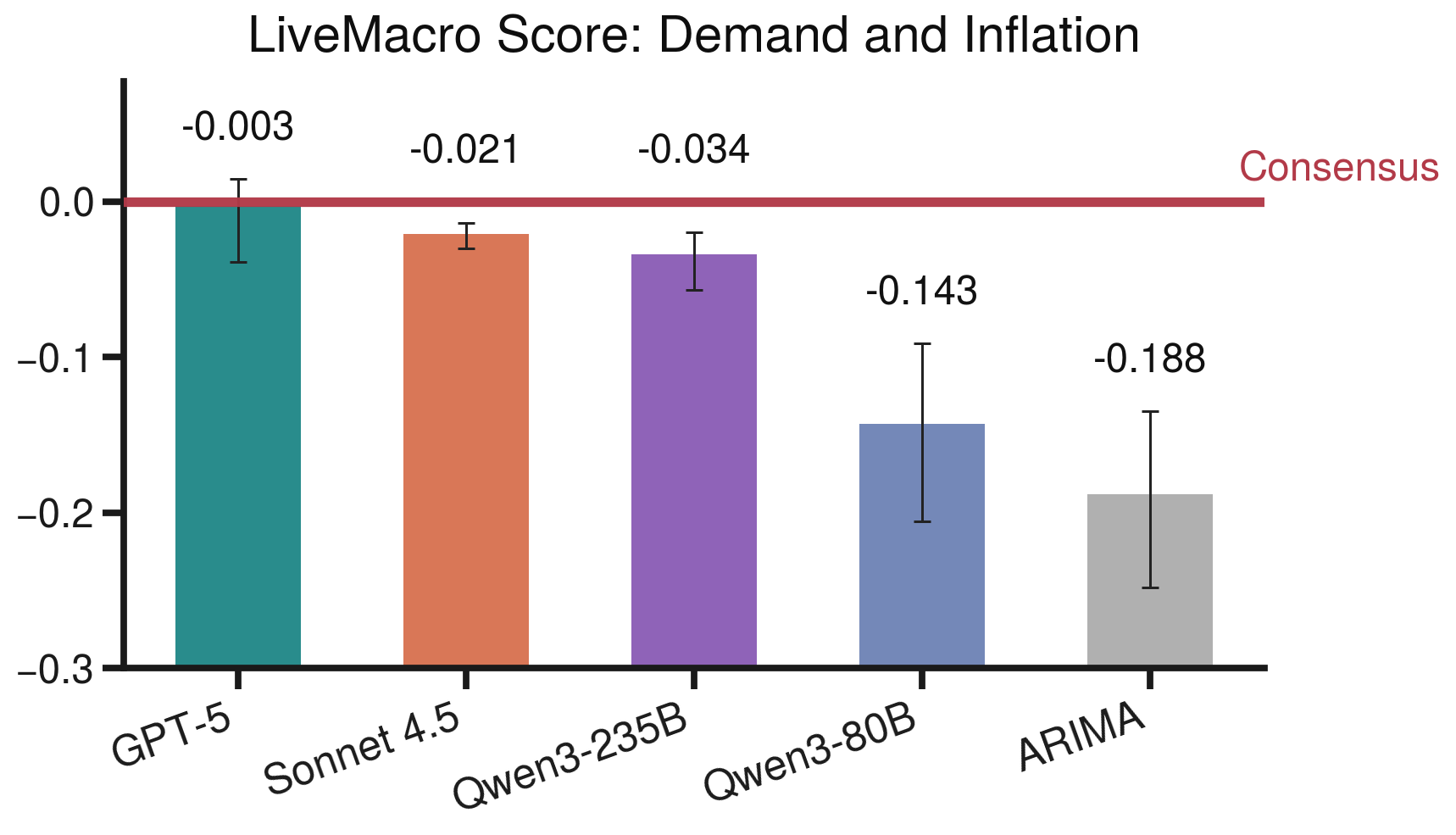}\\[4pt]
\includegraphics[width=0.5\linewidth]{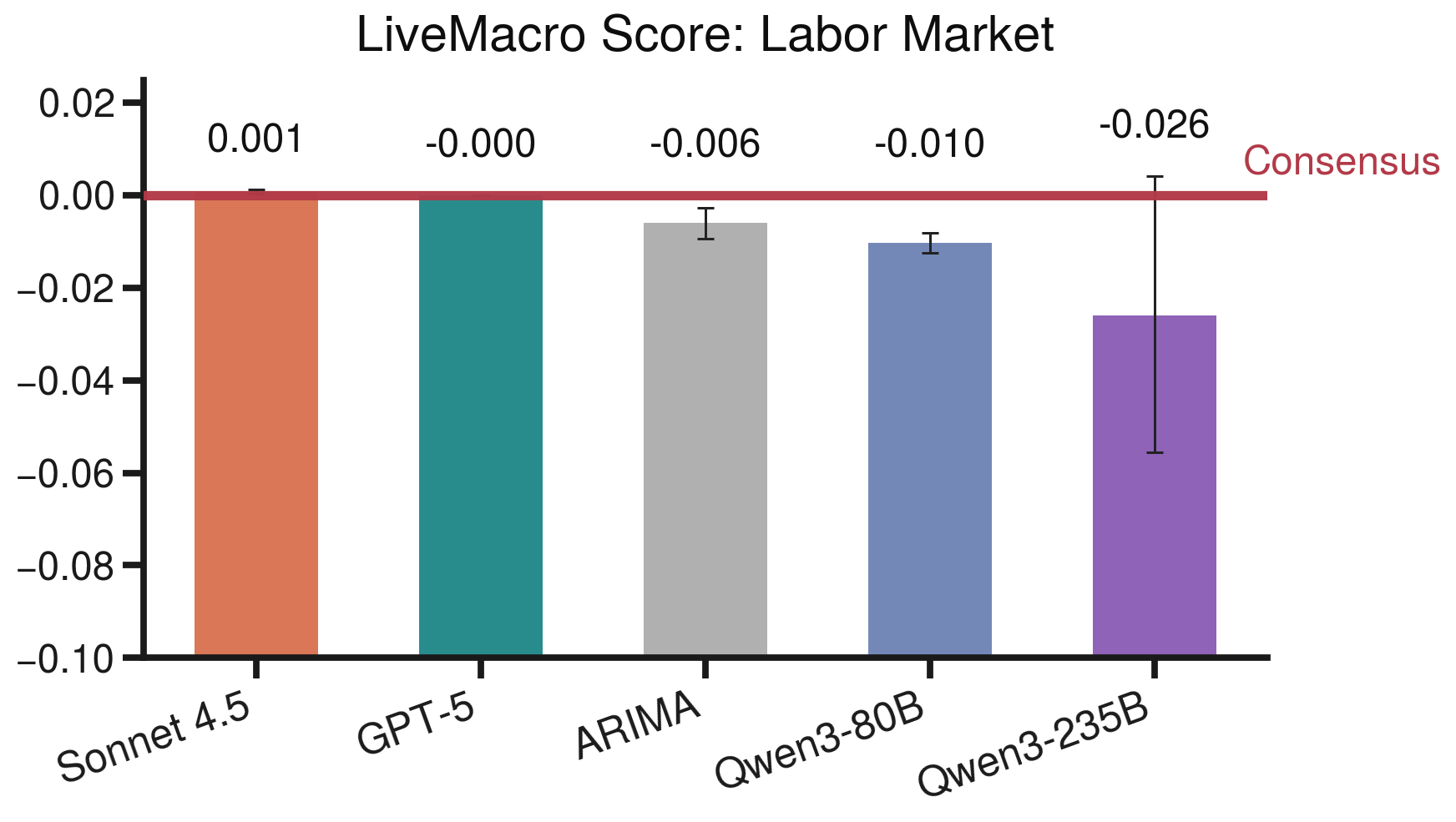}\hfill
\includegraphics[width=0.5\linewidth]{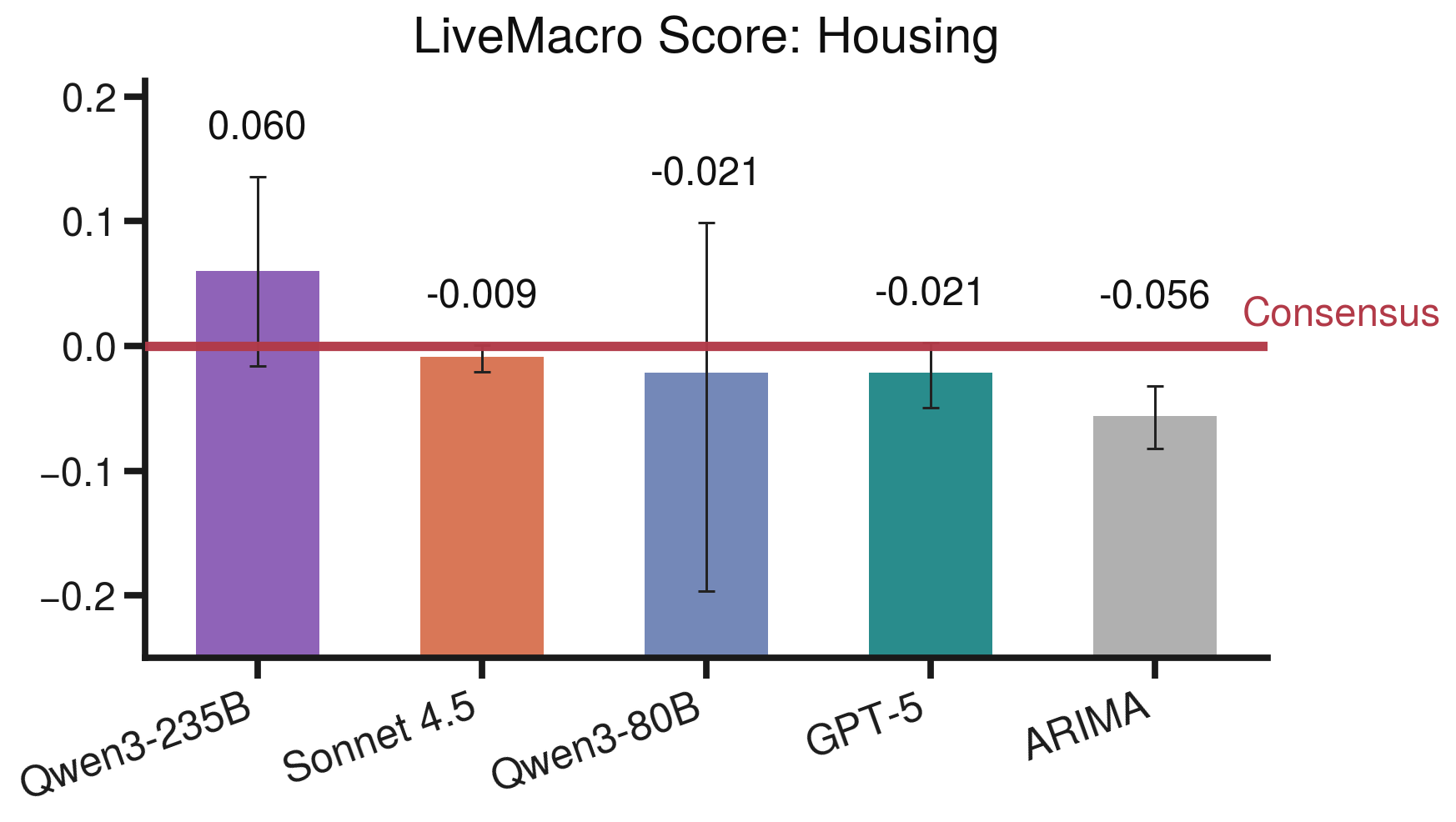}
\caption{LiveMacro Score by theme. Panels show Supply and Production (top-left), Demand and Inflation (top-right), Labor Market (bottom-left), and Housing (bottom-right). Each panel restricts the score to the indicators in the named theme. Positive values indicate that the model exceeds the Bloomberg professional consensus on that theme.}
\label{fig:bdrc-by-theme}
\end{figure*}

\subsection{LiveBetting Score}\label{sec:results-livebetting}

Figure~\ref{fig:polymarket-cum-claude} reports cumulative LiveBetting returns across the February and March target windows for the three indicators with matched institutional Fed nowcasts: real GDP (top), CPI (middle), and the unemployment rate (bottom). Each cumulative return aggregates a continuous stream of hourly \$1 bets within the pre-release window (e.g., $1{,}423$ bets in GPT-5's March real-GDP window), rather than a single binary release-date outcome. The reported curve is therefore a cumulative quantity over more than a thousand bets, and is well-defined only because the agent nowcasts at hourly cadence. On real GDP, the top LLM agents are comparable to the top economist nowcasts: four LLM agents finish the window with positive cumulative returns and three of them beat the Atlanta Fed, with only the New York Fed ahead. On the unemployment rate, the leading LLM agent finishes above the Chicago Fed. Together these two indicators suggest that LLM agents are competitive with the corresponding institutional Federal Reserve nowcasts. On CPI, however, the top economist nowcasts still outperform all LLM agents: the Cleveland Fed and the Bloomberg consensus lead the panel, likely because inflation prediction is a long-standing core focus for economists and relies heavily on established structural signals such as central-bank communications and the well-anchored U.S.\ inflation regime \citep{bernanke2007inflation,faustwright2013forecasting,knotek2017nowcasting}. This gap points to a current limitation of LLM nowcasting on indicators where institutional models are tightly tuned to known economic structure. A detailed per-indicator analysis is in Appendix~\ref{app:livebetting-results-details}.

\subsection{LiveMacro Score by Theme}\label{sec:results-livemacro-theme}

Figure~\ref{fig:bdrc-by-theme} decomposes the LiveMacro Score across the four indicator themes. Each theme corresponds to a block of the U.S.\ economy, so the theme-level view tests whether a model specializes in particular sectors. The decomposition reveals the heterogeneity across themes. On Supply and Production and on Housing, LLM agents materially \emph{exceed} the Bloomberg consensus, led by GPT-5 and Qwen3-235B, respectively (Appendix~\ref{app:livemacro-theme-results-details}). A plausible mechanism is that these themes rely on underlying release components published at higher frequency (e.g., the mortgage applications index and the housing sentiment survey), so the intra-window updates provide fresh information for LLM agents to extract additional signal. On Demand and Inflation, LLM agents \emph{underperform} the consensus, with only GPT-5 sitting near zero. This is consistent with the anchored-expectations regime that protects the professional consensus on headline inflation \citep{bernanke2007inflation,faustwright2013forecasting}. The Labor Market scores cluster around zero across every model, consistent with a signal-to-noise ceiling on the household survey from which the unemployment rate is computed. GPT-5's overall position therefore reflects strong ability on Supply and Production combined with avoidance of the large negative Demand-and-Inflation scores that drag down the other LLM agents. Per-model scores and additional discussion are provided in Appendix~\ref{app:livemacro-theme-results-details}, and Appendix~\ref{app:error-analysis} consolidates the errors the agents share and each model's own pattern of error.

\begin{figure}[h!]
\centering
\includegraphics[width=\linewidth]{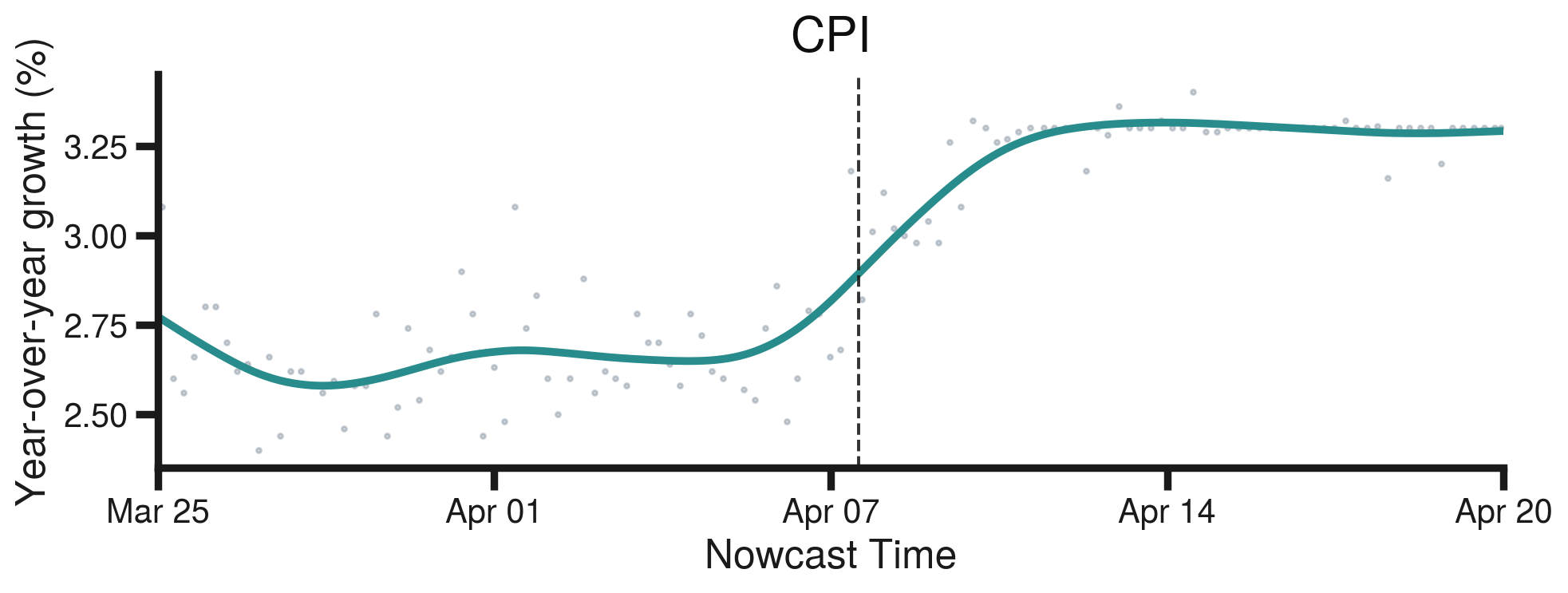}\\[4pt]
\includegraphics[width=\linewidth]{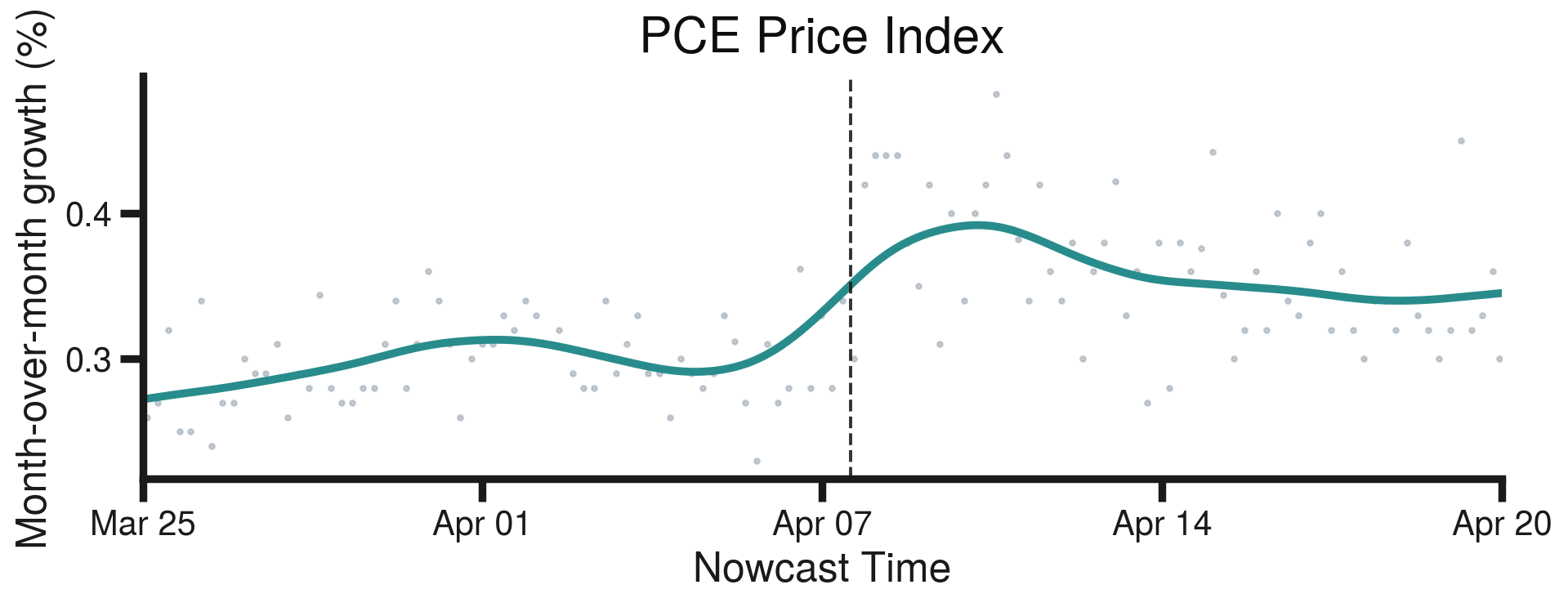}
\caption{GPT-5 nowcasts for March~2026 CPI (top) and PCE price index (bottom). The dashed line at 10:30 ET on April~8, 2026 marks the start of the intraday event window discussed in Section~\ref{sec:case-study}.}
\label{fig:case-cpi-pce}
\end{figure}

\subsection{Case Study: Information Shocks and Nowcast Revisions}\label{sec:case-study}

A distinctive feature of LiveMacroEval is that its hourly nowcasting cadence makes it possible to trace, ex post, how individual macro-information events enter the agent's information set and revise its nowcast, improving interpretability while facilitating economic scientific discovery. As an illustration, Figure~\ref{fig:case-cpi-pce} shows that GPT-5's March~2026 CPI nowcast cluster (top) sits around $2.7\%$ through early April, then jumps to around $3\%$ within a tight intraday window on April~8, 2026 and tracks the realized print thereafter. The PCE nowcast (bottom) shows an analogous upward shift in the same window. Two scheduled, publicly time-stamped events fall inside that window. At 10:30 ET the EIA Petroleum Status Report reported a larger-than-expected drawdown in U.S.\ gasoline inventories alongside a production contraction, an upward signal on the gasoline component of CPI under the competitive storage framework \citep{williams1991storage,deaton1992behaviour}. At 14:00 ET the FOMC meeting minutes explicitly flagged upside inflation risks, the ``central-bank-information component'' channel \citep{rosa2013financial,jarocinski2020deconstructing}. Both events push the nowcast in the same upward direction, and the observed revision is consistent in magnitude with a back-of-the-envelope gasoline-component contribution under the BLS expenditure weights. The event reconstruction and component-contribution calculation are in Appendix~\ref{app:case-study-details}.

\subsection{Tool and Agent Design}\label{sec:tool-agent-design}

We evaluate two agent designs beyond the single web-search agent. The first is a \emph{tool-augmented agent}, the base model with a commercial financial-analysis plug-in. The second is a \emph{multi-agent team}, in which an orchestrator delegates to a researcher and an independent verifier. Both use the same base model, Claude-sonnet-4.5, and live protocol as the plain-prompt control. The evaluation window runs from July to August~2026, after the November~2025 to March~2026 window of Figure~\ref{fig:bdrc}.

The tool-augmented agent ties the plain-prompt control (Table~\ref{tab:agent-design}). Its point estimate is higher, but the confidence interval on the paired difference includes zero, so the difference is not significant. The multi-agent team is the only one of the three configurations with a positive LiveMacro Score, and so the only one to improve on the consensus baseline. Two properties explain the difference. First, the team predicts smaller surprises. Its predicted surprises average $0.26$ in units of each indicator's historical surprise scale, against $0.62$ for those that actually occurred, while the plain prompt averages $0.97$ and overshoots. Second, it gets the direction right more often, on $54\%$ of the individual indicator releases against the plain prompt's $38\%$. Size and direction trade off against each other. Departing from the consensus pays only when the departure is in the right direction often enough to cover the times it is not, so large predicted surprises at a near coin-flip hit rate will score below the consensus line. Appendix~\ref{app:kappa-phi} states this trade-off exactly. Of the three configurations, only the team balances size against direction well enough to beat the consensus. The team reconciles two independent estimates, which may shrink the predicted surprise. The full results are reported in Appendix~\ref{app:agent-design-details}.

\begin{table}[t]
\centering
\footnotesize
\setlength{\tabcolsep}{6pt}
\begin{tabular}{@{}lr@{}}
\toprule
\textbf{Configuration} & \textbf{LiveMacro Score} \\
\midrule
plain prompt (control)    & $-0.080$ \\
\;\;$+$ financial plug-in & $-0.055$ \\
\;\;$+$ multi-agent team  & $\mathbf{+0.020}$ \\
consensus baseline        & $0.000$ \\
\bottomrule
\end{tabular}
\caption{The two agent designs against a plain-prompt control, on the same base model and live protocol. The consensus baseline scores zero by construction. Paired differences and full results are in Appendix~\ref{app:agent-design-details}.}
\label{tab:agent-design}
\end{table}

\section{Conclusion}

We introduced LiveMacroEval, a live, contamination-resistant benchmark for evaluating LLM agents on macroeconomic nowcasting, with two market-based metrics, LiveMacro Score and LiveBetting Score. Over a six-month evaluation, top LLM agents tie the Bloomberg professional consensus on the aggregate LiveMacro Score, and outperform the corresponding institutional Federal Reserve nowcasts on the LiveBetting Score for several indicators, indicating the potential of LLM agents as a real-time technology for macroeconomic nowcasting. The framework's high-frequency cadence further supports event-study-style attribution of nowcast revisions, enabling interpretable analysis of belief updating and surfacing candidate signals for economic scientific discovery.

\section*{Limitations}

Our results are subject to two main limitations. First, the live evaluation contains a limited number of release events. This is a structural constraint of contamination-resistant macroeconomic nowcasting rather than a dataset-size choice: high-value releases arrive only monthly or quarterly, and falling back to historical releases would compromise the live pre-release design. We partially mitigate the constraint through hourly nowcasts within each window, coverage of sixteen indicators, multiple expert baselines, and continued operation of LiveMacroEval so the sample expands over time. The limitation cannot be fully eliminated without sacrificing temporal validity, and it reflects the same sparse-release regime under which professional economist nowcasts are themselves produced, evaluated, and used for decision-making. Although sparse-event evaluation is atypical for traditional NLP benchmarks built on large static datasets, LiveMacroEval is an early attempt at a new mission for NLP: evaluating language agents on decision-relevant live tasks where evidence and outcomes arrive sequentially and cannot be replayed without losing temporal validity.

Second, the indicators and reference nowcasts are all U.S.-denominated. Extending the protocol to non-U.S.\ macroeconomies, each with its own release calendars, consensus surveys, and market reactions, is a natural extension we leave to future work. The euro area is the clearest case, where each of the three ingredients LiveMacroEval needs has a direct analogue, so the protocol ports without conceptual change. The targets and their release calendar carry over. Eurostat and the national statistical offices publish pre-scheduled indicators such as unemployment and industrial production, alongside timely surveys such as the PMIs and the ZEW and ifo indices. The surprise components of such releases move euro-area asset prices within minutes of release, most of the adjustment landing in the first five minutes \citep{andersson2009which,andersen2007realtime}, the high-frequency event-study regularity the LiveMacro Score relies on. The consensus baseline carries over, with the ECB Survey of Professional Forecasters \citep{garcia2003ecbspf} playing the role the Bloomberg ECOS consensus plays here. The institutional comparators carry over, with the Eurosystem producing real-time euro-area nowcasts \citep{angelini2011euro,banbura2013nowcasting} that mirror the Federal Reserve regional-bank nowcasts.

\section*{Acknowledgments}

We thank the Georgia Tech Library for providing access to the Bloomberg Terminal. We also thank the Partnership for an Advanced Computing Environment (PACE) at the Georgia Institute of Technology for the computing resources.

\bibliography{references}

\clearpage

\appendix

% !TEX root = emnlp_2026.tex

% Local environment for the prompt figure in Section~\ref{app:prompt}.
% tcolorbox is loaded by the main file (\usepackage[most]{tcolorbox}).
\newtcolorbox{datacollectionbox}[1][]{%
  enhanced,
  colback=gray!5,
  colframe=black!60,
  boxrule=0.5pt,
  arc=2pt,
  left=4pt, right=4pt, top=4pt, bottom=4pt,
  #1
}

\section{Extended Related Work}\label{app:related-work-extended}

This appendix expands the four threads of related work summarized in Section~\ref{sec:related-work}.

\subsection{Nowcasting in Empirical Macroeconomics}
Methodologically, nowcasting sits at the frontier of empirical macroeconomic forecasting. It is the natural setting in which to exploit the asynchronous, mixed-frequency calendar on which the economy is observed: nowcasting frameworks such as dynamic factor models, mixed-data-sampling (MIDAS) regressions, and their big-data extensions incorporate higher-frequency indicators in real time as they are released \citep{giannone2008nowcasting,marcellino2010factormidas,bok2018nowcasting,banbura2013nowcasting}, an option not available to conventional same-frequency time-series models. The same logic extends to non-traditional, high-dimensional signals: web search activity \citep{choivarian2012google} and machine-readable news text \citep{thorsrud2020words} both deliver incremental nowcasting power over standard macro indicators. Recent machine-learning methods push the frontier further, including sparse-group LASSO for mixed-frequency high-dimensional nowcasting \citep{babii2022machine} and non-parametric mixed-frequency VARs for macroeconomic nowcasting \citep{huber2023nowcasting}. On the institutional side, major central banks therefore produce dedicated nowcast products, most prominently the New York Fed Staff Nowcast \citep{nyfed_nowcast}, the Atlanta Fed GDPNow \citep{atlfed_gdpnow}, the European Central Bank's short-term euro-area GDP nowcast \citep{angelini2011euro}, and the Cleveland Fed inflation nowcast \citep{knotek2017nowcasting}. Methodological progress in macroeconomic forecasting therefore increasingly concentrates on the nowcasting setting rather than on retrospective prediction.

\subsection{LLMs and Machine Learning for Economic and Financial Forecasting}
A growing literature applies LLMs to economic and financial forecasting. \citet{carriero2024macroeconomic} benchmark LLM forecasts against time-series models on FRED-MD, \citet{fariaecastro2025artificial} prompt frontier models for U.S.\ CPI inflation with accuracy comparable to the SPF and Greenbook, and \citet{iadisernia2025prompting} match ECB SPF forecasters by prompting GPT-4o with synthetic economist personas. \citet{ganum2026how} evaluate frontier GPT models on structured macrofinancial questions from IMF Article IV reports, while \citet{lopezlira2023can} show ChatGPT sentiment over news headlines predicts subsequent stock returns. However, these studies evaluate on historical data whose ground truth is plausibly within the training distribution, leaving their results vulnerable to memorization-driven contamination; in contrast, LiveMacroEval evaluates LLMs strictly before each official release, guaranteeing a contamination-free assessment.

\subsection{LLM Memorization and Training-Data Contamination}
Static LLM benchmarks are increasingly compromised by training-data contamination. \citet{golchin2024timetravel} show via guided-prompt completion that memorization of benchmark instances is pervasive, and \citet{gao2025rewindtime} find that prompted knowledge cutoffs only partially suppress post-cutoff facts. These results imply that for widely reported targets such as headline U.S.\ macroeconomic indicators, any evaluation on historical releases is exposed to memorization-driven leakage, which motivates the live, pre-release evaluation design that LiveMacroEval adopts.

\subsection{Live Future-Prediction Benchmarks}
A complementary line avoids memorization by construction, evaluating LLMs on events unresolved at query time. LiveBench \citep{white2025livebench} mitigates contamination with continuously refreshed questions and automatically verifiable ground truth. \citet{halawi2024approaching} show retrieval-augmented ensembles can approach human forecasters on Metaculus, ForecastBench \citep{karger2025forecastbench} scores LLMs via Brier scores on live prediction-market questions, and FutureX \citep{zeng2026futurex} extends live evaluation to LLM agents across 25 general-domain categories. These benchmarks establish the contamination-free methodology but omit scheduled-release macroeconomic indicators and the matched comparison against institutional Federal Reserve nowcasts, the Bloomberg ECOS professional consensus, and econometric baselines that makes macroeconomic nowcasting uniquely amenable to rigorous evaluation.

\section{Extended Preliminary on Nowcasting}\label{app:nowcasting-background}

This appendix expands the Preliminary in Section~\ref{sec:prelim}.

Government statistical agencies publish key macroeconomic indicators such as GDP, inflation, and unemployment only weeks to months after the reference period they cover. Nowcasting is the task of estimating an indicator's value for the current or just-completed reference period before the agency's official release. Unlike ex-post forecasting on historical data, a nowcast is a running estimate of present economic conditions, revised as new data arrive: as the Federal Reserve Bank of New York describes its own Staff Nowcast, ``[n]owcasts of GDP growth are designed to track the economy in real time by incorporating information from an array of indicators as they are released'' \citep{nyfed_nowcast}. Because the target value has not yet been published at prediction time, it cannot have entered the model's training corpus, making the evaluation contamination-free by construction.

Nowcasts are a primary input to monetary and fiscal policymaking. Headline macroeconomic data arrive with multi-week lags and are subject to repeated revision, leaving central banks and fiscal authorities to act under substantial uncertainty about the current state of the economy \citep{bok2018nowcasting}. In a setting where central-bank forward guidance is itself sensitive to timing, more accurate nowcasts translate directly into better-timed and better-calibrated policy interventions. Major central banks therefore produce dedicated nowcast products, most prominently the New York Fed Staff Nowcast \citep{nyfed_nowcast}, the Atlanta Fed GDPNow \citep{atlfed_gdpnow}, and the European Central Bank's short-term euro-area GDP nowcast \citep{angelini2011euro}.

Beyond their slow-updating institutional counterparts, nowcasts are also constrained in cadence: the New York Fed Staff Nowcast \citep{nyfed_nowcast} and the Chicago Fed CHURN \citep{chicagofed_churn} update weekly, and the Atlanta Fed GDPNow updates only when major source-data releases arrive \citep{atlfed_gdpnow}. The methodological discussion of nowcasting frameworks (dynamic factor models, MIDAS regressions, and their machine-learning extensions) is given in Appendix~\ref{app:related-work-extended} to avoid duplication with the related-work survey there.

\section{Extended Indicator Descriptions}\label{app:indicator-details}

The main paper (Section~\ref{sec:tasks-themes}) introduces each indicator's role. This appendix complements that overview with the institutional features most relevant to nowcasting: the issuing agency, the position in the monthly release calendar, and the methodological points that distinguish closely related series. Units and scale guardrails are tabulated separately in Table~\ref{tab:variable-defs}.

\paragraph{Supply and Production.}
\emph{Real Gross Domestic Product} is the broadest quarterly measure of U.S.\ output and the headline business-cycle reference. It is released by the Bureau of Economic Analysis (BEA) about one month after quarter end, and the advance estimate is the target of every institutional Federal Reserve GDP nowcast \citep{bok2018nowcasting}. \emph{Industrial Production Index} is published monthly by the Federal Reserve Board (G.17 release) and captures physical output in manufacturing, mining, and utilities, with manufacturing accounting for roughly three quarters of the total. It provides the highest-frequency direct read on the goods sector. \emph{Manufacturers' New Orders: Durable Goods} is the standard leading indicator of future production, released by the Census Bureau. Its headline is regularly dominated by volatile transportation orders, notably commercial aircraft, so the ex-transportation core is the cleaner signal. \emph{ISM Manufacturing PMI} is the earliest indicator for any reference month, released on the first business day of the following month. It is a diffusion index over five subcomponents (new orders, production, employment, supplier deliveries, inventories) on a 0 to 100 scale with 50 as the expansion threshold, and it materially improves real-time GDP nowcasts \citep{giannone2008nowcasting}.

\paragraph{Demand and Inflation.}
\emph{Consumer Price Index} is the most visible measure of U.S.\ consumer prices. It is released by the Bureau of Labor Statistics (BLS) around the middle of the following month using a modified Laspeyres formula at the upper level. \emph{PCE Price Index} is the formal benchmark for the Federal Reserve's 2\% inflation target, released by the BEA in the Personal Income and Outlays report about one month after the reference period. Unlike CPI it uses Fisher chain weighting that absorbs consumer substitution across goods as relative prices change, and draws its expenditure weights from the national accounts, which is why the Fed treats it as the operational inflation gauge rather than CPI. \emph{Producer Price Index} captures upstream cost pressures from prices received by domestic producers and is a standard leading indicator of consumer-price inflation, released by the BLS in the same week as CPI. We use the Final Demand aggregation, which replaced the Stage of Processing taxonomy in 2014. \emph{Real Personal Consumption Expenditures} is the inflation-adjusted measure of household spending in the national accounts, accounting for roughly two-thirds of U.S.\ GDP and released alongside the PCE Price Index in the same Personal Income and Outlays report. \emph{Retail Sales, Advance Monthly} provides the earliest monthly read on nominal household spending at retail and food-service establishments, released by Census roughly two weeks after the reference month. It excludes the bulk of services consumption and is therefore an early but partial read on consumer demand. \emph{ISM Services PMI} is the headline real-time indicator of services-sector activity, which accounts for most of U.S.\ value added and employment. It is released on the third business day of the month with diffusion-index construction identical to the manufacturing PMI.

\paragraph{Labor Market.}
\emph{Nonfarm Payrolls} reports the monthly net change in paid employment and is typically the single most closely watched U.S.\ data release. It is released by the BLS on the first Friday of the following month in the Employment Situation report, drawn from the Current Employment Statistics establishment survey that covers approximately 120{,}000 businesses representing over 600{,}000 worksites. \emph{Unemployment Rate} measures the share of the labor force without a job and is a core input to Taylor-rule-style policy analyses. It is drawn from the household-side Current Population Survey of about 60{,}000 households conducted over the week containing the 12th and is reported rounded to one decimal place, which sets a hard granularity floor on the achievable nowcast error.

\paragraph{Housing.}
\emph{Building Permits} is the key forward-looking indicator of the residential pipeline, released by the Census Bureau around the middle of the following month jointly with Housing Starts. Permits lead actual ground-breaking by roughly one to two months. \emph{Housing Starts} is a coincident indicator of residential investment activity, released in the same Census report. \emph{New Home Sales} is the timeliest read on primary-market demand, released near the end of the following month by Census and HUD. It is recorded at contract signing rather than at closing, but the underlying transaction count is small, about 600 thousand units SAAR against roughly 5 million for existing homes, so monthly sampling noise is correspondingly larger. \emph{Existing Home Sales} accounts for the bulk of U.S.\ housing transactions and is released by the National Association of Realtors around the 20th of the following month. It is recorded at closing, which typically trails the underlying purchase contract by 30 to 60 days.

\section{Prompt}\label{app:prompt}

We document the prompts issued to each evaluated agent. Every nowcast call consists of a fixed system message that sets the role and output discipline, followed by a templated user message that supplies the target month, the release month, the set of variables to nowcast (each with its title, unit, official sources, and an order-of-magnitude scale guardrail), and the strict output format. At each hourly timestamp we issue two independent calls per agent, one for the \emph{output core} group (output, inflation, and labor) and one for the \emph{demand and sectoral} group (consumption, housing, and services); their components are listed in Table~\ref{tab:variable-defs}. Splitting the panel keeps each call within a manageable output length and isolates failures to a single group. The system and user messages are identical across all evaluated agents. Figure~\ref{fig:prompts} reports the full system message and user prompt template.

\begin{figure*}[tp]
\centering
\begin{minipage}[t]{0.45\linewidth}
\begin{datacollectionbox}[equal height group=prompts]
\textbf{System prompt.}

{\linespread{0.85}\scriptsize
\begin{verbatim}
You are a real-time, web-searching economist
forecasting macro variables. Use web search to
gather the latest available facts, official
releases, trackers, expert consensus, policy
moves, market moves, and news, etc., including
revisions. Identify information shocks and update
your prediction. Think silently and return ONLY
the requested output format, a single line of
space-separated key=value pairs. No extra
commentary.
\end{verbatim}
}

\vspace{0.3em}

\textbf{User prompt.}

{\linespread{0.85}\scriptsize
\begin{verbatim}
Goal: Nowcast macro variables for <target month>.
- Target month = <target month, YYYY-MM>
  (month the data measures)
- Release month = <release month, YYYY-MM>
  (month of official publication)
- Use the full, up-to-date set of official and
  reputable information available as of the
  system time when this response is generated
  (including releases, revisions, trackers,
  consensus, policy/market moves, and relevant
  news) to produce a live nowcast.
- Forecast all and only the variables in the
  variables-and-units list below.
- Geography: United States (U.S.). Use U.S.
  official releases and reputable consensus
  trackers for the U.S.

Whole-month / whole-quarter logic:
- All variables are MONTHLY and measure
  conditions for the ENTIRE target month,
  EXCEPT GDP variables.
- GDP variables (level, QoQ, and YoY) are
  QUARTERLY, and the following rules apply ONLY
  IF any GDP variable is in the list:
    - If no GDP variables are listed, do NOT
      include GDP in any form.
    - For GDP, the target month identifies the
      QUARTER to be forecast, not a monthly
      value. Each GDP variable refers to the
      calendar quarter that CONTAINS the target
      month (e.g., 2025-12 -> 2025Q4).
    - For GDP, the release month is the
      calendar month in which that quarter's
      GDP data are first officially published
      (e.g., 2025Q4 GDP -> release month
      = 2026-01).
- When updating forecasts, assess how new
  information affects the full target month.
  Take into account how much of the month has
  already elapsed and what remains. Information
  may continue to arrive after the month has
  ended (e.g., official statistical releases).

Variables and units:
You must generate numeric forecasts for ALL
variables listed below, every time. No
additional variables are permitted, and none may
be omitted. Each variable's unit is
authoritatively defined here and must be
strictly followed. If a source reports a
variable in a different unit or scale, convert
it to the unit specified below before output.
<list of variables and units>
\end{verbatim}
}
\end{datacollectionbox}
\end{minipage}\hfill
\begin{minipage}[t]{0.45\linewidth}
\begin{datacollectionbox}[equal height group=prompts]
{\linespread{0.85}\scriptsize
\begin{verbatim}
Official sources:
When provided, the following official sources
describe definitions and historical ranges.
Use them for grounding and calibration, not
verbatim copying.
<list of official sources>

Scale guardrails:
Each variable lists an order-of-magnitude hint
in its requested unit. Treat these as acceptable
scale guardrails and only go outside them when
compelling evidence points to an extreme shock.
<list of scale guardrails>

Calibration rules:
- For each variable, use the order-of-magnitude
  guardrails to sanity-check your draws within
  the stated unit. Only deviate if strong
  evidence supports the shift.
- Negative values are explicitly allowed where
  economically meaningful.
- Never emit placeholder tokens such as
  "missing", "nan", "-999", "9999", or zero
  values unless zero is a defensible forecast.
- If uncertainty exists, resolve it by choosing
  your best numeric estimate in the stated units.
- Failure to meet these constraints is considered
  an incorrect response.

Web search:
- You must pull the most CURRENT information
  (official/statistical releases, expert
  consensus, policy decisions, market moves,
  relevant news, etc).
- If data have been revised, use the most
  recently published revision.
- Prefer official sources when available. If no
  official source is available for a variable,
  rely on web search and standard, reputable
  research sources.

Output format (STRICT):
- Output exactly one line of space-separated
  key=value pairs.
- The first two pairs report the target month
  and the release month, in YYYY-MM format and
  in that order; the remaining pairs report the
  indicators listed above.
- Each key must appear exactly once and must
  EXACTLY match the keys in the list.
- Strictly follow each variable's unit in the
  list.
- For percent variables, report values as
  percentages rather than decimals (e.g., 5.2
  represents 5.2%).
- Formatting: values must be pure numbers only
  (no % sign), rounded to at most 2 decimal
  places. Avoid scientific notation. Never use
  commas.

Example (structure only, do not copy verbatim):
<example output line>

Rules:
- No extra text, no explanation.
- Even if uncertain, provide best estimates
  (never NA/missing/-999 placeholders).
- Think silently; output ONLY the line.
\end{verbatim}
}
\end{datacollectionbox}
\end{minipage}
\caption{Prompts issued to each evaluated agent. The system message sets the role and output discipline. The user prompt template supplies the target month, the release month, the set of variables to nowcast, scale guardrails, calibration and web-search rules, and the strict output format. The right-hand panel continues the user prompt from the left. Placeholders in angle brackets are filled in at call time.}
\label{fig:prompts}
\end{figure*}

\paragraph{Variable definitions.}
Table~\ref{tab:variable-defs} lists the 16 series reported in the paper, in theme order within each prompt group. Each row gives the indicator (with its reporting transformation and seasonal adjustment), its unit, the order-of-magnitude scale guardrail in the variable's native unit, and an abbreviated source label. Source labels abbreviate the underlying release: FRED for the St.\,Louis Fed FRED database, Fed G.17 for the Federal Reserve industrial production release, and ISM for the Institute for Supply Management report on business.

\begin{table*}[tp]
\centering
\footnotesize
\setlength{\tabcolsep}{6pt}
\renewcommand{\arraystretch}{1.2}
\begin{tabular}{@{}p{6.4cm}p{2.6cm}p{3.6cm}l@{}}
\toprule
\textbf{Indicator} & \textbf{Unit} & \textbf{Scale guardrail} & \textbf{Source} \\
\midrule
\multicolumn{4}{@{}l}{\textit{Output core: output, inflation, and labor}} \\
\addlinespace[2pt]
Real GDP (QoQ \%, SAAR)                       & percent              & percent                                                  & FRED \\
Industrial Production (MoM \%, SA)            & percent              & percent                                                  & Fed G.17 \\
Durable Goods Orders (MoM \%, SA)             & percent              & percent                                                  & FRED \\
ISM Manufacturing PMI (SA)                    & index                & 2-digit, $O(10^{1}$ to $10^{2})$                         & ISM \\
CPI: All Items (YoY \%)                       & percent              & percent                                                  & FRED \\
PPI: Final Demand (MoM \%, NSA)               & percent              & percent                                                  & FRED \\
PCE Price Index (MoM \%, SA)                  & percent              & percent                                                  & FRED \\
Nonfarm Payrolls (MoM change, SA)             & thousand persons     & 2 to 3-digit typical, up to 5-digit in crises, $O(10^{2}$ to $10^{5})$ & FRED \\
Unemployment Rate (SA, \%)                    & percent              & percent                                                  & FRED \\
\midrule
\multicolumn{4}{@{}l}{\textit{Demand and sectoral activity: consumption, housing, and services}} \\
\addlinespace[2pt]
Real PCE (MoM \%, SA)                         & percent              & percent                                                  & FRED \\
Retail and Food Services Sales (MoM \%, SA)   & percent              & percent                                                  & FRED \\
Housing Starts (SAAR level)                   & thousand units, SAAR & 4-digit, $O(10^{3}$ to $10^{4})$                         & FRED \\
Building Permits (SAAR level)                 & thousand units, SAAR & 3 to 4-digit, $O(10^{3}$ to $10^{4})$                    & FRED \\
Existing Home Sales (SAAR level)              & thousand units, SAAR & 4-digit, $O(10^{3}$ to $10^{4})$                         & FRED \\
New Home Sales (SAAR level)                   & thousand units, SAAR & 3 to 4-digit, $O(10^{3}$ to $10^{4})$                    & FRED \\
ISM Services PMI (SA)                         & index                & 2-digit, $O(10^{1}$ to $10^{2})$                         & ISM \\
\bottomrule
\end{tabular}
\caption{Variable definitions for the 16 series reported in the paper, by prompt group. Each row gives the indicator (with its reporting transformation and seasonal adjustment), its unit, the order-of-magnitude scale guardrail in the variable's native unit, and an abbreviated source label.}
\label{tab:variable-defs}
\end{table*}
\section{Post-Release Recovery}\label{app:postrelease}

\begin{figure*}[tp]
\centering
\includegraphics[width=0.8\linewidth]{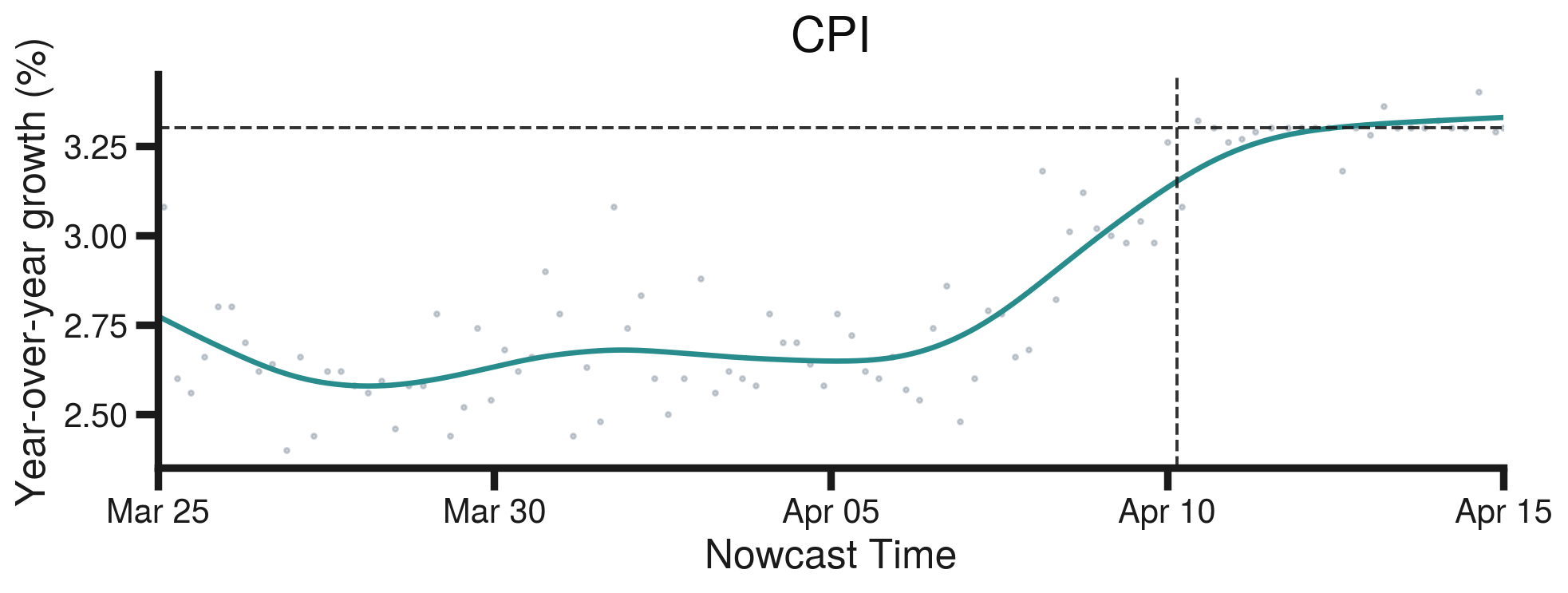}\\[4pt]
\includegraphics[width=0.8\linewidth]{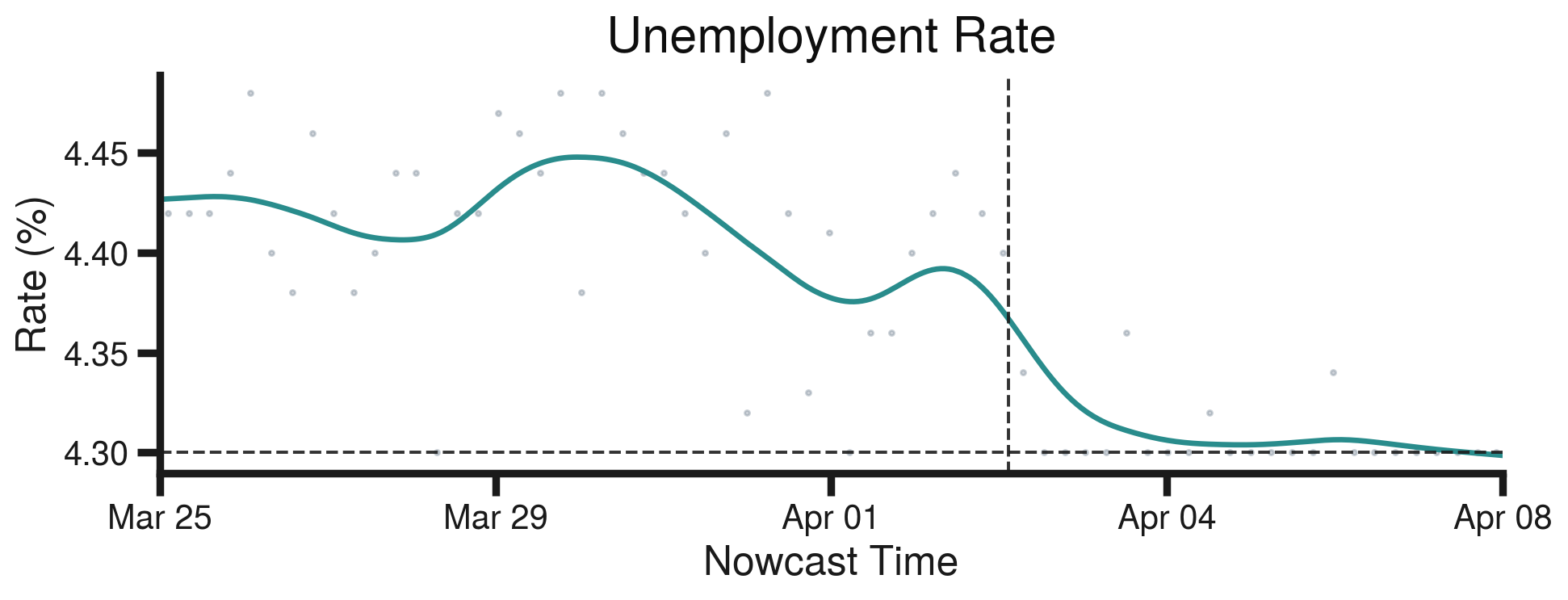}\\[4pt]
\includegraphics[width=0.8\linewidth]{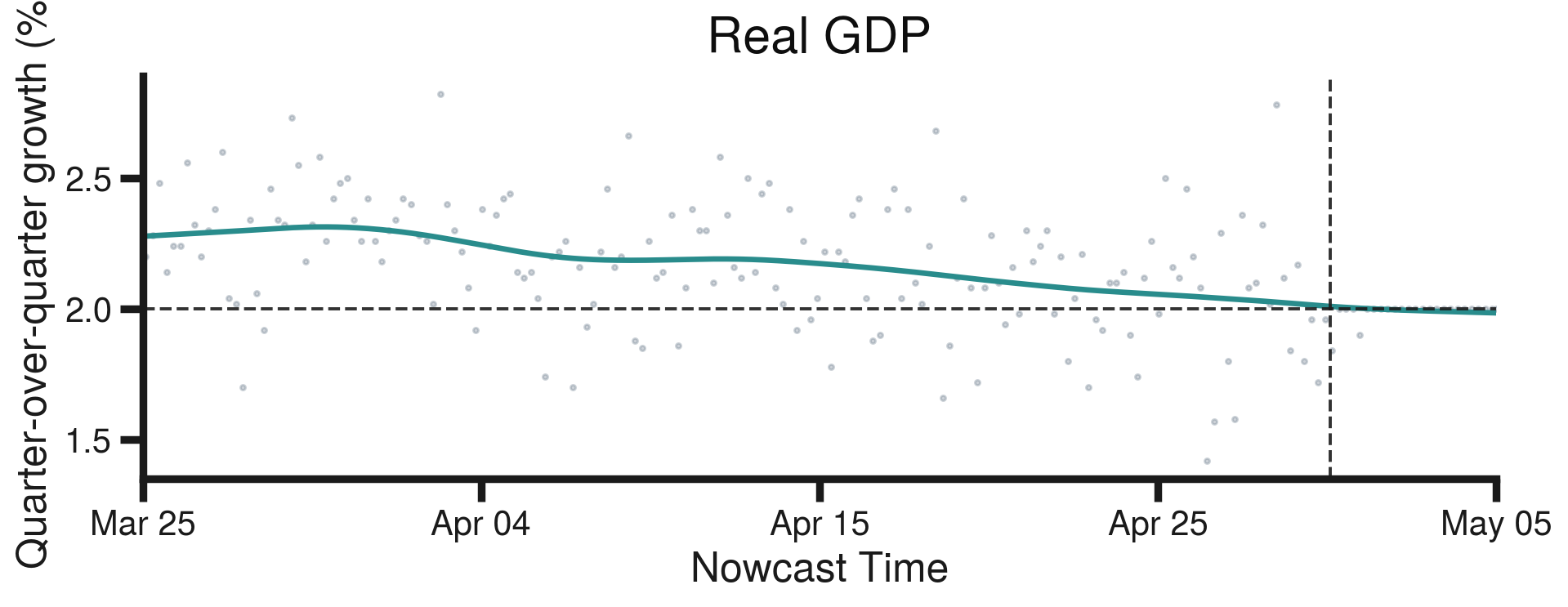}
\caption{Post-release recovery. GPT-5 nowcasts for three target releases continued for five days past the official release: CPI year-over-year for the March~2026 print (top, released April~10, 2026), the unemployment rate for the March~2026 print (middle, released April~3, 2026), and real GDP quarter-over-quarter for the 2026Q1 advance estimate (bottom, released April~30, 2026). Light markers are raw hourly nowcasts. The solid teal line is a cubic smoothing spline with $\lambda$ chosen by generalized cross-validation. The dashed vertical line marks the official release timestamp, and the dashed horizontal line marks the released value.}
\label{fig:postrelease}
\end{figure*}

Figure~\ref{fig:postrelease} reports GPT-5 nowcasts continued for five days past the official release on the three high-attention indicators traded on Polymarket: CPI year-over-year growth rate for the March~2026 print, the unemployment rate for the March~2026 print, and real GDP quarter-over-quarter growth rate for the 2026Q1 advance estimate. The plotting convention matches Figure~\ref{fig:case-cpi-pce}. Across all three series, the post-release nowcast cluster snaps onto the released level within the first day after release and remains there with residual dispersion small relative both to the pre-release cluster spread and to the release-versus-prior-cluster gap. The March CPI nowcast cluster shifts from a roughly $2.6\%$ pre-release band to the $3.3\%$ release. The March unemployment rate snaps from a $4.40$ to $4.45\%$ band to $4.30\%$. The 2026Q1 GDP nowcast collapses from a $2.1$ to $2.3\%$ band to $2.0\%$. The agent's web-search pipeline therefore reliably reads the official value once it is published and reports it in the requested numeric format.

Two mechanisms could in principle drive the pre-release dispersion. First, the agent retrieves all publicly available information but a residual uncertainty about the as-yet-unreleased value leaves a genuine nowcast error. Second, the agent fails to retrieve a publicly available pre-release signal, so the observed dispersion is a retrieval problem rather than a forecasting problem. Because the same retrieval pipeline cleanly recovers the released value on these indicators within hours of publication, the dominant pre-release failure mode is unlikely to be retrieval. This is consistent with reading the LiveMacro and LiveBetting scores as measurements of nowcasting skill rather than information-retrieval competence.

\section{Are LLM Agents Copying Institutional Nowcasts?}\label{app:contamination-check}

A natural concern is that web-search-enabled LLM agents may retrieve and copy the Bloomberg consensus or the Federal Reserve regional-bank nowcasts, rather than producing independent estimates. This is not the central concern of LiveMacroEval for three reasons. First, the Bloomberg ECOS consensus is distributed solely through the paid Bloomberg Terminal subscription \citep{bloomberg2024ecos} and the cumulative-median consensus used as our comparator is not republished in real time on the open web, so the agents cannot directly retrieve it via web search. Second, even setting Bloomberg aside, the LiveBetting results in Section~\ref{sec:results-livebetting} already show that LLM agent and Federal Reserve nowcast paths diverge enough to generate materially different cumulative Polymarket returns using the same betting strategy. This would not be observable if the agents were simply tracking the Fed series. Third, the agents produce hourly nowcasts that revise on every incoming information event, whereas the Federal Reserve regional-bank nowcasts publish on weekly or event-triggered schedules with multi-day gaps. The dense intra-window revision path the agents produce is intrinsically beyond what a copy-of-institutional-nowcast strategy can deliver, and producing it without an LLM agent would require continuous human analyst labor.

Although copying is therefore not our main concern, we report three direct tests to show that the agents are not in fact copying the Bloomberg or Federal Reserve series.

\paragraph{Setup and notation.}
All three tests compare one agent's nowcast series against one public forecast series for the same indicator and reference month, and all three truncate strictly at the release timestamp, so nothing observed after the release can enter. We call each (agent, indicator, reference month) triple a \emph{series} and compute every statistic within a series before pooling, so a heavily covered indicator cannot dominate. Write $\hat X_t$ for the agent's nowcast at hour $t$, transformed into the comparator's units, and $P_t$ for the public forecast at that time, which is the Bloomberg consensus or a Federal Reserve nowcast depending on the comparator. Two quantities carry the tests. The \emph{gap} $G_t = |\hat X_t - P_t| / \sigma_i$ is how far the agent's level sits from the public forecast, divided by the release-surprise scale $\sigma_i$ of Appendix~\ref{app:capture-score-details} so that it is comparable across indicators, and we quote gaps in units of $\sigma$ throughout. The \emph{revision} $\Delta_t = \hat X_t - \hat X_{t-1}$ is the change the agent makes between consecutive hourly nowcasts.

Each test asks a different question of these two quantities. The convergence test reads the level, whether $G_t$ falls toward zero as the release nears. The first-difference test reads co-movement, whether the agent's revisions move together with the comparator's. The event-time test reads timing, whether the agent's largest revisions land when the Federal Reserve publishes. A copier would answer yes to all three. Tests~2 and~3 are restricted to CPI and real GDP, which have Federal Reserve regional-bank nowcast coverage and a long pre-release nowcasting window. Each table below defines the statistic it reports.

\paragraph{Test 1: level convergence.}
This test asks whether an agent's nowcast moves onto the public forecast as the release approaches, which is what a copier or an averager would do. Table~\ref{tab:contam-convergence} reports the median \emph{terminal gap}, meaning $G_t$ evaluated on the last nowcast before the release, together with the median within-series Spearman correlation $\rho_{\mathrm{S}}$ between $G_t$ and the hours remaining until release. Copying would leave two marks, a terminal gap near zero and a clearly positive $\rho_{\mathrm{S}}$, meaning the gap closes as the release nears. Neither appears. Across all sixteen indicators the final nowcast sits a median $0.42\sigma$ from the Bloomberg consensus, at least $0.5\sigma$ away in $46\%$ of series and at least $1\sigma$ away in $25\%$, and it sits farther still from every Federal Reserve nowcast except the Chicago Fed unemployment nowcast. The dynamics show no closing either. Against Bloomberg $\rho_{\mathrm{S}} = -0.001$, and the median gap plateaus between $0.39\sigma$ and $0.58\sigma$ across every time-to-release bin from two weeks out to the final six hours, never approaching zero (Table~\ref{tab:contam-gap-time}). Against the Cleveland Fed CPI nowcast, Atlanta Fed GDPNow and the Chicago Fed unemployment nowcast, $\rho_{\mathrm{S}}$ is negative, so the gap widens as the release nears, which is the opposite of convergence.

\paragraph{Test 2: co-movement of revisions.}
The first-difference correlation $r_\Delta$ between each LLM agent's daily nowcast series and each comparator's daily series is small for every (model $\times$ variable $\times$ comparator) pair. Across the cells in Table~\ref{tab:contam-fd} the median absolute correlation is $0.11$, and the sign of $r_\Delta$ is split exactly ten to ten across the twenty cells, which is what independent noise produces rather than copying. 

\paragraph{Test 3: timing of large revisions.}
The event-time concentration test in Table~\ref{tab:contam-event} provides the strongest evidence. For each Federal Reserve agency update, we ask whether GPT-5's large revisions, those in the top $10\%$ by $|\Delta_t|$, cluster within a $\pm 24$h window of the update, with $\Delta\mathrm{share} > 0$ being the exact pattern copying would produce. None of the three Fed cells for GPT-5 is positive-and-significant. The two significant cells (Cleveland Fed on CPI and Atlanta Fed GDPNow on real GDP) both carry $\Delta\mathrm{share} < 0$ (anti-clustering, the opposite of that pattern), and the NY Fed Staff Nowcast cell on real GDP is non-significant. For Atlanta Fed GDPNow, GPT-5 shows $\Delta\mathrm{share} = -0.04$ in the anti-clustering direction, so its large revisions land away from GDPNow publication events rather than concentrating around them. Taken together, none of the three tests finds what a web-search agent averaging or copying institutional and consensus nowcasts would leave behind.

\begin{table}[htbp]
\centering
\footnotesize
\setlength{\tabcolsep}{4pt}
\begin{tabular}{@{}llrrr@{}}
\toprule
\textbf{Comparator} & \textbf{Target} & \textbf{Series} & \textbf{Gap} & $\rho_{\mathrm{S}}$ \\
\midrule
Bloomberg ECOS      & all $16$ & $196$ & $0.42$ & $-0.00$ \\
Cleveland Fed CPI   & CPI      & $13$  & $0.82$ & $-0.16$ \\
Atlanta Fed GDPNow  & GDP      & $6$   & $0.75$ & $-0.16$ \\
NY Fed staff        & GDP      & $6$   & $1.66$ & $+0.04$ \\
Chicago Fed CHURN   & unemp.   & $13$  & $0.18$ & $-0.15$ \\
\bottomrule
\end{tabular}
\caption{Level-convergence test. ``Target'' is the indicator the comparator forecasts. ``Gap'' is the median terminal gap, the absolute distance between the agent's last pre-release nowcast level and the public forecast level, in units of the release-surprise scale $\sigma_i$, so a copier would show $0$. $\rho_{\mathrm{S}}$ is the median within-series Spearman correlation between the absolute gap and hours remaining until release, so $\rho_{\mathrm{S}} > 0$ means the gap closes toward release (convergence) and $\rho_{\mathrm{S}} < 0$ means it widens (divergence).}
\label{tab:contam-convergence}
\end{table}

\begin{table}[htbp]
\centering
\footnotesize
\setlength{\tabcolsep}{6pt}
\begin{tabular}{@{}lrr@{}}
\toprule
\textbf{Hours before release} & \textbf{Gap} & \textbf{Series} \\
\midrule
$336+$ (14d$+$)   & $0.49$ & $62$ \\
$168$ to $336$    & $0.53$ & $143$ \\
$96$ to $168$     & $0.58$ & $177$ \\
$48$ to $96$      & $0.53$ & $177$ \\
$24$ to $48$      & $0.45$ & $179$ \\
$12$ to $24$      & $0.39$ & $92$ \\
$6$ to $12$       & $0.46$ & $157$ \\
$0$ to $6$        & $0.41$ & $98$ \\
\bottomrule
\end{tabular}
\caption{Median absolute gap to the Bloomberg consensus in units of $\sigma_i$, by time remaining until release, with each series weighted equally within a bin. A copying or averaging agent would fall toward $0$ in the bottom rows. The gap instead plateaus near half a surprise standard deviation throughout.}
\label{tab:contam-gap-time}
\end{table}

\begin{table*}[tp]
\centering
\footnotesize
\setlength{\tabcolsep}{5pt}
\renewcommand{\arraystretch}{1.15}
\begin{tabular}{@{}llrrrr@{}}
\toprule
\textbf{Variable} & \textbf{Comparator} & \textbf{GPT-5} & \textbf{Claude-sonnet-4.5} & \textbf{Qwen3-235B} & \textbf{Qwen3-80B} \\
\midrule
CPI (YoY)          & Bloomberg ECOS                & $+0.11$ & $+0.02$ & $-0.10$ & $+0.17$ \\
CPI (YoY)          & Cleveland Fed Nowcast         & $+0.10$ & $-0.13$ & $-0.01$ & $+0.30$ \\
Real GDP (QoQ)     & Bloomberg ECOS                & $-0.30$ & $-0.11$ & $-0.37$ & $+0.57$ \\
Real GDP (QoQ)     & Atlanta Fed GDPNow            & $-0.03$ & $-0.03$ & $-0.14$ & $-0.01$ \\
Real GDP (QoQ)     & NY Fed Staff Nowcast          & $+0.17$ & $+0.37$ & $+0.01$ & $+0.03$ \\
\bottomrule
\end{tabular}
\caption{First-difference co-movement test. $r_\Delta$ is the Pearson correlation between the agent's day-over-day changes and the comparator's, computed on the release-truncated daily series and pooled across reference periods. It is a correlation of \emph{changes}, not of levels, so it asks whether the two revise together rather than whether they sit at similar values. An agent copying the comparator would revise when it revises and produce $r_\Delta \approx 0.8$ to $1.0$. Values near zero with signs split evenly are what independent noise looks like.}
\label{tab:contam-fd}
\end{table*}

\begin{table*}[tp]
\centering
\footnotesize
\setlength{\tabcolsep}{6pt}
\renewcommand{\arraystretch}{1.15}
\begin{tabular}{@{}llrrrrr@{}}
\toprule
\textbf{Variable} & \textbf{Fed Comparator} & $\Delta\mathrm{share}$ & \textbf{perm $p$} & \textbf{KS $p$} & \textbf{\# Agent Revisions} & \textbf{\# Fed Updates} \\
\midrule
CPI (YoY)         & Cleveland Fed Nowcast  & $-0.10$ & $<\!0.001$ & $0.08$ & $2{,}533$ & $29$ \\
Real GDP (QoQ)    & Atlanta Fed GDPNow     & $-0.04$ & $0.04$     & $0.07$ & $4{,}289$ & $26$ \\
Real GDP (QoQ)    & NY Fed Staff Nowcast   & $-0.00$ & $0.96$     & $0.36$ & $4{,}289$ & $41$ \\
\bottomrule
\end{tabular}
\caption{Event-time concentration test for GPT-5 against the Federal Reserve nowcasts. A revision is \emph{large} if $|\Delta_t|$ falls in the top $10\%$ for that series. $\Delta\mathrm{share}$ is the fraction of large revisions landing within $\pm 24$h of a Fed update, minus the same fraction computed over all revisions, so it is the excess concentration of big moves around Fed publication times. $\Delta\mathrm{share} > 0$ is the copying pattern and $\Delta\mathrm{share} < 0$ means large revisions avoid those moments. \emph{perm $p$} is a two-sided permutation $p$-value from $B = 1{,}000$ random reassignments of the revision magnitudes across the observed revision timestamps. \emph{KS $p$} is a two-sided Kolmogorov--Smirnov test comparing the distribution of $|\Delta_t|$ inside the windows against outside. Fed updates within two days of a previously kept update are dropped so the windows stay disjoint.}
\label{tab:contam-event}
\end{table*}

\section{What the Nowcast Rationales Contain}\label{app:reasoning-traces}

This appendix asks what an agent says it did, using the reasoning output of GPT-5. The agent emits a short prose rationale with every value it reports, and we collected $7{,}078$ of these blocks from $305$ nowcast runs on two independent servers, covering the target months June through August~2026. The median block runs $262$ characters. These are self-reports, so they record what the agent states about its own procedure rather than what produced the number, and we read them on those terms. 

\paragraph{Reported behavior.}
Table~\ref{tab:traces-behaviour} reports what the rationales contain. A majority name a statistical agency, meaning the BLS, the BEA, the Census Bureau or the ISM, at $56.2\%$. In $47.0\%$ of the blocks the agent exercises explicit judgement over its own figure. A further $12.1\%$ state the vintage of the data, which matters for growth-based targets because the calculation depends on which prior release is used as the base. Explicit arithmetic is reported in $9.1\%$, where the agent computes a metric rather than reporting one it retrieved.

\begin{table}[htbp]
\centering
\footnotesize
\setlength{\tabcolsep}{4pt}
\begin{tabular}{@{}lrr@{}}
\toprule
 & \textbf{Share} & $\mathbf{n}$ \\
\midrule
names a statistical agency      & $56.2\%$ & $3{,}979$ \\
judges its own figure           & $47.0\%$ & $3{,}324$ \\
states the data vintage         & $12.1\%$ & $858$ \\
performs explicit arithmetic    & $9.1\%$  & $647$ \\
\bottomrule
\end{tabular}
\caption{What the $7{,}078$ rationale blocks contain, ordered by share. The behaviors can co-occur within one block, so the shares do not sum to one.}
\label{tab:traces-behaviour}
\end{table}

\paragraph{Citation composition.}
Almost every block names its sources, at $98.6\%$, and the corpus carries $9{,}430$ citations across $167$ domains. Official statistical sources account for $62.1\%$ of them, led by \texttt{census.gov} at $16.2\%$, \texttt{bea.gov} at $15.9\%$, \texttt{bls.gov} at $11.6\%$, \texttt{federalreserve.gov} at $6.1\%$, and \texttt{ismworld.org} at $5.3\%$. These are the sites that publish the release itself and the historical series behind it, which is what an agent needs in order to reconstruct a number. Consensus aggregators are a small share, with \texttt{tradingeconomics.com} at $3.5\%$ and \texttt{investing.com} at $1.5\%$. This speaks to the copying question of Appendix~\ref{app:contamination-check} from a second direction. The tests there are temporal and show that an agent's nowcast does not converge onto the consensus as the release approaches. The citation composition here shows that what the agent reads is overwhelmingly the primary release rather than a forecast aggregator. \texttt{reddit.com} accounts for $2.8\%$ of citations, which suggests the agent also reads community sentiment.

\paragraph{Checking the arithmetic.}
We pulled $652$ explicit calculations out of the prose and recomputed each one ourselves, allowing a $2\%$ tolerance because the agent states most of its results as rounded approximations. $93.4\%$ match the figure the agent reports. We then repeated the check with a second program written from scratch, which found $494$ calculations and agreed on $94.7\%$ of them. Of those, $399$ combine three or more numbers, so they are multi-step derivations rather than a single retrieved figure restated. To confirm the check can fail, we also ran it on three calculations we knew to be wrong, and it rejected all three.

\paragraph{Departures from the computed value.}
Of the $604$ blocks holding both a checked calculation and a stated forecast, the forecast differs from the calculation in $69.5\%$ of cases. Most departures give a reason. Adjustment language appears in $40.0\%$ of them, and language about the period ahead in $40.5\%$, where the calculation covers the prior period. The block below shows retrieval, calculation, and adjustment in sequence.

\begin{quote}
\footnotesize
``Calculate 2026 Q2 level: baseline Q1 level \$24,066.0 billion; growth 1.3\% SAAR $\rightarrow$ new level $=24{,}066.0 \times (1 + 0.013) \approx 24{,}379.0$. But given mixed signals (soft IP, high inflation), shave modestly to \$24,050.00 billion to reflect dampened momentum in latter half of quarter.''
\end{quote}

\section{Information Available to the Professional and Institutional Benchmarks}\label{app:benchmark-information}

This appendix documents the information behind each of the professional and institutional benchmarks introduced in Section~\ref{sec:comparators}, so that the fairness of the comparison against the LLM agents can be audited.

\paragraph{Bloomberg.}
The ECOS panel aggregates individual point forecasts submitted ahead of each release. Bloomberg describes the surveyed pool as a diverse group of traders, portfolio managers, think tanks, and academics, drawn from relationships with roughly 1,600 forecasters, and reports around 80 individual estimates on the most widely followed series \citep{bloomberg2024ecos}. The panel's edge is therefore whatever information those institutions can bring to bear, which may include paid data subscriptions, commercial data vendors, and non-public internal sources. We frame these as information potentially available to a submitting forecaster rather than as what any single estimate actually uses, since the survey publishes the contributor's name, firm, and estimate but not the inputs behind it.

\paragraph{Federal Reserve.}
The five regional-bank nowcasts we use are transparent about their inputs, and Table~\ref{tab:benchmark-inputs} summarizes them. Four of the five run on essentially publicly available government and statistical data, the same releases a web-search LLM agent can read on the open web. The only genuine data advantage is the Chicago Fed CHURN, which blends in proprietary alternative feeds.

\begin{table*}[tp]
\centering
\footnotesize
\setlength{\tabcolsep}{5pt}
\renewcommand{\arraystretch}{1.2}
\begin{tabular}{@{}p{0.19\linewidth}p{0.47\linewidth}c p{0.15\linewidth}@{}}
\toprule
\textbf{Fed nowcast (target)} & \textbf{Disclosed data inputs} & \textbf{Public inputs} & \textbf{Documentation} \\
\midrule
Atlanta Fed GDPNow (real GDP) & The public monthly source data the BEA itself uses, from the Census Bureau, the BLS, and other agencies, covering consumer and construction spending, international trade, and business inventories, mapped to GDP subcomponents through bridge equations. The raw data and model parameters are published in a downloadable spreadsheet at each release & Yes & \citet{atlfed_gdpnow} \\
New York Fed Staff Nowcast (real GDP) & A panel of public macroeconomic indicators in a dynamic factor model, with each incoming release entering the nowcast as news & Yes & \citet{nyfed_nowcast,okeeffe2025component} \\
St.~Louis Fed Real GDP Nowcast (real GDP) & An economic news index built from the content of key public monthly data releases & Yes & \citet{stlfed_realgdp_nowcast} \\
Cleveland Fed Inflation Nowcasting (CPI) & A small number of public series at mixed frequencies, namely daily oil prices, weekly gasoline prices, and monthly CPI and PCE readings, in a model-switching design & Yes & \citet{clevelandfed_inflation_nowcast,knotek2017nowcasting} \\
Chicago Fed CHURN (unemployment rate) & Public job-finding and separation rates derived from the BLS Current Population Survey, blended through partial least squares with the Bloomberg consensus forecast for the unemployment rate and with private high-frequency indicators from Google, Indeed, Morning Consult, and Haver & Partly & \citet{chicagofed_churn} \\
\bottomrule
\end{tabular}
\caption{Disclosed data inputs behind the five Federal Reserve regional-bank nowcasts used as comparators. Four of the five run on essentially publicly available government and statistical data, the same releases a web-search LLM agent can read. The Chicago Fed CHURN is the only one that adds proprietary feeds.}
\label{tab:benchmark-inputs}
\end{table*}

\section{Auto-ARIMA Baseline}\label{app:arima-baseline}

As an econometric baseline alongside the LLM agents, LiveMacroEval includes a univariate auto-ARIMA model fit per indicator, following the automatic ARIMA order selection approach \citep{hyndman2008automatic}. For each indicator $i$, we fit an $\mathrm{ARIMA}(p, d, q)$ specification on the indicator's own historical series and produce a single point forecast for the target reference period.

\paragraph{Model and order selection.}
The model is implemented in Python with the \texttt{statsmodels} ARIMA estimator. The differencing order $d$ is set from the canonical FRED transform code attached to each series in the FRED-MD and FRED-QD vintage files \citep{mccracken2016fredmd, mccracken2021fredqd}, which encodes the transformation under which the series is treated as stationary in the macro forecasting literature. In our 16-indicator panel $d \in \{0, 1, 2\}$, with most monthly growth-rate series at $d{=}1$ and headline inflation indices at $d{=}2$. With $d$ fixed by the transform code, the autoregressive and moving-average orders are selected jointly by AIC minimization \citep{hyndman2008automatic} over the grid $p \in \{0, 1, \ldots, 12\}$ and $q \in \{0, 1, 2, 3\}$. The trend term is fixed at a constant when $d{=}0$, a linear time trend when $d{=}1$, and is suppressed when $d{=}2$, matching the conventional treatment of integrated series in macroeconomic forecasting. Stationarity and invertibility constraints during likelihood maximization are not enforced, so the estimator can reach the unconstrained AIC-minimizing root configuration. The ISM Manufacturing PMI and ISM Services PMI series are treated separately: both are bounded diffusion indices for which the FRED transform code is not directly applicable, so we fit an AR(1) on the log level and forecast back through the exponential.

\paragraph{Data and refit cadence.}
For each target reference period, we use the FRED vintage immediately preceding the target month. The model is therefore refit at the start of each prediction window on the up-to-that-point historical sample, with a minimum of 60 observations required for a fit to be issued. The fitted model produces a single point forecast for the target month, and that forecast is held fixed for the entire prediction window. By construction the baseline does not condition on intra-window information releases, prediction-market prices, or news flow, providing a frozen reference against which the LLM agents' continuously updated nowcasts can be benchmarked. In the live-scoring pipeline of Section~\ref{sec:metric}, the single ARIMA point forecast is broadcast across all pre-release hourly timestamps for that target month, so the LiveMacro and LiveBetting scores collapse to a single horizon-invariant value for this baseline.

\section{Nowcast Error in Real Time}\label{app:nowcast-error}

We report each indicator's real-time nowcast squared relative error against the eventual official release. For indicator $i$, the relative error at time $t$ is $r_{i,t} = (\hat y_{i,t} - y_i)/s_i$, where $\hat y_{i,t}$ is the model prediction, $y_i$ is the official release, and $s_i$ is the mean absolute release value across the target months we evaluate. The plotted quantity is its square $r_{i,t}^2$. This scaling renders errors dimensionless and directly comparable across indicators with otherwise incommensurable units (e.g., levels in thousands versus percentage points). The Bloomberg consensus relative error is computed in the same way and plotted as a reference.\footnote{Bloomberg typically begins publishing its consensus only in the final week before a release. We take the first published economist estimate as the earliest available value and carry it backward to earlier dates, allowing a direct comparison between early model nowcasts and the first Bloomberg estimate.} We report this metric one curve at a time, per indicator, and not as a cross-indicator total. Appendix~\ref{app:conventional-aggregators} shows what goes wrong when it is summed into a single ranking, and why the LiveMacro Score weights by market impact instead.

\begin{figure*}[p]
\centering
\includegraphics[width=0.8\linewidth]{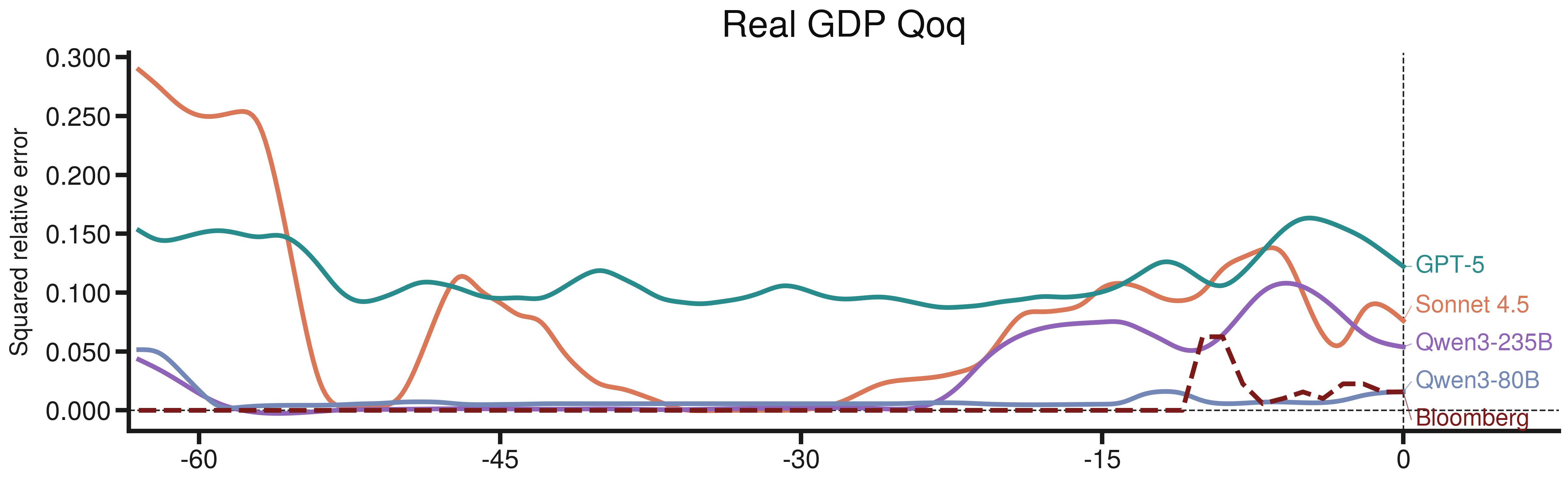}\\[4pt]
\includegraphics[width=0.8\linewidth]{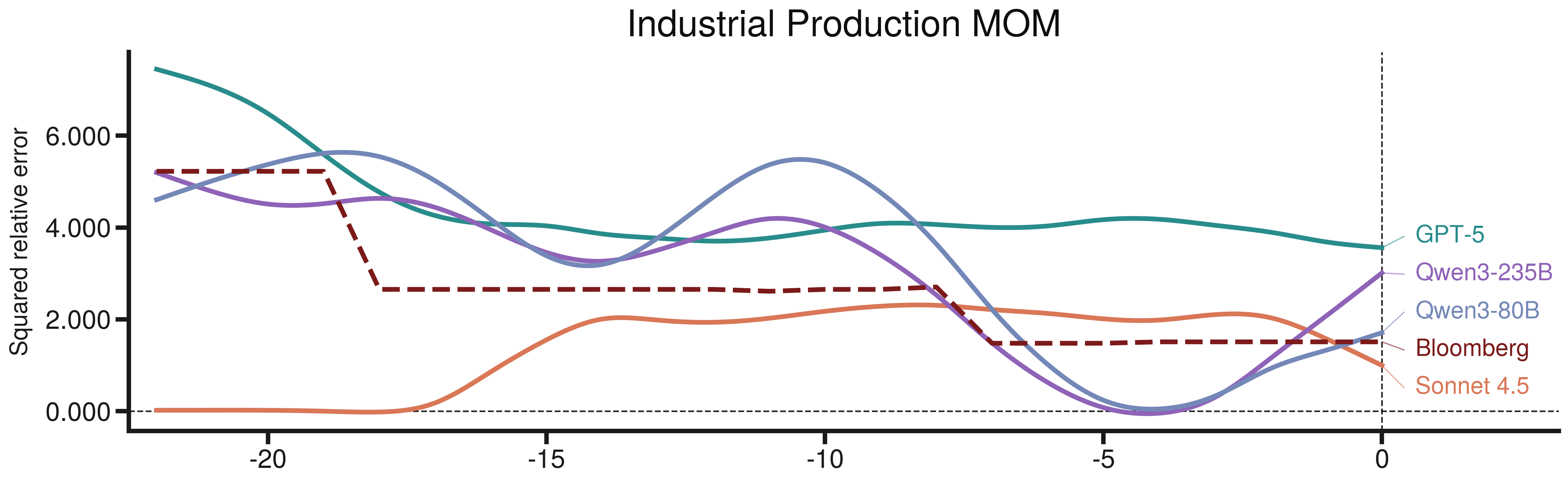}\\[4pt]
\includegraphics[width=0.8\linewidth]{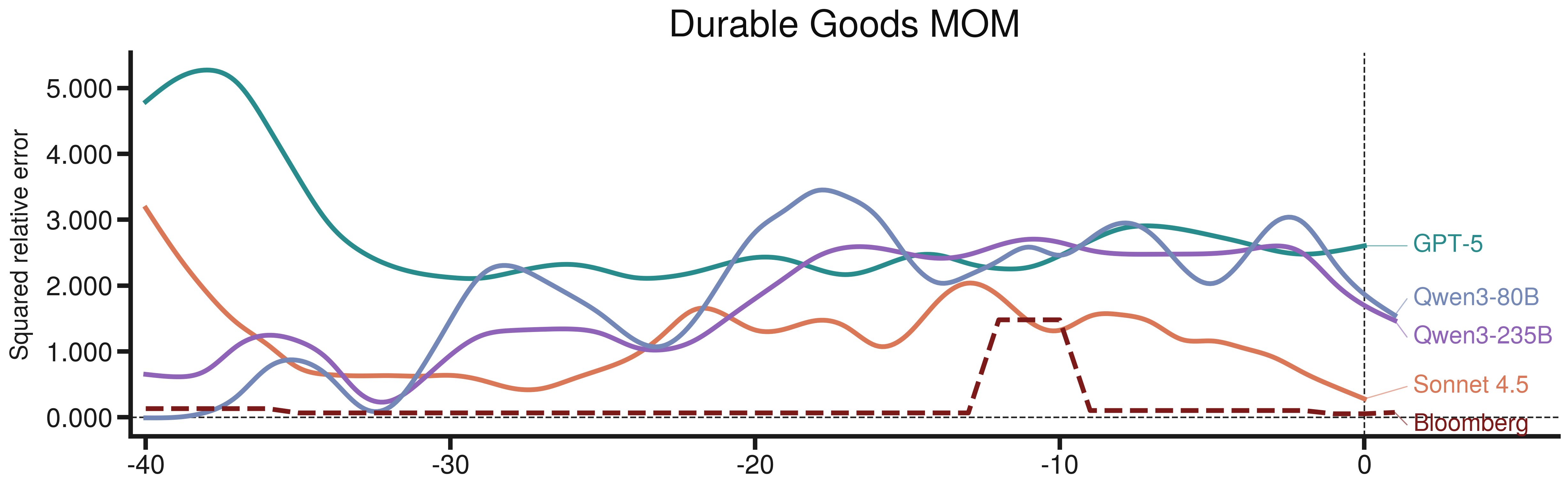}\\[4pt]
\includegraphics[width=0.8\linewidth]{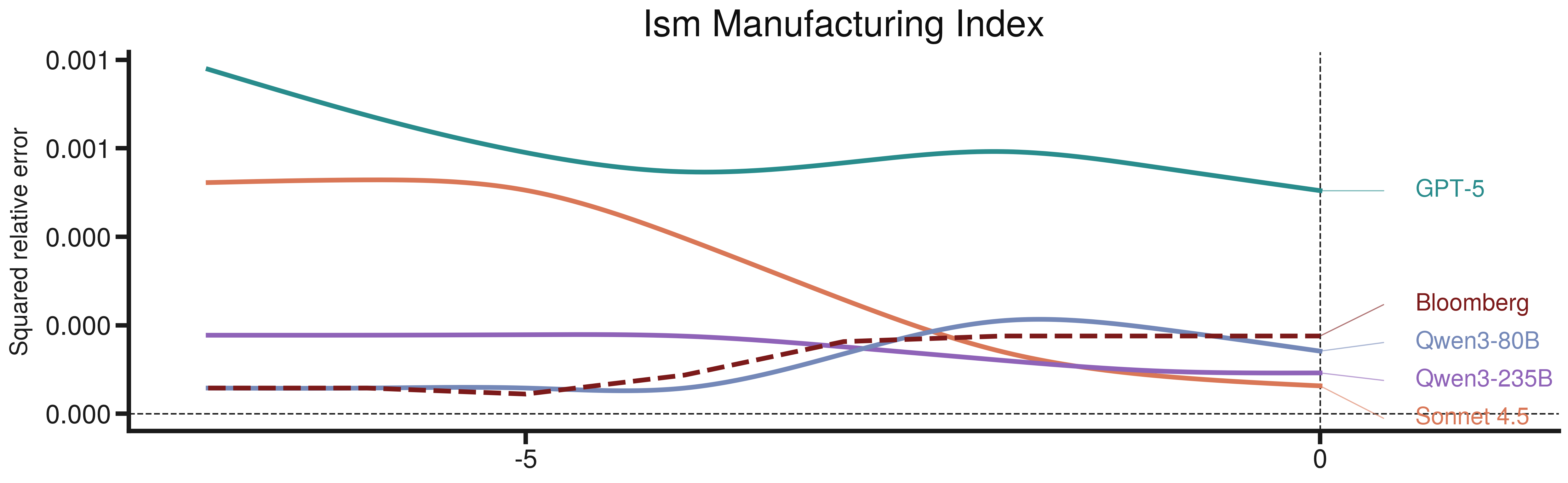}
\caption{Supply and Production: real-time nowcast error against the official release, by indicator and model, averaged across the February and March~2026 releases. Panels (top to bottom): real GDP (QoQ), industrial production (MoM), durable goods orders (MoM), ISM manufacturing index.}
\label{fig:err-supply-production}
\end{figure*}

\begin{figure*}[p]
\centering
\includegraphics[width=0.7\linewidth]{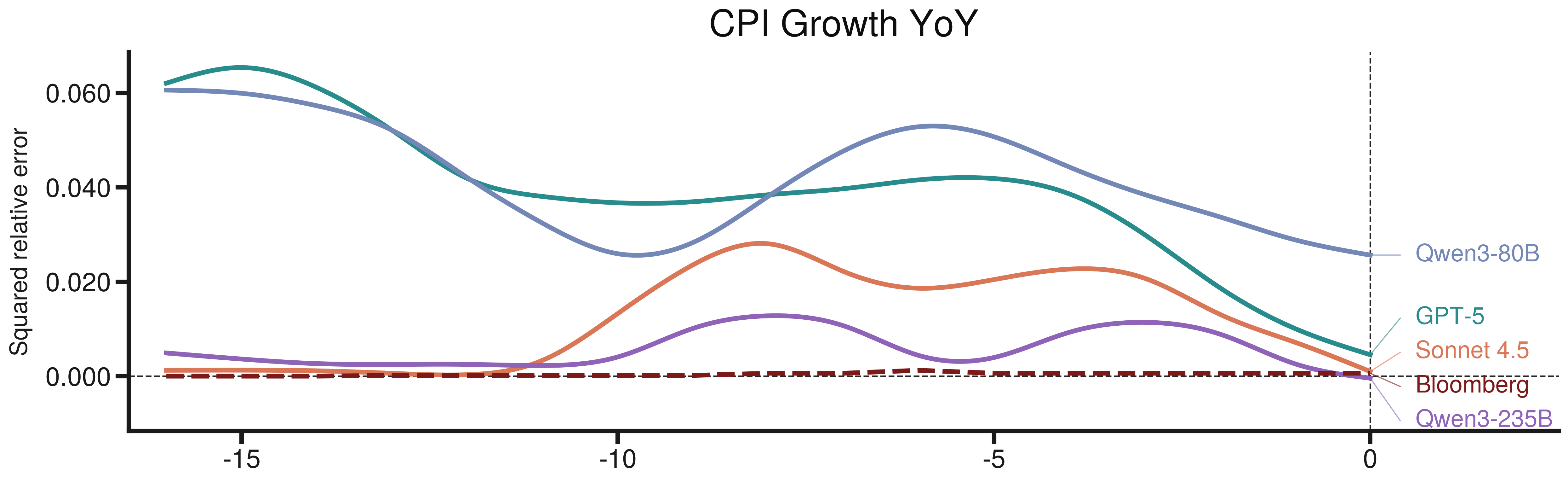}\\[4pt]
\includegraphics[width=0.7\linewidth]{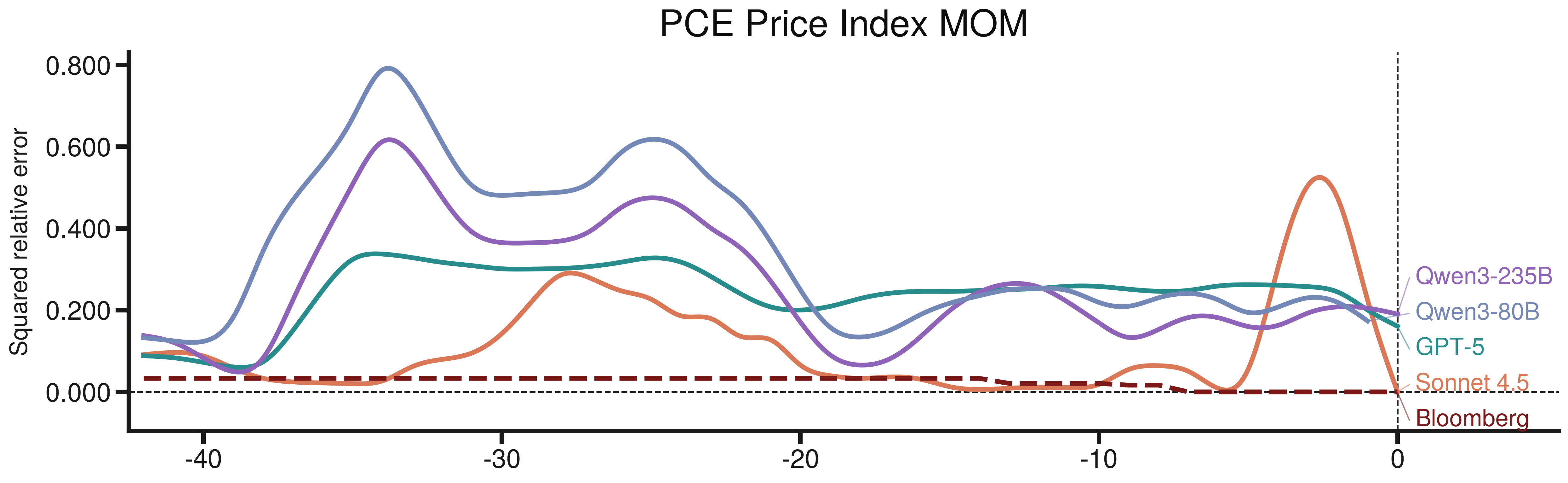}\\[4pt]
\includegraphics[width=0.7\linewidth]{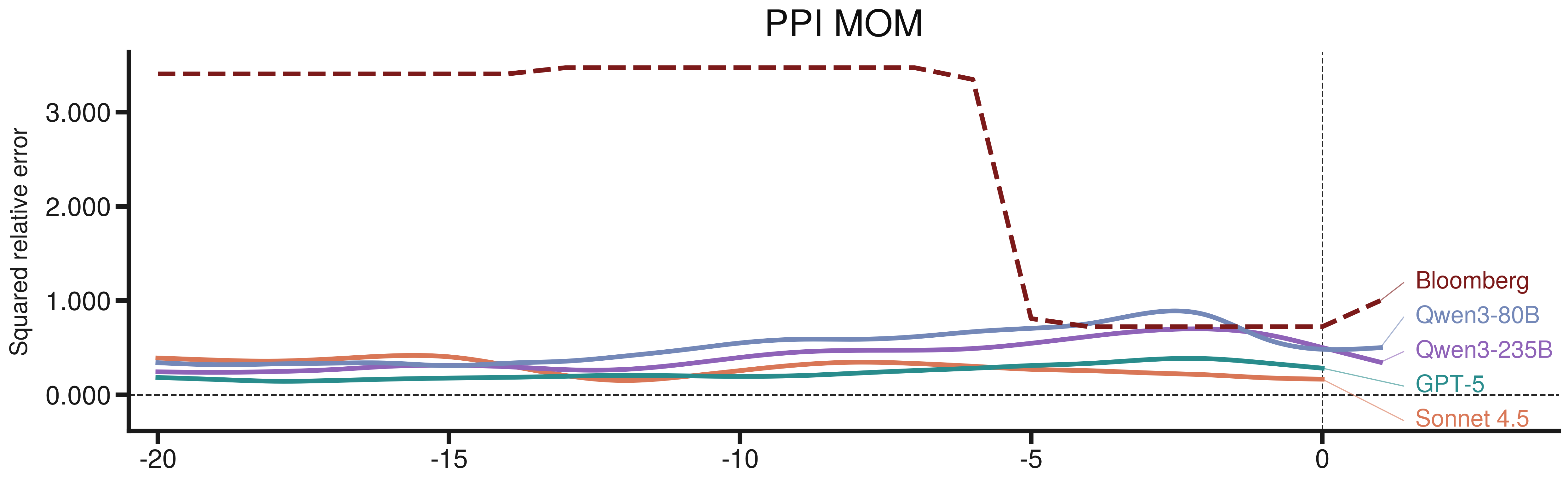}\\[4pt]
\includegraphics[width=0.7\linewidth]{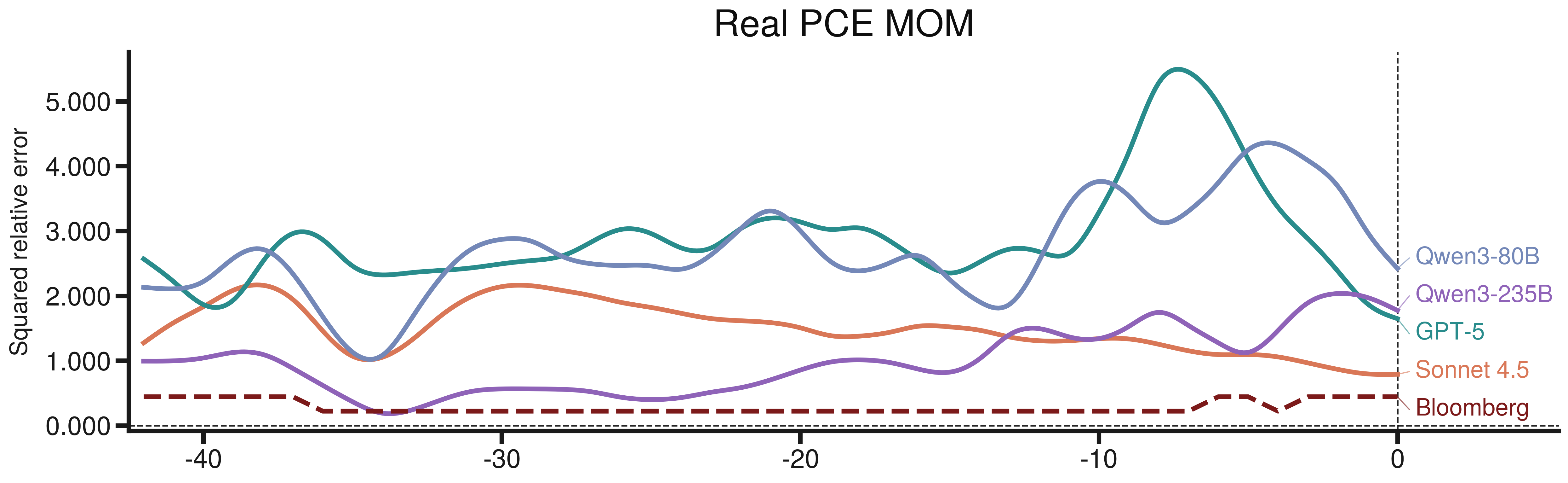}\\[4pt]
\includegraphics[width=0.7\linewidth]{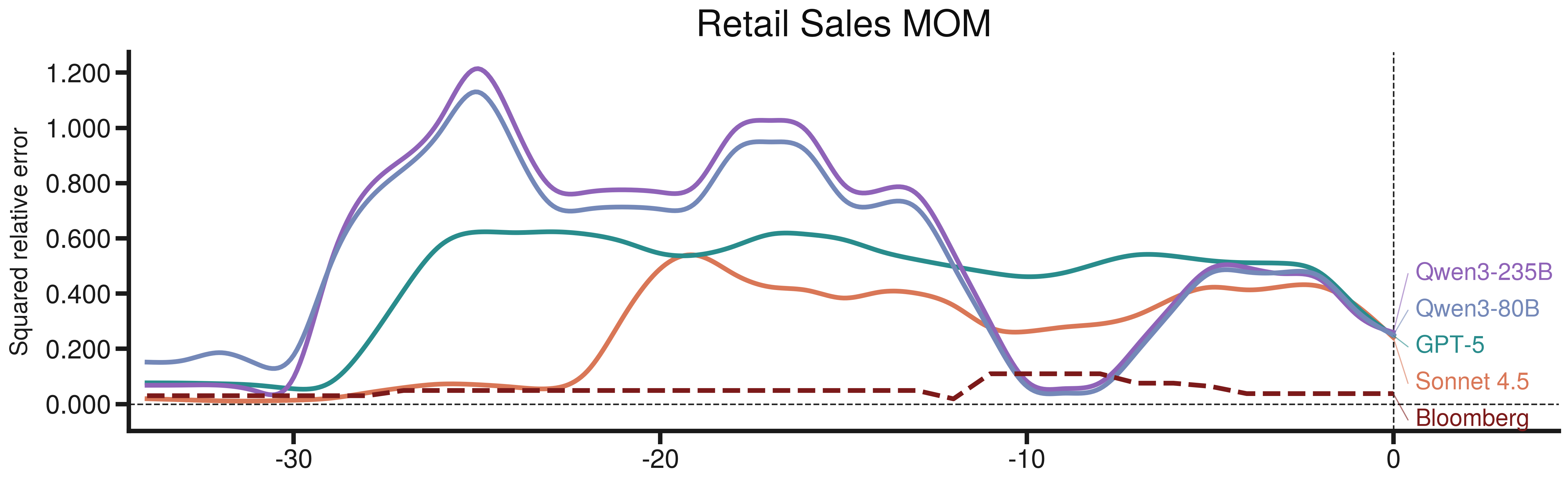}\\[4pt]
\includegraphics[width=0.7\linewidth]{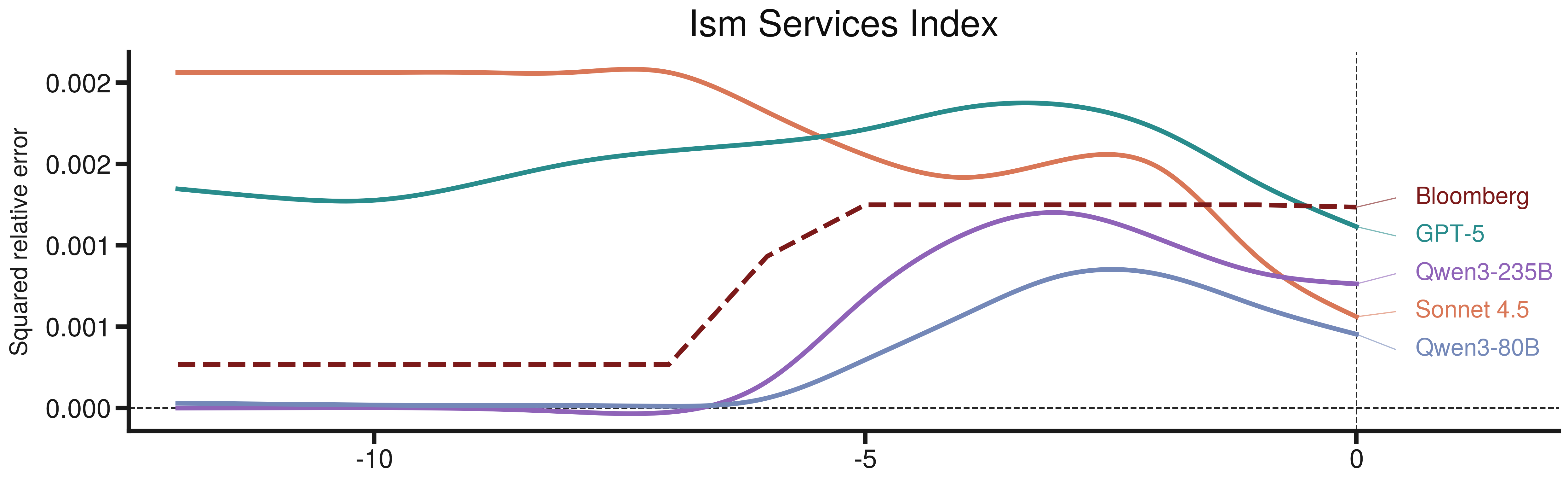}
\caption{Demand and Inflation: real-time nowcast error against the official release, by indicator and model, averaged across the February and March~2026 releases. Panels (top to bottom): CPI (YoY), PCE price index (MoM), PPI (MoM), real PCE (MoM), retail sales (MoM), ISM services index.}
\label{fig:err-demand-inflation}
\end{figure*}

\begin{figure*}[p]
\centering
\includegraphics[width=0.8\linewidth]{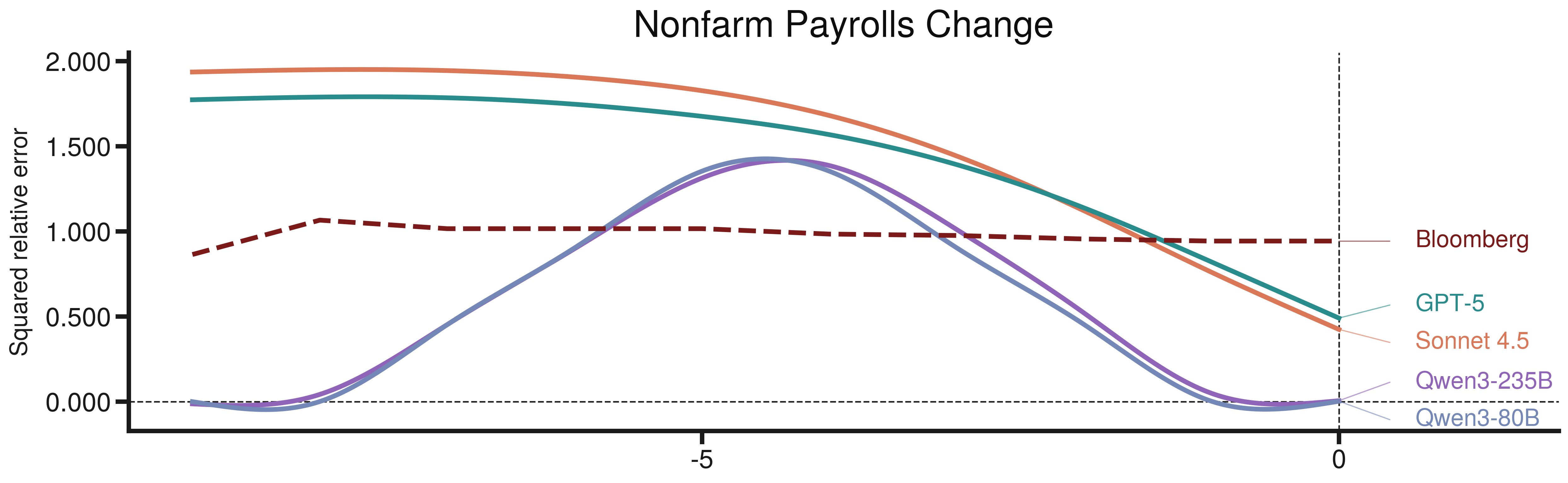}\\[4pt]
\includegraphics[width=0.8\linewidth]{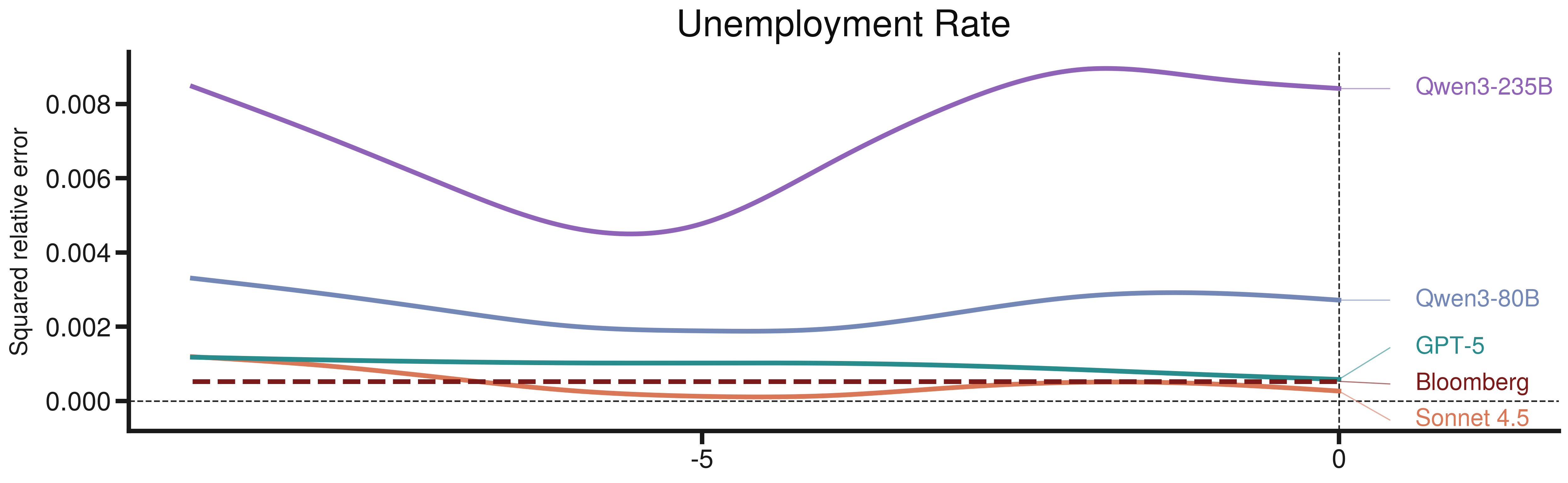}
\caption{Labor Market: real-time nowcast error against the official release, by indicator and model, averaged across the February and March~2026 releases. Panels (top to bottom): nonfarm payrolls (MoM change), unemployment rate.}
\label{fig:err-labor}
\end{figure*}

\begin{figure*}[p]
\centering
\includegraphics[width=0.8\linewidth]{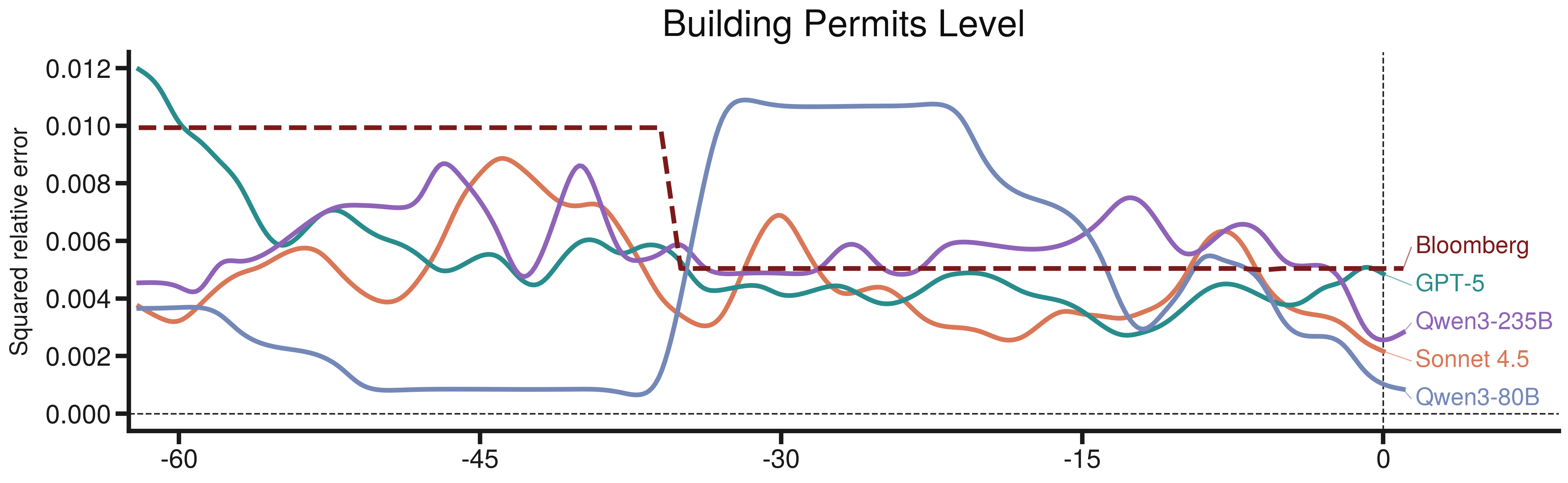}\\[4pt]
\includegraphics[width=0.8\linewidth]{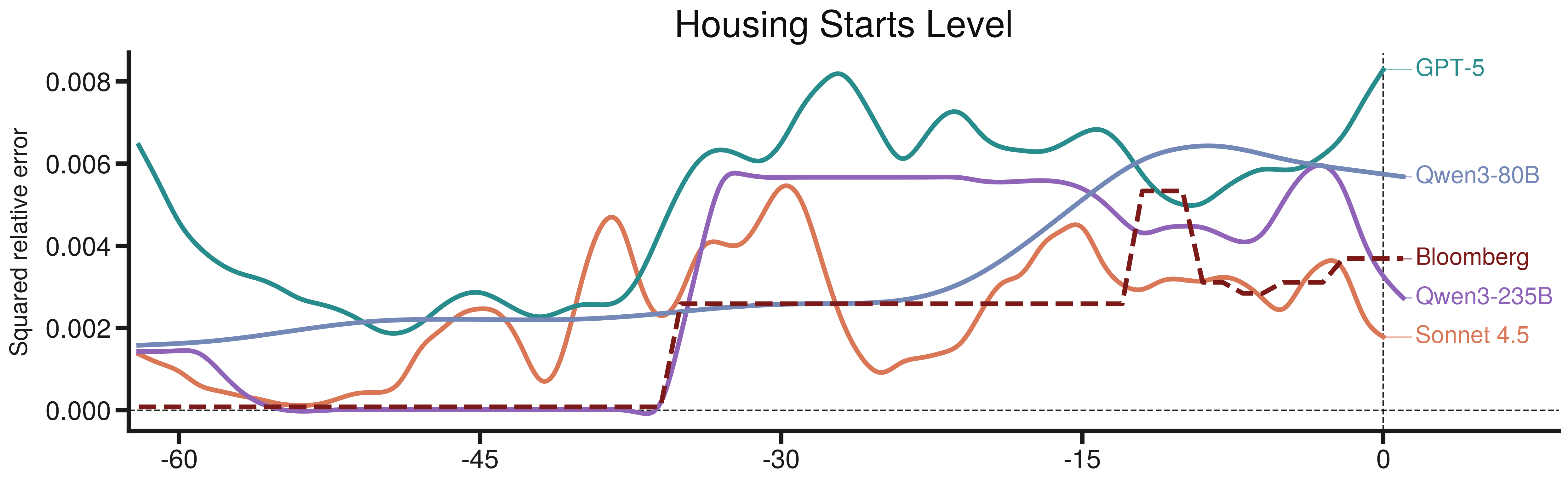}\\[4pt]
\includegraphics[width=0.8\linewidth]{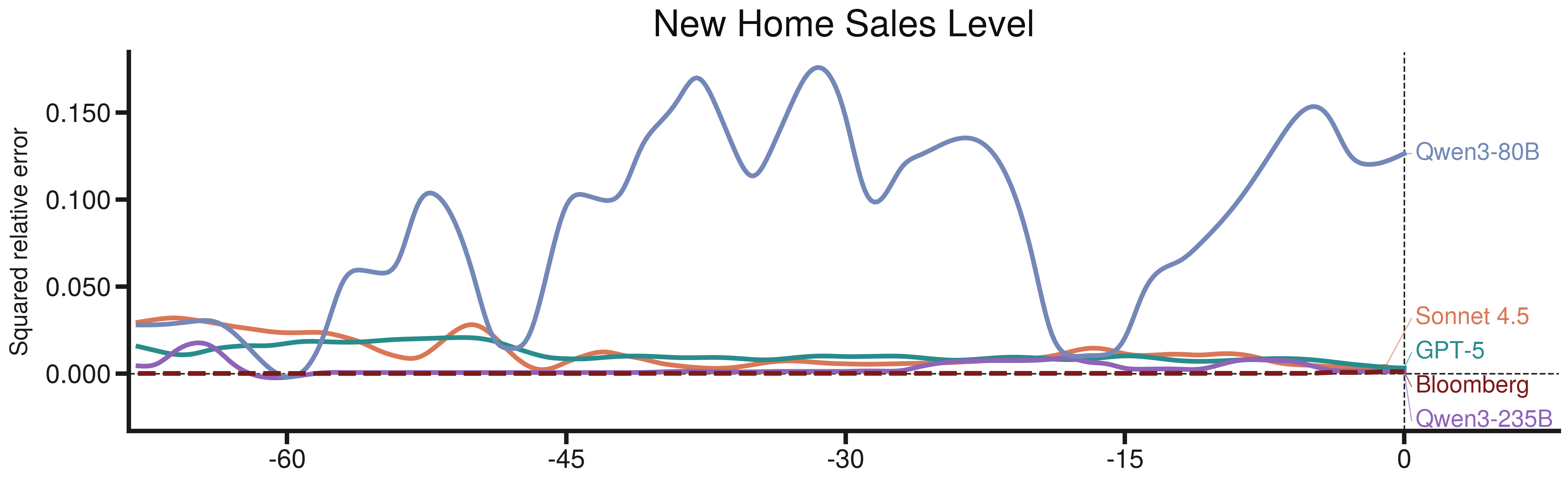}\\[4pt]
\includegraphics[width=0.8\linewidth]{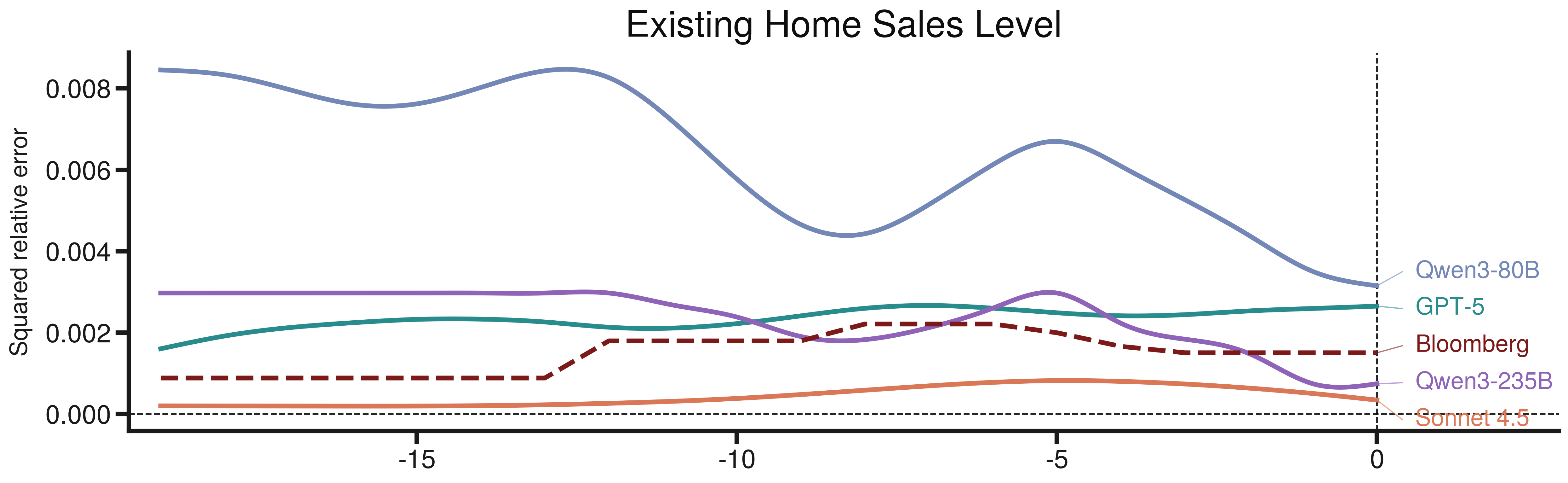}
\caption{Housing: real-time nowcast error against the official release, by indicator and model, averaged across the February and March~2026 releases. Panels (top to bottom): building permits (SAAR level), housing starts (SAAR level), new home sales (SAAR level), existing home sales (SAAR level).}
\label{fig:err-housing}
\end{figure*}

Figures~\ref{fig:err-supply-production} through \ref{fig:err-housing} report the model error and the Bloomberg consensus error for the February and March~2026 releases, organized by macroeconomic theme as defined in Section~\ref{sec:tasks-themes}. We treat the plotted target months as independent samples. At each integer day relative to the release, we first compute the mean squared relative error within each target month, then average across target months. Plotted curves are smoothed with a GCV-selected smoothing spline fit to the daily-binned, 3-day-averaged series.

For supply and production (Figure~\ref{fig:err-supply-production}), the consensus error on real GDP and durable goods orders is small from the outset and no model materially improves on it. Industrial production and the ISM manufacturing index begin with large model errors that decline sharply toward release, with Claude-sonnet and Qwen3 narrowly meeting or crossing consensus in the final days.

For demand and inflation (Figure~\ref{fig:err-demand-inflation}), CPI, the PCE price index, real PCE, and retail sales all show consensus errors close to zero, and no model improves on consensus. PPI is the clearest counter-example: consensus error is persistently high and every model falls below consensus near release. For the ISM services index, consensus error rises in the final days while model errors decline, so all models beat consensus at resolution. This case illustrates the value of continuous high-frequency updating, which the consensus does not exploit.

For the labor market (Figure~\ref{fig:err-labor}), every model beats consensus on nonfarm payrolls, where the consensus error remains large throughout the window. The unemployment rate is tightly tracked by consensus, and only Claude-sonnet and GPT-5 cross below the consensus level.

For housing (Figure~\ref{fig:err-housing}), the value of high-frequency updating is most visible. Consensus error remains flat in the days immediately before release, whereas model nowcasts continue to update and typically fall below consensus near resolution. Claude-sonnet performs consistently well across the four housing series, while GPT-5 and Qwen3 are weakest in this theme.

\clearpage

\clearpage

\clearpage

\clearpage

\clearpage

\section{LiveMacro Score: Full Construction}\label{app:capture-score-details}

This appendix gives the full construction of the LiveMacro score summarized in Section~\ref{sec:metric}.

\subsection{Setup and construction}

\paragraph{Defining events.}
Section~\ref{sec:metric} writes the LiveMacro score in a compact notation: each event carries one realized return $r$ and a vector of surprises across the 16 scored indicators. The score formula in this appendix is identical. The only detail we make precise is what constitutes an event in the actual data: not all 16 indicators release in isolation, and a few share a single official timestamp (e.g., nonfarm payrolls and the unemployment rate at 8:30 ET on the first Friday of each month), so a single announcement-window return $r_g$ responds to several concurrent surprises at once. To handle this jointly \citep{gurkaynak2020missing}, we group all releases sharing a timestamp into a single \emph{timestamp-group event} $g \in \mathcal{G}$, indexed by its release timestamp $T_g$. Each event $g$ yields one return $r_g$ and a 16-dimensional standardized-surprise vector, with structural zeros for fields not released in $g$. This grouping motivates the $g$ subscript used throughout the rest of this appendix.

\paragraph{Standardized surprises and historical regression.}
For each scored field $i \in \mathcal{I}$ and release $t$, let $X_{i,t}$ denote the released value, $X^c_{i,t}$ the prevailing market consensus, $T_{i,t}$ the realized release timestamp, and $\hat X_{i,t}$ the latest model nowcast strictly before $T_{i,t}$.\footnote{Alternative definitions of $\hat X_{i,t}$, such as the median nowcast over a longer pre-release window (e.g., the last three days), can be used when evaluating model performance at different horizons. We use the latest pre-release nowcast here because the market consensus is itself collected during the final days before release and finalized immediately beforehand, so this choice aligns the model's information set with the consensus's.} Following \citet{scotti2016surprise}, we standardize both the realized and the model-implied surprises by the field-specific historical surprise scale $\sigma_i = \mathrm{sd}(X_{i,t} - X^c_{i,t})$:
\begin{equation}
S_{i,t} = \frac{X_{i,t} - X^c_{i,t}}{\sigma_i}, \qquad \hat S_{i,t} = \frac{\hat X_{i,t} - X^c_{i,t}}{\sigma_i}.
\end{equation}
For each field $i$ released in event $g$ at $t = T_g$, we set $S_{i,g} = S_{i,t}$ and $\hat S_{i,g} = \hat S_{i,t}$; for fields not released in $g$, $S_{i,g} = \hat S_{i,g} = 0$. The market reaction in event $g$ is the high-frequency log return on front-month E-mini S\&P~500 futures over a $[-5, +30]$ minute window around $T_g$ \citep{andersen2007realtime}:
\begin{equation}
r_g = \log P^{ES}(T_g + 30\text{m}) - \log P^{ES}(T_g - 5\text{m}).
\end{equation}
The mapping from standardized surprises to event-window returns is estimated on a long historical sample $\mathcal{G}^{\mathrm{hist}}$ spanning Jan 2010 to Oct 2025 via the pooled Huber-ridge regression
\begin{equation}\label{eq:hist-reg}
r_g = \sum_{i \in \mathcal{I}} \beta_i S_{i,g} + u_g, \quad g \in \mathcal{G}^{\mathrm{hist}},
\end{equation}
fitted without a constant, following the high-frequency event-study convention of \citet{gurkaynak2020missing}: under the identifying assumption that the announcement-window return responds only to the released surprise, a zero surprise carries no expected return. The fit yields the field-specific surprise sensitivities $\{\beta_i\}_{i \in \mathcal{I}}$, frozen during evaluation (estimator details in the next subsection, the estimated $\hat\beta_i$ are reported in Table~\ref{tab:hist-beta}).

\paragraph{Predicted return and LiveMacro Score.}
The model's predicted stock-equivalent shock for each live event is
\begin{equation}
\hat Q_g = \sum_{i \in \mathcal{I}} \beta_i \hat S_{i,g},
\end{equation}
and the model's and consensus's predicted announcement-window returns are
\begin{equation}\label{eq:r-hat}
\hat r_g = \hat Q_g, \qquad r^{c}_g = 0,
\end{equation}
where the consensus has zero model-implied surprise ($\hat Q^c_g \equiv 0$), so its predicted return is identically zero under the no-intercept design of Equation~\ref{eq:hist-reg}. The Bloomberg consensus is the strongest publicly available pre-release expectation for each indicator, so the LiveMacro Score measures whether the model's surprises carry information about the realized announcement-window market response above and beyond what the consensus already prices in. On the live evaluation sample $\mathcal{G}^{\mathrm{live}}$, the score compares $\hat r_g$ and $r^{c}_g$ against the realized return $r_g$:
\begin{equation}
\begin{aligned}
\mathrm{LiveMacro}(r, \hat r, r^c) &= \frac{\mathrm{SS}_{\mathrm{cons}} - \mathrm{SS}_{\mathrm{model}}}{\mathrm{SS}_{\mathrm{cons}} + \mathrm{SS}_{\mathrm{model}}}, \\
\mathrm{SS}_{\mathrm{model}} &= \sum_{g} \big(r_g - \hat r_g\big)^2, \\
\mathrm{SS}_{\mathrm{cons}} &= \sum_{g} \big(r_g - r^{c}_g\big)^2,
\end{aligned}
\end{equation}
where the sums run over $g \in \mathcal{G}^{\mathrm{live}}$. Because $\hat r_g = \hat Q_g$ and $r^{c}_g = 0$ under the no-intercept design, this matches the main paper's $(r - \hat Q)^2$ versus $(r - Q^c)^2$ comparison from Section~\ref{sec:metric} exactly. The score is bounded by construction in $[-1, +1]$: $+1$ denotes a model whose nowcasts perfectly anticipate the realized announcement-window market returns; $0$ denotes performance that exactly matches the consensus reference (no equities-relevant signal beyond what consensus already prices in); and values approaching $-1$ indicate predictions arbitrarily worse than the consensus.

\paragraph{Economic interpretation.}
Economically, the score quantifies how much of the realized announcement-window equity-market response is anticipated by the model's nowcasts above and beyond what is already priced in by the consensus. Because the field-specific weights $\{\beta_i\}$ are estimated from the long-run surprise-to-return mapping of each macro variable, the score automatically rewards skill on releases that historically move the market most (e.g., CPI, nonfarm payrolls, GDP) and discounts skill on smaller-impact releases, yielding a single equities-relevant ranking of macro-nowcasting models that respects the economic importance of each variable.

\paragraph{Theme decomposition.}
The score is free to define on any subset of indicators. For any theme $\theta$ with indicator set $\mathcal{I}^\theta \subseteq \mathcal{I}$, restricting the event sum in $\mathrm{SS}_{\mathrm{model}}$ and $\mathrm{SS}_{\mathrm{cons}}$ to releases of indicators in $\mathcal{I}^\theta$ yields a theme-specific LiveMacro score. We use this to report per-theme scores for the four macroeconomic blocks (supply and production, demand and inflation, labor market, housing), which identify the segments in which each LLM agent has the greatest predictive value relative to the institutional and professional consensus nowcasts.

\subsection{Estimation}

\paragraph{Design of the market-reaction variable.}
The announcement-window equity return is the most widely watched economic-utility measure of a macro release among market participants and economists, motivating our choice of $r_g$ as the response variable for the LiveMacro score. Two properties justify it. First, it is a directly priceable cash quantity that prices each release at the weight the market itself assigns to it, and among such priceable measures it is the canonical metric in the macro-announcement literature \citep{andersen2003micro,andersen2007realtime,boyd2005stock}. Second, estimating $\{\beta_i\}$ on tight intraday windows around pre-scheduled releases removes confounding from contemporaneous news, so $\{\beta_i\}$ admits a causal interpretation and matches the standard target of high-frequency event-study identification in empirical macroeconomics and finance \citep{gurkaynak2020missing,nakamura2018high,bauer2023alternative}. We use a $[-5, +30]$ minute window: it is short enough that little other major news is expected to arrive, and its timing is known in advance, so the measured return is attributable to the released surprise alone \citep{kuttner2001monetary,andersen2003micro}.

We measure $r_g$ on the front-month E-mini S\&P~500 futures contract (ticker ES, CME Globex), the canonical equity-market shock proxy in the high-frequency macro-announcement and monetary-policy identification literature \citep{andersen2007realtime,gurkaynak2020missing,nakamura2018high,bauer2023alternative}. ES is the most liquid equity-index instrument worldwide and trades on a near-twenty-four-hour electronic schedule across weekdays, so it produces a quoted price at every release timestamp in our panel, including the 8:30 ET slot in which CPI, nonfarm payrolls, advance retail sales, and the GDP advance estimate are released, well before the NYSE cash session opens. As a broad-index futures contract with the deepest book among equity instruments, its high-frequency quotes carry far lower microstructure noise than the cash S\&P~500, which is computed from non-synchronously trading constituents, or any individual stock, and the front-month tenor concentrates the bulk of that liquidity.

\paragraph{Historical regression sample.}
The LiveMacro score uses a frozen historical event sample. We estimate the surprise scales $\sigma_i$ and the equity-market sensitivity vector $\{\beta_i\}_{i \in \mathcal{I}}$ using first-release macro surprises from January~2010 through October~2025. After grouping simultaneous releases by timestamp, the historical regression sample contains 2{,}163 timestamp-group events across the 16 scored fields. The fitted $\beta_i$ are fixed before any live outcomes are scored.

\paragraph{Joint identification of $\{\beta_i\}$.}
Equation~\ref{eq:hist-reg} writes each event-level return $r_g$ against the full 16-dimensional surprise row $S_g = (S_{1,g}, \ldots, S_{16,g})^\top$, with $S_{i,g} = 0$ for fields not released in $g$. Identification of each $\beta_i$ then comes from cross-event variation in $S_{i,g}$, with the joint specification purging the partial co-movement induced by sibling releases on the same timestamp. A separate per-field regression would omit the sibling surprises and incur omitted-variable bias proportional to their covariance with $S_{i,g}$ \citep{gurkaynak2020missing}.

\paragraph{Huber-ridge estimator.}
Two features of the historical event sample motivate moving away from plain OLS. First, the design matrix is sparse: most timestamp-group rows have a single nonzero entry, so individual $\beta_i$ are weakly identified and OLS slopes are noisy. Second, announcement-window equity returns have heavy tails, with a small fraction of events (notably the March 2020 dislocation and a handful of high-volatility CPI and nonfarm-payrolls releases) producing residuals an order of magnitude larger than the typical event. To address both we estimate $\beta$ by minimizing the penalized robust objective
\begin{equation}
\hat\beta = \operatorname*{arg\,min}_{\beta} \; \sum_{g \in \mathcal{G}^{\mathrm{hist}}} \rho_\delta\!\left( r_g - S_g^\top \beta \right) + \lambda \, \|\beta\|_2^2,
\end{equation}
where $\rho_\delta$ is the Huber loss \citep{huber1964robust}
\begin{equation}
\rho_\delta(u) = \begin{cases} \tfrac{1}{2} u^2, & |u| \le \delta, \\ \delta\,|u| - \tfrac{1}{2}\delta^2, & |u| > \delta, \end{cases}
\end{equation}
and the L2 penalty on $\beta$ provides regularization in the presence of sparse columns and near-collinear surprises \citep{hoerl1970ridge}. The tuning constant is $c = 1.345$, the standard choice that yields approximately 95\% asymptotic efficiency under Gaussian errors, and the threshold $\delta$ is adapted from a rescaled median absolute deviation of the working residuals, $\delta = c \cdot 1.4826 \cdot \mathrm{median}_g |u_g - \mathrm{median}_g u_g|$, recomputed each IRLS step. The ridge penalty is fixed at $\lambda = 21.54$, selected by blocked chronological five-fold cross-validation under this no-intercept Huber-ridge specification on the historical $(S, r)$ over a log-spaced grid of 25 values from $10^{-4}$ to $10^{4}$, and frozen thereafter. We solve the M-estimator by iteratively reweighted least squares (IRLS) initialized at the OLS-ridge solution, with Huber weights $w_g = 1$ for $|u_g| \le \delta$ and $w_g = \delta / |u_g|$ otherwise. On the full historical sample the fit converges in 23 iterations at a relative-$\beta$ tolerance of $10^{-7}$, with converged Huber scale $\hat\sigma = 1.85 \times 10^{-3}$ and threshold $\delta = 2.49 \times 10^{-3}$ (in log-return units). Bit-identical $\hat\beta$ is recovered across runs because IRLS is deterministic given $(S, r, \lambda, c)$.

\paragraph{Heteroskedasticity-robust covariance of $\hat\beta$.}
The covariance of $\hat\beta$ accounts for both the IRLS weighting and the heteroskedasticity-robust adjustment of \citet{mackinnon1985some}. Let $X$ denote the $n \times p$ surprise design matrix used in Equation~\ref{eq:hist-reg}, let $W = \mathrm{diag}(w_g)$ collect the converged Huber weights, and let $\psi_\delta(u) = \mathrm{clip}(u, -\delta, +\delta)$ denote the Huber influence function. The HC1-style robust covariance estimator is
\begin{equation}
\begin{aligned}
V_\beta &= \frac{n}{n - \mathrm{df}_{\mathrm{eff}}} \, A^{-1} B A^{-1}, \\
A &= X^\top W X + \lambda I, \\
B &= X^\top \mathrm{diag}\!\left(\psi_\delta(u_g)^2\right) X,
\end{aligned}
\end{equation}
where $\mathrm{df}_{\mathrm{eff}} = \mathrm{tr}(X A^{-1} X^\top W)$ is the effective degrees-of-freedom adjustment for the penalized weighted fit ($\mathrm{df}_{\mathrm{eff}} \approx 13.4$ on the full sample, against the nominal $p = 16$ parameters). Marginal standard errors $\sqrt{\mathrm{diag}(V_\beta)}$ are reported per field in Table~\ref{tab:hist-beta}. The full $V_\beta$ is retained for the bootstrap step below.

\paragraph{Estimated surprise-to-return sensitivities.}
Table~\ref{tab:hist-beta} reports the field-level $\hat\beta_i$ and HC1 marginal standard errors from the historical Huber-ridge fit (Equation~\ref{eq:hist-reg}) over the 2{,}163 timestamp-group events in $\mathcal{G}^{\mathrm{hist}}$. Coefficients are in units of $10^{-4}$, so $\hat\beta_i$ is approximately the basis-point response of the announcement-window E-mini return to a one-standard-deviation surprise in field $i$. Signs align with standard high-frequency event-study evidence \citep{andersen2007realtime,gurkaynak2020missing}: positive activity surprises (real GDP, ISM manufacturing, nonfarm payrolls, retail sales, new home sales) lift equity returns, while positive inflation surprises (CPI, PPI, PCE price index) and a positive unemployment surprise depress them. CPI carries the largest absolute sensitivity at $-11.24 \times 10^{-4}$ per standardized surprise, followed by ISM manufacturing at $+10.38 \times 10^{-4}$, consistent with their prominence as the most market-moving releases in the panel.

This structure is what makes freezing $\{\beta_i\}$ defensible. The coefficients trace the cash-flow and discount-rate response of equities to a macro surprise rather than a transient correlation, and two decades of macro-announcement studies find the same sign and magnitude pattern, with inflation surprises depressing equity returns and positive activity surprises lifting them \citep{flannery2002macroeconomic,savor2013how}. The magnitudes are stable over comparable spans. A one-standard-deviation nonfarm-payroll surprise moved the two-year Treasury yield by roughly $6$ basis points over 1991 to 1995 \citep{balduzzi2001economic} and by $4.95$ basis points over 2004 to 2018 \citep{benamar2021demand}, the same sign and order of magnitude three decades apart.

\begin{table*}[tp]
\centering
\footnotesize
\setlength{\tabcolsep}{8pt}
\renewcommand{\arraystretch}{1.15}
\begin{tabular}{@{}lrr@{}}
\toprule
\textbf{Indicator} & $\hat\beta_i$ ($\times 10^{-4}$) & SE ($\times 10^{-4}$) \\
\midrule
\multicolumn{3}{@{}l}{\textit{Supply and Production}} \\
\addlinespace[2pt]
Real GDP (QoQ \%)               & $+5.567$ & $1.241$ \\
Industrial Production (MoM \%)  & $+0.567$ & $1.161$ \\
Durable Goods Orders (MoM \%)   & $+1.099$ & $0.920$ \\
ISM Manufacturing PMI           & $+10.382$ & $1.525$ \\
\midrule
\multicolumn{3}{@{}l}{\textit{Demand and Inflation}} \\
\addlinespace[2pt]
CPI: All Items (YoY \%)         & $-11.240$ & $1.594$ \\
PPI: Final Demand (MoM \%)      & $-2.258$ & $1.074$ \\
PCE Price Index (MoM \%)        & $-3.372$ & $1.436$ \\
Real PCE (MoM \%)               & $+2.581$ & $1.538$ \\
Retail Sales (MoM \%)           & $+6.095$ & $1.911$ \\
ISM Services PMI                & $+2.441$ & $1.487$ \\
\midrule
\multicolumn{3}{@{}l}{\textit{Labor Market}} \\
\addlinespace[2pt]
Nonfarm Payrolls (MoM change)   & $+4.937$ & $1.544$ \\
Unemployment Rate               & $-1.473$ & $1.814$ \\
\midrule
\multicolumn{3}{@{}l}{\textit{Housing}} \\
\addlinespace[2pt]
Housing Starts (SAAR)           & $+1.984$ & $0.827$ \\
Building Permits (SAAR)         & $-1.501$ & $0.970$ \\
Existing Home Sales (SAAR)      & $+0.470$ & $1.211$ \\
New Home Sales (SAAR)           & $+4.063$ & $1.234$ \\
\bottomrule
\end{tabular}
\caption{Estimated equity-market sensitivities $\hat\beta_i$ from the historical Huber-ridge regression in Equation~\ref{eq:hist-reg}, with HC1 heteroskedasticity-robust marginal standard errors (SE). Sample: 2{,}163 timestamp-group events from January 2010 through October 2025. Coefficients give the announcement-window log-return response of the front-month E-mini S\&P~500 futures contract to a one-standard-deviation surprise in field $i$, in units of $10^{-4}$ (approximately basis points).}
\label{tab:hist-beta}
\end{table*}

\subsection{Inference and robustness}

\paragraph{Parametric-bootstrap confidence intervals.}
We propagate the historical-regression uncertainty in $\hat\beta$ into the live-sample LiveMacro score by parametric bootstrap \citep{efron1993introduction}. Let $L$ be the Cholesky factor of $V_\beta$, so that $L L^\top = V_\beta$. For $b = 1, \ldots, B$ with $B = 10{,}000$ we draw
\begin{equation}
\beta^{(b)} = \hat\beta + L \, z^{(b)}, \quad z^{(b)} \overset{\mathrm{iid}}{\sim} \mathcal{N}(0, I_p),
\end{equation}
hold $S$, $\hat S$, and $r$ on the live sample fixed across draws, and recompute the LiveMacro score and its companions on each draw. Reported 90/95\% intervals are percentile intervals of the resulting empirical distribution. The one-sided $p$-values are the corresponding empirical tail probabilities. Theme-level intervals (Appendix~\ref{app:livemacro-theme-results-details}) use the same draws sliced to the relevant field indices, with $V_\beta$ restricted to the theme block and re-factored. 

\paragraph{Sampling-based confidence intervals.}
We also report a sampling-based confidence interval. We take the model's nearest pre-release nowcasts as samples, recompute the headline LiveMacro Score with each sample in place of the terminal nowcast (holding $\hat\beta$ and the Bloomberg final consensus fixed), and report the resulting mean and Student-$t$ $95\%$ interval across the samples.

\begin{figure}[htbp]
\centering
\includegraphics[width=0.95\linewidth]{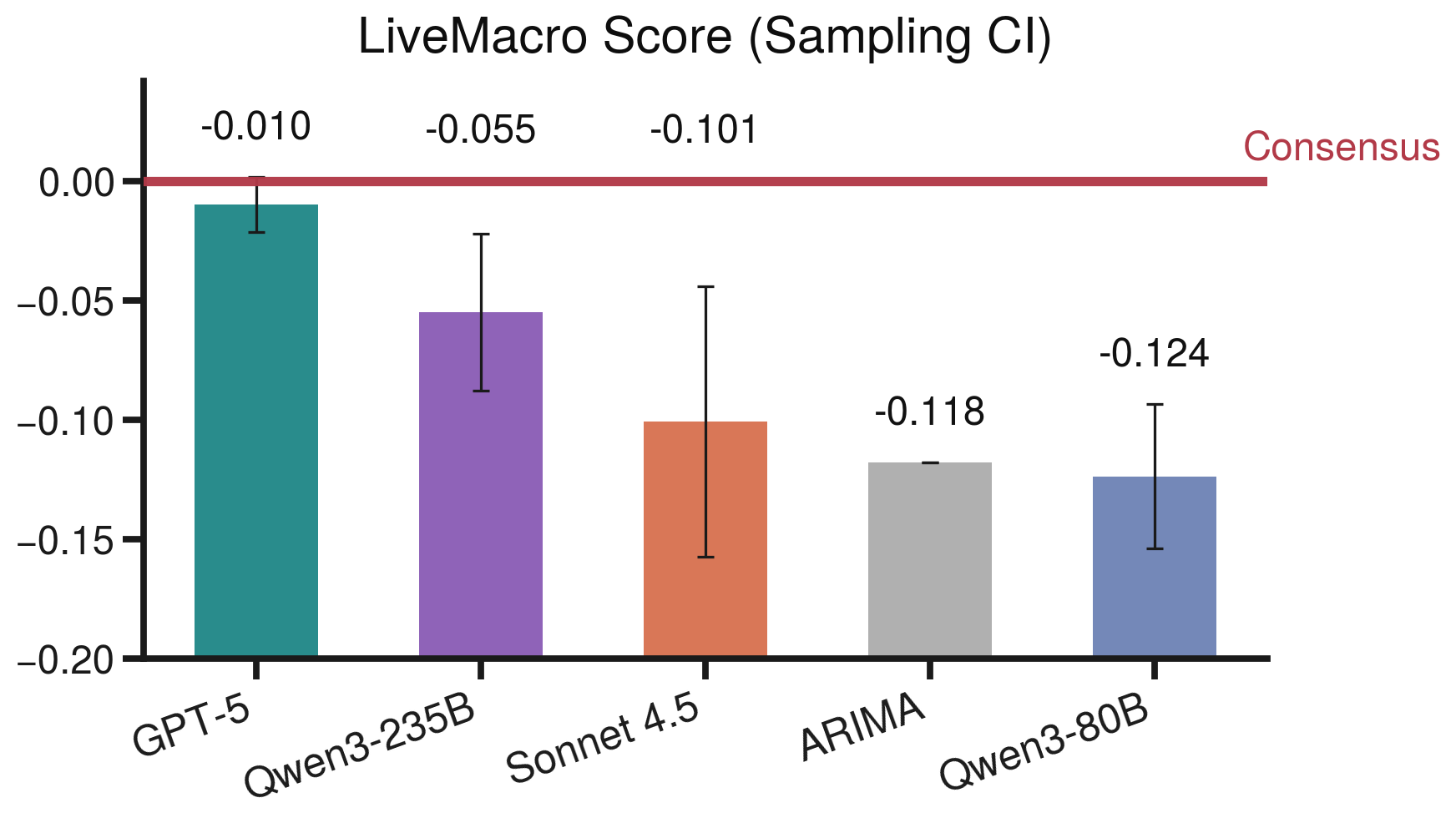}
\caption{LiveMacro Score with sampling-based confidence intervals. Bars are sample means and whiskers are Student-$t$ $95\%$ CIs, using the 5 most recent pre-release nowcasts per (field, event) as samples. The horizontal line is the Bloomberg-consensus reference at zero.}
\label{fig:sampling-livemacro-no-agent}
\end{figure}

Figure~\ref{fig:sampling-livemacro-no-agent} reports the result on the LiveMacro Score. GPT-5 leads the panel at $-0.010$, with the upper bound of its $95\%$ CI just crossing the Bloomberg-consensus reference line at zero, so it remains statistically indistinguishable from the consensus. The remaining models (Qwen3-235B at $-0.055$, Claude-sonnet-4.5 at $-0.101$, auto-ARIMA at $-0.119$, and Qwen3-80B at $-0.124$) retain the headline ranking of Section~\ref{sec:results-livemacro}. The sampling CI propagates near-release prediction-snapshot variability while holding $\hat\beta$ fixed, pinning down prediction-emission stability within the live window.

The point estimates under the sampling CI are uniformly somewhat lower than under the headline parametric-bootstrap CI of Section~\ref{sec:results-livemacro} (e.g., GPT-5 moves from $+0.004$ to $-0.010$). This gap reflects a structural disadvantage faced by the LLM agents under this construction rather than a deterioration in nowcast quality. The Bloomberg consensus on which the sampling CI conditions is the \emph{terminal} consensus median, which continues to be updated up to the announcement and incorporates economist submissions made within one to two hours of the official release \citep{kurov2019price}. By contrast, the agent samples are drawn at the hourly cadence of the nowcasting pipeline, so the five most recent pre-release samples reach up to roughly six hours before release. They necessarily condition on a strictly earlier information set than the consensus they are scored against. The like-for-like alignment of information sets achieved by the headline terminal-snapshot design therefore breaks under the sampling CI, with the asymmetry working systematically against the LLM agents.

\paragraph{Evaluation horizon.}
The headline LiveMacro Score compares the model's latest pre-release nowcast against Bloomberg's final consensus median, a terminal-snapshot design that aligns the two information sets at the same point in time. Our continuous nowcasting setup queries the model throughout the prediction window, and the last pre-release nowcast is produced under the broadest available information set, since any signal observable to an earlier nowcast in the window remains observable at the final query. The terminal nowcast therefore represents the model's full predictive ability under the high-frequency updating design. The matching design on the Bloomberg side is the natural counterpart: the Bloomberg survey allows economists to revise their submissions throughout the two weeks preceding each release and accepts new posts until the announcement, with the published consensus continuously re-computed as the median of standing forecasts \citep{kurov2019price}. Matching the terminal model nowcast against Bloomberg's terminal consensus thus delivers the cleanest like-for-like comparison.

\begin{figure}[htbp]
\centering
\includegraphics[width=0.95\linewidth]{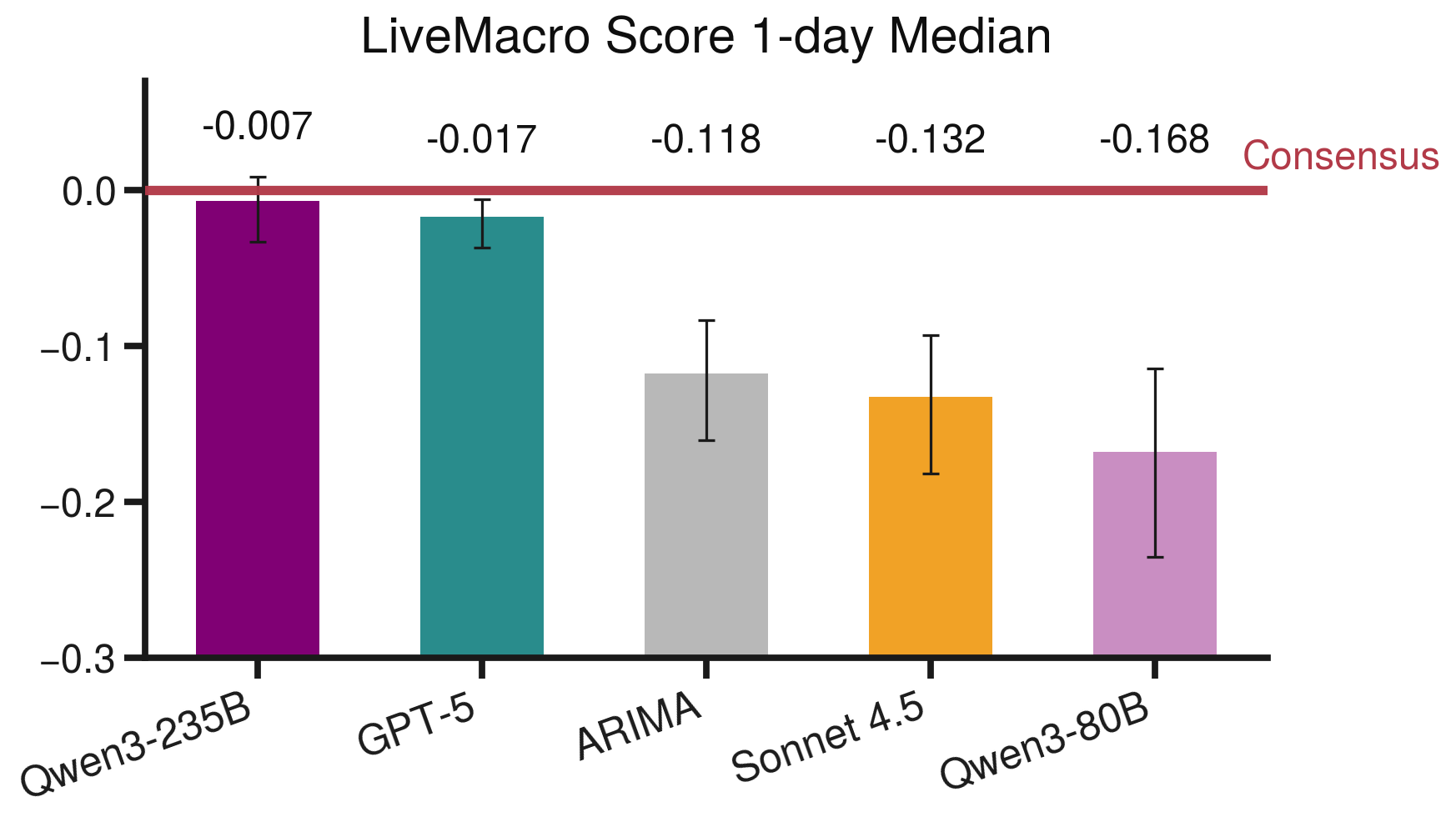}
\caption{LiveMacro Score at the 1-day median evaluation horizon. For each event, the model nowcast is the within-day median of model snapshots on the single calendar day preceding release, and the Bloomberg consensus is the cumulative-median calendar value at the end of that day. The score is then computed with the frozen historical $\{\hat\beta_i\}$ of Section~\ref{sec:metric}. Bars are point estimates and whiskers are 90\% parametric-bootstrap intervals. Qwen3-235B narrowly edges out GPT-5 at the top, with both top models remaining at the level of the Bloomberg consensus. The auto-ARIMA baseline is invariant by construction, and Claude-sonnet-4.5 falls below the auto-ARIMA line at this horizon.}
\label{fig:median1d-livemacro}
\end{figure}

Because the score construction is agnostic to the choice of $\hat X_{i,t}$, we can also evaluate the model at coarser horizons by aggregating model and consensus snapshots over a fixed pre-release window. We re-score using the per-event within-day median of model snapshots and the cumulative-median Bloomberg consensus at the end of the 1, 3, and 7 calendar days before each release. Figure~\ref{fig:median1d-livemacro} reports the LiveMacro Score at the 1-day median horizon. The auto-ARIMA baseline is invariant across horizons by construction. Among the LLM agents, the top two swap positions: Qwen3-235B becomes the model closest to consensus at $-0.007$ with a $90\%$ bootstrap interval $[-0.033, +0.009]$ that straddles zero, narrowly ahead of GPT-5 at $-0.017$. Both top models remain at the consensus level, with GPT-5 undergoing a sign flip from $+0.004$ to $-0.017$ that sits well inside its bootstrap band. Qwen3-80B moves modestly, from $-0.122$ to $-0.168$. The one non-trivial change is Claude-sonnet-4.5, whose score drops from $-0.029$ to $-0.132$ and falls below the auto-ARIMA baseline at this horizon.

The cross-model dispersion in horizon sensitivity is itself informative. A score that is stable across horizons indicates that the agent's nowcast has largely converged before the final pre-release day, so the within-day median and the terminal snapshot deliver comparable predictive content. GPT-5's small horizon shift is consistent with a temporally stable forecast that does not whipsaw during the prediction window. Qwen3-235B moves slightly toward consensus at the 1-day horizon, suggesting that its day-of nowcast distribution is at least as informative as its single terminal prediction. Claude-sonnet-4.5's larger drop implies that its terminal nowcast carries late-arriving signal that the within-day median dilutes, a profile consistent with greater intra-window revision volatility and a more time-sensitive prediction path. Full 3-day and 7-day median results are reported in the supplementary materials.

\begin{figure}[htbp]
\centering
\includegraphics[width=0.95\linewidth]{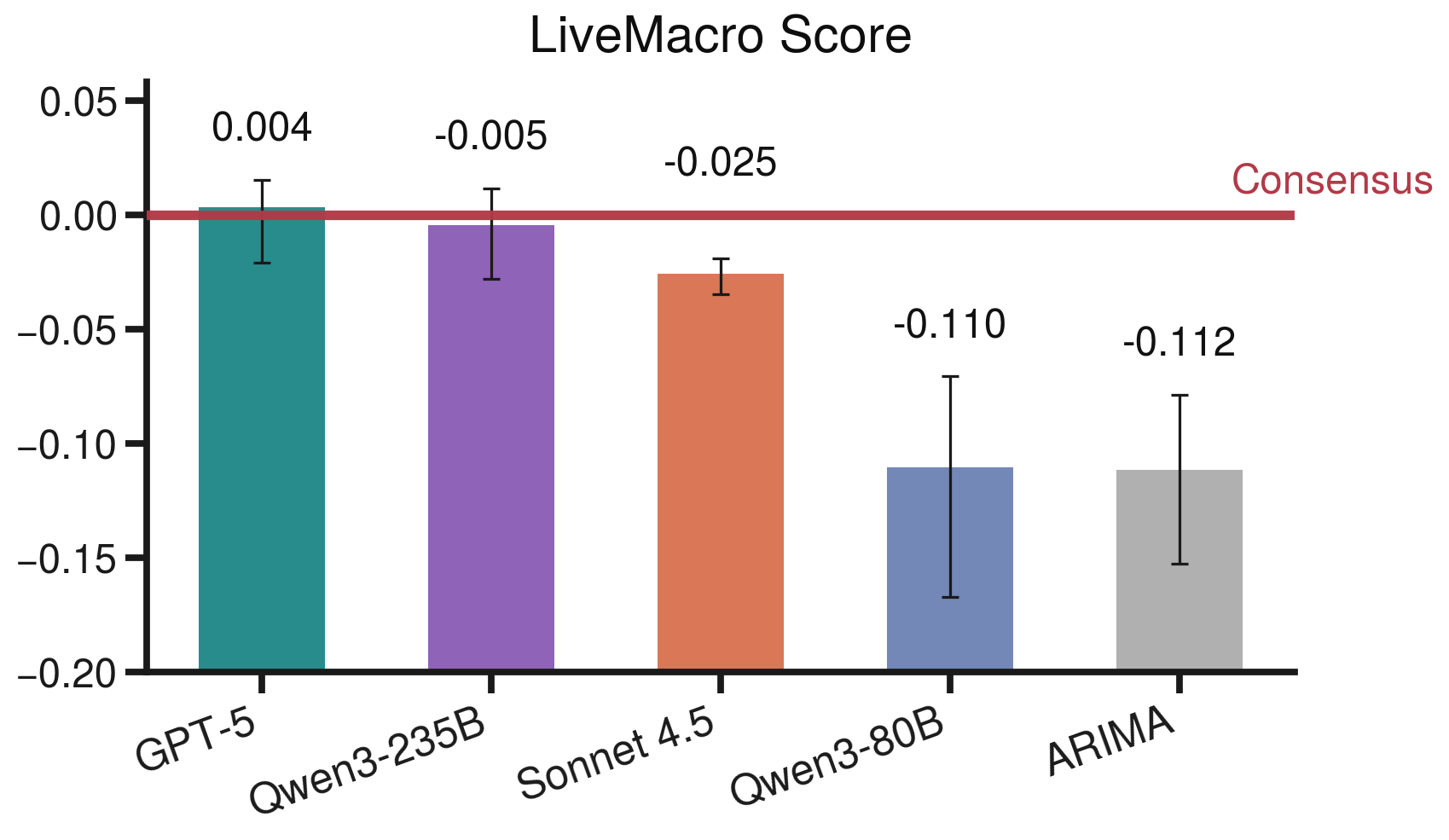}
\caption{LiveMacro Score with 90\% bootstrap confidence intervals, refit on the COVID-excluded historical sample (120 of 2{,}163 timestamp-group events dropped). The ordering of the LLM agents and the sign of the top estimate are preserved relative to the full-sample baseline in Section~\ref{sec:results-livemacro}, and shifts are uniformly small across the panel.}
\label{fig:nocovid-livemacro-no-agent}
\end{figure}

\paragraph{Robustness excluding the COVID-19 window.}
A standard concern in empirical macroeconomics is that the early COVID-19 period violates the linearity and stationarity assumptions implicit in regressions estimated on long historical samples, because it may reflect a distinct economic regime \citep{lenza2022how}. The Huber loss already downweights such observations, but as an additional robustness check we refit the entire historical regression on the COVID-excluded sample, dropping all 120 timestamp-group events with $T_g$ between 2020-03-01 and 2020-12-31 inclusive (5.55\% of the 2{,}163-event sample).
We then re-run the live-scoring pipeline using the no-COVID $\hat\beta$ and a freshly drawn $V_\beta$, holding the live sample, the consensus, and the model predictions fixed.
The headline conclusions of Section~\ref{sec:results-livemacro} survive on the LiveMacro Score (Figure~\ref{fig:nocovid-livemacro-no-agent}). GPT-5 remains the only model with a positive point estimate, sitting just above the Bloomberg-consensus line, and its 90\% CI straddles zero in both the baseline and the no-COVID fit. The remaining LLM agents stay below consensus, with Qwen3-80B and the auto-ARIMA baseline clustered at the bottom, preserving the ordering of Section~\ref{sec:results-livemacro} up to a swap between those two that lies well inside the 90\% bootstrap intervals. Movements from the full-sample baseline are uniformly small, with every LiveMacro Score point estimate shifting by less than $0.025$ in absolute terms and essentially no change at the top of the ranking. Table~\ref{tab:covid-robustness} reports the re-scored values. The one borderline case is Qwen3-235B, whose point estimate moves close enough to zero that its 90\% CI no longer excludes the consensus line. Its shift of $+0.023$ is the largest in the panel and is comparable to the half-width of its own 90\% bootstrap interval in Figure~\ref{fig:bdrc}, so it is on the order of the coefficient sampling uncertainty already reported.

The ranking is stable because the coefficients that carry the weight barely move. Fifteen of the sixteen coefficients keep their sign under the COVID-excluded fit, the exception being the unemployment rate, which accounts for $0.57\%$ of the $\beta^2$ weight. The two dominant coefficients shift least of all, CPI by $+2.7\%$ and ISM manufacturing by $+1.7\%$, and together they carry $61\%$ of the priced-shock variance. The large proportional moves fall on near-zero, low-weight releases and therefore scarcely enter the score. We retain the full-sample $\hat\beta$ as the headline because COVID-window releases were themselves high-impact macro events that moved equity markets, contain genuine surprise information about the surprise-to-return mapping, and are already downweighted by the Huber loss. The no-COVID specification serves as a sensitivity check.

\begin{table}[htbp]
\centering
\footnotesize
\setlength{\tabcolsep}{3pt}
\renewcommand{\arraystretch}{1.15}
\begin{tabular}{@{}lrrrc@{}}
\toprule
\textbf{Model} & \textbf{Frozen} & \textbf{No-COVID} & $\Delta$ & \textbf{Rank} \\
\midrule
GPT-5             & $+0.004$ & $+0.004$ & $-0.001$ & $1 \to 1$ \\
Qwen3-235B        & $-0.028$ & $-0.005$ & $+0.023$ & $2 \to 2$ \\
Claude-sonnet-4.5 & $-0.029$ & $-0.025$ & $+0.003$ & $3 \to 3$ \\
auto-ARIMA        & $-0.118$ & $-0.112$ & $+0.006$ & $4 \to 5$ \\
Qwen3-80B         & $-0.122$ & $-0.110$ & $+0.012$ & $5 \to 4$ \\
\bottomrule
\end{tabular}
\caption{LiveMacro Score under the frozen $\hat\beta$ and under the COVID-excluded $\hat\beta$, on the identical live release sample. $\beta$ is re-estimated on the historical sample with the 2020 COVID window dropped, and the live nowcasts, consensus, and releases are held fixed, so every change is attributable to the coefficients alone. GPT-5 remains the only model above the Bloomberg-consensus line, and the only reordering is between auto-ARIMA and Qwen3-80B at the bottom of the panel, whose intervals overlap heavily.}
\label{tab:covid-robustness}
\end{table}

\subsection{Why conventional error aggregators fail}\label{app:conventional-aggregators}

Section~\ref{sec:metric} argues that weighting by measured market impact is what lets sixteen per-indicator accuracies collapse into one comparable number. This subsection makes that argument quantitative. We recompute the two conventional alternatives on the identical per-release grid the LiveMacro Score consumes, so the comparison holds the data fixed and varies only the aggregator. Table~\ref{tab:aggregator-shares} reports, for each indicator, its share of the raw summed squared error, its share of the equal-weight relative-error sum, and its economic weight under the LiveMacro Score.

\paragraph{Raw error sums fail by scale domination.}
A squared error summed across the sixteen indicators is dominated by whichever series carries the largest units. More than $99.99\%$ of the summed squared error comes from the five level series, the payroll count and the four housing counts. Existing home sales alone accounts for $45\%$ of it, even though its estimated $|\hat\beta_i|$ is the smallest of the sixteen (Table~\ref{tab:hist-beta}). The eleven percentage and index series together contribute less than $0.01\%$. A ranking built on this sum is therefore a ranking on housing and payroll levels, whatever the agent does on inflation or activity.

\paragraph{Relative error removes the scale problem but ranks unstably.}
The natural fix normalizes each indicator by its own scale. This is the relative-error metric defined and plotted per indicator in Appendix~\ref{app:nowcast-error}, read here as a cross-indicator aggregate rather than one curve at a time. It divides the squared error by $s_i^2$, where $s_i$ is the mean absolute release value for that indicator, and it fails in two further ways. First, the denominator sits near zero for the month-over-month percentage-change series. The mean absolute release is $0.16$ for real PCE, $0.40$ for industrial production and $0.48$ for PPI, so a small absolute miss becomes a large relative error. PPI alone is $42\%$ of the relative-error sum while CPI, the release with the largest $|\hat\beta_i|$ in the panel, is $0.1\%$. This is the near-zero-denominator failure of percentage and relative errors documented in the forecasting literature \citep{hyndman2006another,makridakis1993accuracy}. Second, a sum of squared near-zero-denominator terms is dominated by a few extreme events. Between $30.5\%$ and $39.3\%$ of each model's relative-error total comes from a single indicator, and dropping each model's single worst event reorders the panel, moving Qwen3-235B from fourth to second and Claude-sonnet-4.5 from third to fourth, with a rank correlation of $0.70$ between the two orderings. A ranker this sensitive to one event out of the whole live sample is not a reliable basis for comparing agents. The Huber-ridge estimator behind the LiveMacro Score downweights such events by construction.

\paragraph{Equal weighting is not what the evidence supports.}
Even scale-free and outlier-robust, summing sixteen indicators equally treats every release as equally worth forecasting, and the announcement-return evidence says otherwise. Weighting each indicator by its share of announcement-return variance, which is proportional to $\hat\beta_i^2$, puts $33\%$ of the priced shock on CPI, and $79\%$ on CPI, the ISM manufacturing index, retail sales and real GDP together. The effective number of indicators under these weights is $4.7$, against $16$ under equal weighting. The relative-error sum instead puts CPI on the same footing as existing home sales, which carries $0.06\%$ of the priced shock. Evaluating a forecast by a statistical error that ignores its economic value is a long-documented gap in the forecast-evaluation literature \citep{leitch1991economic,granger2000economic,elliott2008economic}, and economic surprise indices aggregate relevance-weighted standardized surprises rather than equal ones \citep{scotti2016surprise}. The $\hat\beta_i$ of the LiveMacro Score are exactly those relevance weights, measured from announcement-window returns rather than assumed.

\begin{table}[htbp]
\centering
\footnotesize
\setlength{\tabcolsep}{4pt}
\begin{tabular}{@{}lrrr@{}}
\toprule
 & \textbf{Raw} & \textbf{Relative} & \textbf{Economic} \\
\textbf{Indicator} & \textbf{error} & \textbf{error} & \textbf{weight} \\
\midrule
CPI                    &    $<0.01$ &     $0.12$ &     $33.0$ \\
ISM Manufacturing      &    $<0.01$ &    $<0.01$ &     $28.2$ \\
Retail sales           &    $<0.01$ &      $7.3$ &      $9.7$ \\
Real GDP               &    $<0.01$ &      $3.0$ &      $8.1$ \\
Nonfarm payrolls       &     $16.0$ &     $11.5$ &      $6.4$ \\
New home sales         &     $11.1$ &     $0.18$ &      $4.3$ \\
PCE price index        &    $<0.01$ &      $2.0$ &      $3.0$ \\
Real PCE               &    $<0.01$ &     $13.6$ &      $1.7$ \\
ISM Services           &    $<0.01$ &    $<0.01$ &      $1.6$ \\
PPI                    &    $<0.01$ &     $41.6$ &      $1.3$ \\
Housing starts         &     $20.5$ &     $0.08$ &      $1.0$ \\
Building permits       &      $7.3$ &     $0.03$ &     $0.59$ \\
Unemployment rate      &    $<0.01$ &     $0.02$ &     $0.57$ \\
Durable goods orders   &    $<0.01$ &      $8.4$ &     $0.32$ \\
Industrial production  &    $<0.01$ &     $12.1$ &     $0.08$ \\
Existing home sales    &     $45.1$ &     $0.02$ &     $0.06$ \\
\bottomrule
\end{tabular}
\caption{What each aggregator is actually measuring, in percent. ``Raw error'' is the indicator's share of the summed squared nowcast error across the sixteen indicators, ``Relative error'' its share of the scale-normalized sum of Appendix~\ref{app:nowcast-error}, and ``Economic weight'' its share of announcement-return variance, proportional to $\hat\beta_i^2$. Rows are ordered by economic weight. Each column sums to $100$. The three columns disagree almost completely, which is the point.}
\label{tab:aggregator-shares}
\end{table}

\section{Simulated Polymarket Betting Return: Full Construction}\label{app:polymarket-return-details}

This appendix gives the full construction of the simulated Polymarket betting return summarized in Section~\ref{sec:metric}.

For each indicator $i$ that trades on Polymarket as a categorical macro market, the market consists of mutually exclusive numeric buckets $\mathcal{B}_i$. Each bucket $b \in \mathcal{B}_i$ trades as a binary contract that pays \$1 at resolution if the released value falls in $b$ and \$0 otherwise; let $p_b(\tau) \in [0, 1]$ denote its Polymarket price at time $\tau$. At each prediction hour $\tau$ strictly before the release timestamp $T_i$, we map the agent's most recent nowcast to the bucket $\hat b_i(\tau)$ that contains it and place a \$1 bet on that bucket, purchasing $1/p_{\hat b_i(\tau)}(\tau)$ shares. Once the release determines the winning bucket $b_i^\star$, the cumulative return on the $N_i$ hourly bets placed on indicator $i$ is
\begin{equation}
R_i = \frac{1}{N_i} \sum_{\tau = 1}^{N_i} \frac{\mathbf{1}\{\hat b_i(\tau) = b_i^\star\}}{p_{\hat b_i(\tau)}(\tau)} - 1.
\end{equation}

\paragraph{Bucket pricing.}
The entry price $p_{\hat b_i(\tau)}(\tau)$ is the crowd's pool-implied probability for that bucket, so bucket liquidity is priced into the return by construction: a \$1 stake on a bucket the crowd already favors buys few shares and pays little even when it wins, while the same stake on a bucket the crowd has underpriced buys many shares and pays a large multiple if the agent turns out to be right. Market prices of this kind are themselves near-unbiased probability forecasts \citep{wolfers2004prediction}, which is what makes a positive $R_i$ readable as beating the crowd. We fix the stake at \$1 per hour rather than optimizing bet size, following the $1/N$ logic that makes naive equal weighting a robust benchmark in asset pricing \citep{demiguel2009optimal}, so the score measures nowcast quality rather than bet sizing.

\paragraph{Polymarket return sample.}
For the Polymarket-style betting return, we restrict the comparison to target variables that, during the live window, admit both a bucketed Polymarket macro market and a matched institutional nowcast reference. Three variables qualify: CPI YoY, unemployment rate, and real GDP QoQ. The corresponding markets are economically non-trivial. Polymarket reports total trading volume of \$69{,}472 for the January~2026 CPI YoY market, \$308{,}214 for the March~2026 unemployment-rate market, and \$455{,}385 for the Q4~2025 GDP growth market \citep{polymarket2026janinflation,polymarket2026marchunemp,polymarket2026q4gdp}. We fetch Polymarket bucket prices from the Polymarket API at hourly frequency. For each release we use the latest estimate strictly before the release timestamp, matching the rule used in the capture score. Returns are reported for the same LLM agents, the auto-ARIMA baseline, the Bloomberg consensus, and the available regional-Fed nowcasts.

\section{Detailed Analysis of LiveMacro Score Results}\label{app:livemacro-results-details}

This appendix expands the headline LiveMacro score result reported in Section~\ref{sec:results-livemacro} with per-model numbers, uncertainty bands, and additional analysis.

\paragraph{Interpretation of the zero baseline.}
The Bloomberg consensus aggregates the medians of 50 to 80 professional forecasters per release, and is the strongest publicly available pre-release expectation for U.S.\ headline macro indicators. Systematically outperforming this consensus on scheduled macro releases is extraordinarily difficult \citep{faustwright2013forecasting}. The natural prior is therefore that most nowcasts will not produce a meaningfully positive surprise capture. A positive score reads as return-generating signal at the announcement window: the model's predicted shock aligns with realized E-mini returns beyond what the consensus already impounds, so the residual is priceable.

\paragraph{GPT-5 versus the consensus.}
Against this benchmark, GPT-5 is the only model with a positive score, at $+0.004$, with a $90\%$ parametric-bootstrap confidence interval that narrowly straddles zero. By construction, a score of $+0.004$ implies a $\approx 0.8\%$ reduction in the mean-squared error of the predicted announcement-window equity-return shock relative to the consensus baseline. Equivalently, GPT-5's nowcasts account for an additional $\approx 0.8\%$ of the variance in announcement-window E-mini returns beyond what the consensus already prices in. Set against the documented difficulty of beating the professional consensus, this is an economically non-trivial gain.

\paragraph{Remaining models.}
The remaining four models post negative scores. Claude-sonnet-4.5 and Qwen3-235B cluster together at approximately $-0.028$, statistically indistinguishable from the consensus reference, in effect reproducing the consensus signal without adding any equity-relevant information. The auto-ARIMA baseline at $-0.118$ and Qwen3-80B at $-0.122$ are the worst performers, with overlapping $90\%$ bootstrap intervals. The ARIMA baseline issues a single point forecast at the start of each prediction window and does not condition on intra-window news, so it operates at a structural information disadvantage relative to the real-time benchmarks. The spread between GPT-5 and the remaining LLM agents indicates that the capacity to outperform a strong expectation benchmark on a heterogeneous set of macro releases is far from automatic across the current frontier.

\paragraph{Per-indicator decomposition.}
The aggregate score of Section~\ref{sec:results-livemacro} can be decomposed into additive per-indicator contributions. We allocate the quadratic realized-return score across indicators by an exact Shapley allocation, so the contributions sum to each model's published LiveMacro Score, including on timestamp-group events where several indicators release together. Table~\ref{tab:indicator-decomposition} reports two quantities per indicator. The \emph{average} is the equal-weight mean of the four LLM agents, which is a summary of the panel rather than an ensemble forecast. The \emph{best} is the ex-post best agent for that indicator, which is an oracle diagnostic and a capability upper bound, not a deployable single model. A positive contribution means the agent's implied surprise improves on the Bloomberg consensus for that indicator.

The decomposition makes the heterogeneity precise. The average agent contributes positively on 4 of the 16 indicators, namely ISM manufacturing ($+0.0012$), existing home sales ($+0.0011$), building permits ($+0.0007$), and industrial production ($+0.0004$), all activity or housing-supply series. Its three largest drags are CPI ($-0.0228$), retail sales ($-0.0077$), and the PCE price index ($-0.0075$), which together contribute $-0.0380$ and account for $87\%$ of the average agent's aggregate score of $-0.0437$. The per-indicator best model is positive on 12 of 16, against 4 of 16 for the average, and the identity of that best model changes from indicator to indicator. GPT-5 leads on eight of the sixteen and each of the other three agents leads on at least two, so per-indicator strength is spread across model families.

Real GDP is not the weak spot the aggregate might suggest. Three of the four agents contribute positively there, Qwen3-235B at $+0.0016$, GPT-5 at $+0.0012$, and Qwen3-80B at $+0.0005$. The negative average of $-0.0017$ comes from Claude-sonnet-4.5 alone, at $-0.0100$. Real GDP is also middling rather than worst on the real-time relative error of Appendix~\ref{app:nowcast-error}. The LiveBetting return points the same way, with three agents finishing ahead of the Atlanta Fed GDPNow on real GDP (Section~\ref{sec:results-livebetting}). The shortfall is concentrated instead on the anchored-inflation and retail series.

The pattern has an economic reading. U.S.\ inflation expectations have been firmly anchored for two decades \citep{bernanke2007inflation}. In that regime survey and consensus forecasts of inflation are very hard to improve on \citep{ang2007do,faustwright2013forecasting}, so the agents rarely get ahead of the consensus on CPI or the PCE price index. Only GPT-5 on CPI and Claude-sonnet-4.5 on the PCE price index contribute positively, and the average agent is negative on both. The Cleveland Fed further runs a purpose-built daily oil-and-gasoline model on this target \citep{knotek2017nowcasting}. Retail sales and the other hard demand series reverse sharply from month to month, and the agents smooth toward the trailing trend and miss the turn. The wins fall where timely real-time signal matters most and the consensus edge is smallest. Survey and soft indicators such as the ISM PMI carry the most incremental nowcasting information early in the data cycle, before the hard data arrive \citep{giannone2008nowcasting,banbura2013nowcasting}, building permits is a forward-looking component of the Conference Board Leading Economic Index, and a web-search agent can additionally exploit real-time news and sentiment, which is known to improve macroeconomic forecasts of output, inflation, and unemployment \citep{kalamara2022making}.

\begin{table*}[tp]
\centering
\footnotesize
\setlength{\tabcolsep}{8pt}
\renewcommand{\arraystretch}{1.15}
\begin{tabular}{@{}lrl@{}}
\toprule
\textbf{Indicator} & \textbf{Avg.\ contribution} & \textbf{Best contribution (model)} \\
\midrule
\multicolumn{3}{@{}l}{\textit{Supply and Production}} \\
\addlinespace[2pt]
Real GDP                & $-0.0017$ & $+0.0016$ (Qwen3-235B) \\
Industrial Production   & $+0.0004$ & $+0.0008$ (GPT-5) \\
Durable Goods           & $-0.0009$ & $-0.0003$ (Qwen3-80B) \\
ISM Manufacturing       & $+0.0012$ & $+0.0093$ (GPT-5) \\
\midrule
\multicolumn{3}{@{}l}{\textit{Demand and Inflation}} \\
\addlinespace[2pt]
CPI                     & $-0.0228$ & $+0.0088$ (GPT-5) \\
PCE Price Index         & $-0.0075$ & $+0.0009$ (Claude-sonnet-4.5) \\
PPI                     & $-0.0001$ & $+0.0028$ (GPT-5) \\
Real PCE                & $-0.0002$ & $+0.0011$ (GPT-5) \\
Retail Sales            & $-0.0077$ & $-0.0020$ (GPT-5) \\
ISM Services            & $-0.0011$ & $+0.0004$ (Qwen3-80B) \\
\midrule
\multicolumn{3}{@{}l}{\textit{Labor Market}} \\
\addlinespace[2pt]
Nonfarm Payrolls        & $-0.0013$ & $+0.0003$ (Claude-sonnet-4.5) \\
Unemployment Rate       & $-0.0017$ & $\phantom{+}0.0000$ (GPT-5) \\
\midrule
\multicolumn{3}{@{}l}{\textit{Housing}} \\
\addlinespace[2pt]
Building Permits        & $+0.0007$ & $+0.0019$ (GPT-5) \\
Housing Starts          & $-0.0021$ & $-0.0003$ (Claude-sonnet-4.5) \\
New Home Sales          & $-0.0002$ & $+0.0060$ (Qwen3-235B) \\
Existing Home Sales     & $+0.0011$ & $+0.0020$ (Qwen3-80B) \\
\bottomrule
\end{tabular}
\caption{Per-indicator decomposition of the aggregate LiveMacro Score of Figure~\ref{fig:bdrc}. Each entry is an additive share, in LiveMacro Score points, of the aggregate score, computed on the identical setting: the same four LLM agents, the same live evaluation window, and an exact Shapley allocation whose contributions sum to each model's published aggregate. A positive value means the agent improves on the Bloomberg consensus for that indicator. ``Avg.'' is the equal-weight mean of the four agents, whose aggregate score is $-0.0437$. ``Best'' is the ex-post per-indicator best agent, a capability upper bound rather than one deployable model.}
\label{tab:indicator-decomposition}
\end{table*}

\section{Alternative Performance Measures}\label{app:livemacro-directional}

\paragraph{LiveMacro Directional Score.}
We also construct a \emph{directional} version of the LiveMacro Score. The two measure complementary properties. The LiveMacro Score penalizes the squared distance between the predicted and realized announcement-window returns, so it rewards well-scaled magnitude. The directional version asks only whether the model's predicted return matches the sign of the realized announcement-window market move, weighting each event by its economic magnitude, so it rewards directional accuracy alone. The complementarity runs in two ways. First, squared-error scores can be dominated by a few large events, while a sign-based aggregation gives a more stable read on broad-based directional skill. Second, the sign of the predicted return determines whether a long or short position into the announcement is profitable, so directional accuracy is the property that maps most directly onto the profit and loss of event-driven trades \citep{andersen2007realtime}.

The score takes the magnitude-weighted hit-rate form
\begin{equation}
\mathrm{LMD} = \frac{\sum_g |r_g| \, \mathbf{1}\{\operatorname{sgn}(\hat r_g) = \operatorname{sgn}(r_g)\}}{\sum_g |r_g|},
\end{equation}
where the sum runs over $g \in \mathcal{G}^{\mathrm{live}}$, $\mathrm{LMD} \in [0,1]$, $\hat r_g = \hat Q_g$ is the model's predicted announcement-window return (Equation~\ref{eq:r-hat}), and $r_g$ is the realized announcement-window market return. A score of $1$ means the model gets the sign right on every event by weighted mass, $0$ means it is wrong on every event, and $0.5$ is the coin-flip reference, the expected score of a predictor whose sign is independent of the outcome.

\begin{figure}[htbp]
\centering
\includegraphics[width=0.95\linewidth]{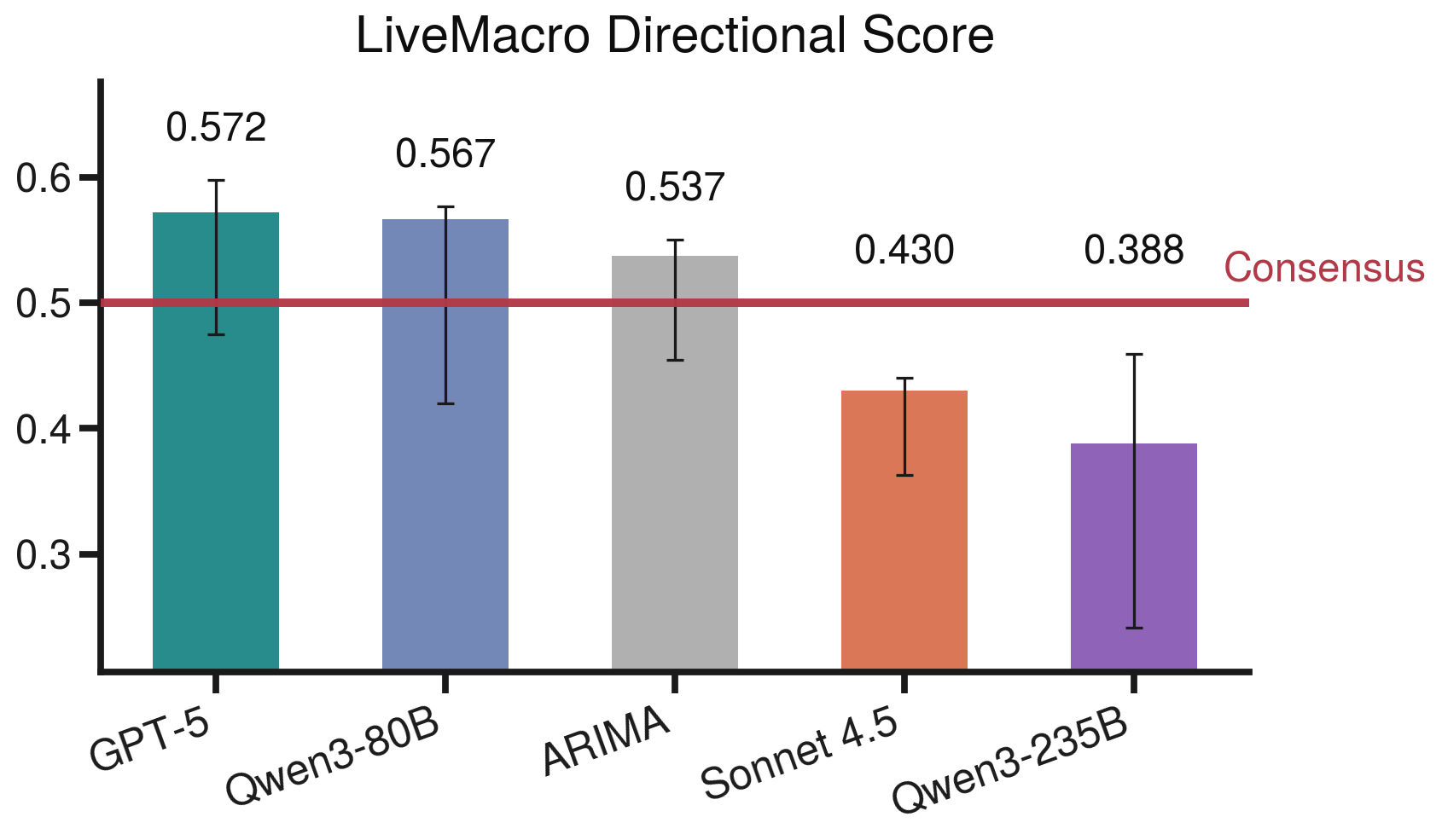}
\caption{LiveMacro directional score (magnitude-weighted hit rate against $\operatorname{sign}(r_g)$) across models on the live evaluation sample. Error bars are $90\%$ parametric-bootstrap confidence intervals over $\{\beta_i\}$. The dashed line at $0.5$ is the coin-flip reference.}
\label{fig:bdrc-wdhr}
\end{figure}

Figure~\ref{fig:bdrc-wdhr} reports the LiveMacro directional score across models. GPT-5 leads at $0.572$, with Qwen3-80B and the auto-ARIMA baseline statistically indistinguishable just below. These three are the only models at or above the $0.5$ coin-flip reference. Claude-sonnet-4.5 and Qwen3-235B fall not only below coin-flip but below the minimally informed auto-ARIMA baseline, with Qwen3-235B last at $0.388$.

A model can score well on the LiveMacro Score either by getting signs right or by shrinking predictions toward zero on uncertain events. The LiveMacro directional score isolates the first channel, weighting each event by the size $|r_g|$ of the move that an event-driven trade would have been sized against, and is by construction immune to the cautious-magnitude strategy that can flatter the LiveMacro Score. Read jointly, the two scores separate genuine forecasting skill from scores flattered by shrinkage. Appendix~\ref{app:kappa-phi} gives an exact form of this trade-off, decomposing any LiveMacro Score into a position size and a directional alignment. This partitions the panel by the source of LiveMacro Score performance. GPT-5's stability at the top of both signals a forecast that is both directionally correct and well-scaled, the profile most likely to translate into positive risk-adjusted returns on positions taken around announcements. Claude-sonnet-4.5 and Qwen3-235B sit near the top of the LiveMacro Score but below coin-flip on direction, indicating that their LiveMacro Score performance comes from conservative magnitude calibration that limits squared loss rather than from genuine directional skill, and would not translate into profitable long-short positioning at announcements. Qwen3-80B shows the opposite asymmetry, with a clear directional signal on the realized market move even though its overall magnitude calibration is weaker on the LiveMacro Score. 

\section{Detailed Analysis of LiveBetting Score Results}\label{app:livebetting-results-details}

This appendix expands the LiveBetting score results reported in Section~\ref{sec:results-livebetting} with per-indicator dynamics and additional discussion.

\paragraph{Setup.}
We report cumulative returns from a simulated Polymarket-style bet executed under the trading rule defined in Section~\ref{sec:metric}. For each indicator we plot the complete nowcasting timeline across the February and March target windows and aggregate to a two-month cumulative return. Real GDP is reported quarterly, so both windows target the same Q1~2026 advance estimate.

\paragraph{Why cumulative LiveBetting returns can be large.}
Two features of the construction jointly explain the large cumulative returns observed in Figure~\ref{fig:polymarket-cum-claude}. First, the score places a fresh \$1 stake every hour, so each pre-release window contains hundreds to thousands of independent \$1 bets rather than one terminal bet at the release. Second, the per-bet payoff is the inverse of a small bucket price: a \$1 stake at a share price of $p$ buys $1/p$ shares, and at resolution each winning share pays \$1, so a single winning bet at $p = 0.11$ returns $1/0.11 \approx 9.1\times$ the stake. When the agent's modal bucket coincides with the eventually realized bucket persistently at low prevailing prices, this multiplicative payoff compounds across the hourly stream. As a concrete illustration, between 2026-02-27 05:00 and 2026-03-03 18:00 in the Q1~2026 real-GDP market every one of the New York Fed Staff Nowcast's $110$ stitched hourly bets in the kept window lay on the eventual winning bucket ``$2.0$--$2.5\%$'' at share prices ranging from \$0.07 to \$0.125 (mean \$0.1103), so each \$1 bet bought roughly nine winning shares and the segment alone contributes a cumulative gain on the order of $\sim 8\times$ the staked capital. Such large per-window numbers are therefore mechanical consequences of (i) the high-frequency \$1-per-hour stake schedule and (ii) the inverse-price share economics of Polymarket contracts, not an artifact of any single binary outcome.

\paragraph{Real GDP.}
Figure~\ref{fig:polymarket-cum-claude} (top) reports cumulative Polymarket returns on the Q1~2026 real GDP nowcast. The New York Fed Staff Nowcast is the strongest series on the panel. The two Qwen variants follow, and all three series rise above the Bloomberg consensus inside the final week before release. Claude-sonnet-4.5 ranks next, narrowly beating the Atlanta Fed GDPNow. GPT-5 follows, opening with negative cumulative returns and crossing into positive territory near day~12, suggesting an effective information update. All four LLM agents and the New York Fed Nowcast end the window with positive cumulative Polymarket returns. Positive cumulative returns admit a direct economic interpretation: at the moments bets are placed, the agent's real-time conditional distribution places more mass on the eventually realized bucket than the Polymarket crowd's distribution does.\footnote{Two channels drive this gap. Through an accuracy channel, the agent's modal bucket matches the eventually realized bucket more often than the crowd-implied modal bucket, averaged across bet timestamps. Through a timing channel, the agent revises its nowcast in response to incoming information before the market reprices the corresponding contract, locking in a low entry price $p_{\hat b(\tau)}(\tau)$ on a bucket that still wins at resolution.} The St.~Louis Fed Real GDP Nowcast is the only series with negative cumulative returns throughout the two-month window, failing to beat the Polymarket crowd at any point.

The Atlanta and New York Fed nowcasts are the most closely watched institutional GDP nowcasts in the U.S.\ market. The New York Fed Staff Nowcast also leads the academic literature on big-data dynamic-factor nowcasting, with a methodology that has been refined across successive generations: the foundational dynamic-factor framework \citep{giannone2008nowcasting,bok2018nowcasting}, the pandemic-robust redesign \citep{nyfed_nowcast}, and a recent extension that combines national-accounts identities with the dynamic-factor model \citep{okeeffe2025component}. Its first-place finish is therefore unsurprising. The non-trivial finding is that three LLM agents beat the Atlanta Fed GDPNow, which itself revises several times per month off scheduled macro releases \citep{atlfed_gdpnow}. Around day~9 of the window, Claude-sonnet-4.5 and GDPNow undergo near-simultaneous revisions that swing both series from negative to positive cumulative returns. The co-movement is suggesting that the LLM agent ingests the same information shock in real time and updates its posterior accordingly. Finally, the Bloomberg consensus substantially outperforms the Polymarket crowd. This pattern is consistent with our framing of the consensus as a strong professional-expectation benchmark.

\paragraph{Headline CPI.}
Figure~\ref{fig:polymarket-cum-claude} (middle) reports cumulative Polymarket returns on the February-print and March-print CPI nowcasts. The Cleveland Fed Inflation Nowcasting model, the only institutional Fed nowcast in our reference set that targets U.S.\ inflation, achieves the highest cumulative return. The Bloomberg consensus is second. The two Qwen variants follow, both narrowly above the Polymarket crowd. GPT-5 is positive through the early February-print window but ends the two-month aggregation in negative territory. Claude-sonnet-4.5 is the worst performer, accumulating near-total losses in the first weeks of the window and recovering only partially on late-window revisions.

The institutional-and-consensus lead on inflation reflects the regime of well-anchored U.S.\ inflation expectations sustained by the Federal Reserve's inflation-stabilization mandate, a regularity widely attributed to the credibility of that mandate \citep{bernanke2007inflation,faustwright2013forecasting}. The Cleveland Fed nowcast further exploits daily oil and gasoline prices through a deterministic-model-switching design that has historically outperformed both the Blue Chip consensus and the Survey of Professional Forecasters on headline-inflation nowcasting \citep{knotek2017nowcasting}.

\paragraph{Why the CPI curves diverge.}
The six CPI curves trend in opposite directions, and the hourly cadence lets us attribute each trend to the bets that drive it. The betting return moves for two reasons, whether the staked bucket matched the final release and how cheap that bucket was when bought. The February print resolved to the $2.4\%$ bucket and the March print to the $\ge 2.8\%$ bucket, and in both cases the correct bucket was cheap early and expensive later, priced at the crowd's implied probability, so being right early paid far more than being right late. One consequence is purely mechanical. Once the crowd prices the correct bucket near one, a fresh stake is break-even and merely grows the denominator, pulling a model's cumulative return toward zero, which is downward for a model sitting above zero and upward for a model sitting below it.

The strong forecasters therefore decline without making a new error. The Cleveland Fed nowcast stakes the winning bucket in all $375$ of its hourly bets, yet its curve still slides from $+138\%$ to $+91.5\%$, because the denominator grows from $247$ to $375$ break-even April bets. The same force drags the Bloomberg consensus from $+66\%$ to $+44\%$. Table~\ref{tab:cpi-decomposition} separates this dilution from genuine forecast error using the counterfactual return that would have resulted had March contributed exactly zero profit.

Claude-sonnet-4.5 is the mirror image of the same force. It carries a persistent high-inflation lean, holding $2.5\%$ for $143$ consecutive hours in February, which pinned it near the $-100\%$ floor. The energy-driven hot March partly vindicated that lean and its win rate more than triples, from about $10\%$ to $38\%$. On 2026-04-05 it finally commits to the correct $\ge 2.8\%$ bucket, and because its curve sits far below zero those break-even bets dilute it upward, from $-82\%$ to $-72\%$. The Bloomberg consensus shows the same arithmetic earlier in the window and in the other direction. It starts at $-100\%$, then turns up on 2026-03-03 when the consensus lands on the right bucket, and that one cheap and correct bet lifts it from $-100\%$ to $+42\%$. The consensus and the Cleveland Fed stay high because their modal bucket was right in both months, and the Cleveland Fed's daily oil-and-gasoline model \citep{knotek2017nowcasting} is built to price exactly this kind of energy shock, a structural edge the agents lack.

The other agents fall through two distinct errors around the same shock, and the event window pins down each one. Qwen3-80B under-reacts. It is correct in February, but in March it stays anchored at $2.6$ to $2.7\%$ and misses the $\ge 2.8\%$ surge entirely, winning none of its $128$ March bets. Qwen3-235B over-reacts in the wrong month. The moment the oil shock lands on 2026-03-04 it flips its February bet to $\ge 2.7\%$, pricing into the February index a surge that the late-February shock was too early to move, and collapsing a $+156\%$ lead to near zero. GPT-5 is unstable rather than wrong in one direction. It changes bucket frequently and keeps reverting to a $2.4\%$ low print in April, so its $-40\%$ comes from many noisy misses rather than one bad call.

\begin{table*}[tp]
\centering
\footnotesize
\setlength{\tabcolsep}{8pt}
\renewcommand{\arraystretch}{1.15}
\begin{tabular}{@{}lrrrrr@{}}
\toprule
\textbf{Series} & \textbf{Feb win rate} & \textbf{Feb profit (\$)} & \textbf{Mar profit (\$)} & \textbf{Final return} & \textbf{If Mar $=0$} \\
\midrule
Cleveland Fed     & $100\%$ & $+341.5$ & $+1.7$   & $+91.5\%$ & $+91.1\%$ \\
Bloomberg         & $71\%$  & $+164.2$ & $+1.7$   & $+44.2\%$ & $+43.8\%$ \\
Qwen3-80B         & $70\%$  & $+169.9$ & $-128.0$ & $+11.2\%$ & $+45.3\%$ \\
Qwen3-235B        & $42\%$  & $+1.9$   & $+1.7$   & $+1.0\%$  & $+0.5\%$ \\
GPT-5             & $31\%$  & $-64.3$  & $-86.5$  & $-40.2\%$ & $-17.1\%$ \\
Claude-sonnet-4.5 & $10\%$  & $-190.6$ & $-79.4$  & $-72.0\%$ & $-50.8\%$ \\
\bottomrule
\end{tabular}
\caption{Decomposition of the cumulative CPI LiveBetting return into its February and March segments. ``Feb win rate'' is the February win rate, and the dollar columns are realized segment profits on the \$1-per-hour stake schedule over $247$ February and $128$ March bets. ``If Mar $=0$'' is the counterfactual final return had March contributed exactly zero profit, that is the February profit spread over all $375$ bets. Where it matches the actual final return, as for the Cleveland Fed, Bloomberg, and Qwen3-235B, the late decline is entirely the mechanical dilution of a growing denominator rather than a new forecast error.}
\label{tab:cpi-decomposition}
\end{table*}

\paragraph{Unemployment rate.}
Figure~\ref{fig:polymarket-cum-claude} (bottom) reports cumulative Polymarket returns on the February-print and March-print unemployment-rate nowcasts. Claude-sonnet-4.5 is the strongest series on the panel, with a cumulative two-month return materially above every other LLM agent and the institutional reference. GPT-5 follows. It posts winning bets in the early days of the February window, stops beating the Polymarket crowd after roughly day~5, and ends the window in negative territory. The Chicago Fed CHURN nowcast sits below GPT-5 and also finishes negative. Qwen3-80B, Qwen3-235B, and the Bloomberg consensus all bottom out at $-100\%$ across the entire two-month timeline. This floor is reached only when the agent's modal bucket fails to coincide with the eventually realized bucket at every bet timestamp, so every staked dollar is lost at resolution. Both Claude-sonnet-4.5 and GPT-5 thus clear the Bloomberg consensus on unemployment, while the Qwen variants tie with it at the $-100\%$ floor.

Unlike the GDP and CPI cases, the institutional and professional consensus do not display an advantage over the leading LLM agents on unemployment nowcasting. The Chicago Fed CHURN model outperforms both the Bloomberg consensus and a random-walk benchmark from 2018 to 2025, but its weekly cadence is coarse and its feature set excludes textual macro news \citep{chicagofed_churn}.

\section{Detailed Analysis of LiveMacro Score by Theme}\label{app:livemacro-theme-results-details}

This appendix expands the theme decomposition reported in Section~\ref{sec:results-livemacro-theme} with per-model numbers and additional discussion.

\paragraph{Per-model breakdown.}
Figure~\ref{fig:bdrc-by-theme} decomposes the LiveMacro score across the four indicator themes. GPT-5 is the most consistent performer, with the tightest cross-theme spread in the panel: it posts $+0.057$ on Supply and Production, sits essentially at the Bloomberg consensus on Demand and Inflation ($-0.003$) and on Labor Market ($\approx 0$), and falls slightly below consensus on Housing ($-0.021$), so it matches or ties the consensus in three of the four themes. Qwen3-80B is the weakest LLM agent overall, sitting in the bottom two in three themes, and the auto-ARIMA baseline posts the worst single-model score in two themes (Demand and Inflation at $-0.188$ and Housing at $-0.056$). Qwen3-235B behaves as a specialist: it leads on Housing ($+0.060$) and is essentially at consensus on Supply and Production ($-0.001$), but is clearly negative on Demand and Inflation ($-0.034$) and Labor Market ($-0.026$).

\paragraph{Cross-theme numerical detail.}
Three numerical patterns complement the main-paper interpretation in Section~\ref{sec:results-livemacro-theme}. On Supply and Production and on Housing the largest single-model gains in the panel are concentrated, with GPT-5 leading Supply and Production at $+0.057$ and Qwen3-235B leading Housing at $+0.060$. On Demand and Inflation only GPT-5 sits at the consensus, with the remaining LLM agents and the auto-ARIMA baseline posting the largest negative gaps in the panel. On Labor Market the scores cluster within $\pm 0.03$ of zero across every model, consistent with the household-survey signal-to-noise ceiling identified for the unemployment rate.

\section{Cross-Model Error Analysis}\label{app:error-analysis}

This appendix consolidates the recurring errors across the four LLM agents. We separate the failures the agents share from each model's own pattern of error, working from each agent's final pre-release nowcast against the official release and against the Bloomberg consensus on the live scoring grid.

\paragraph{Shared failure: the anchored-inflation floor.}
On headline inflation the professional consensus is close to exact, so there is very little room for an agent to add signal. Across the five scored CPI months the consensus misses the release by at most $0.1$ percentage point in four of them, and on the PCE price index it is exactly right on four of the five prints. A baseline this accurate is effectively unbeatable, which is why every agent posts a negative Demand and Inflation score and only GPT-5 sits at the consensus (Appendix~\ref{app:livemacro-theme-results-details}). The same pattern appears in the betting results, where the Cleveland Fed and the consensus lead every agent on CPI (Section~\ref{sec:results-livebetting}). This advantage is structural rather than incidental. It rests on the anchored-expectations regime that has held U.S.\ inflation forecasts close to target \citep{bernanke2007inflation,faustwright2013forecasting} and, for the Cleveland Fed, on a daily oil-and-gasoline model built for exactly this series \citep{knotek2017nowcasting}.

\paragraph{Shared failure: sharp month-over-month reversals.}
The second blind spot is the turning point. When a series reverses direction sharply, every agent smooths toward the trailing trend and misses the turn in the same direction. Table~\ref{tab:err-shared} collects the three clearest cases in the live window. On February durable goods all four agents over-predict a decline and two of them predict outright growth. On March industrial production all four predict growth into a contraction. On March retail sales all four understate the surge, nowcasting between $+0.4$ and $+0.6$ against a $+1.7$ print. Durable goods orders turn on aircraft bookings, industrial production on energy output, and retail sales on gasoline receipts, so each series can reverse sharply from one month to the next. The misses are shared rather than model-specific. The contrast with Appendix~\ref{app:nowcast-error} is informative. Where the consensus is stale and intra-window updating helps, every agent beats it near release on nonfarm payrolls, PPI and the ISM services index. The shared failures are therefore specific to anchored and reversal releases rather than a blanket weakness.

\begin{table}[htbp]
\centering
\footnotesize
\setlength{\tabcolsep}{4pt}
\begin{tabular}{@{}lrrr@{}}
\toprule
 & \textbf{Durable} & \textbf{Industrial} & \textbf{Retail} \\
 & \textbf{goods} & \textbf{production} & \textbf{sales} \\
\midrule
Official release    & $-1.4$  & $-0.5$  & $+1.7$ \\
Bloomberg consensus & $-1.2$  & $+0.1$  & $+1.4$ \\
\midrule
GPT-5               & $-0.50$ & $+0.30$ & $+0.40$ \\
Claude-sonnet-4.5   & $-0.25$ & $+0.19$ & $+0.60$ \\
Qwen3-235B          & $+0.50$ & $+0.20$ & $+0.60$ \\
Qwen3-80B           & $+0.89$ & $+0.15$ & $+0.60$ \\
\bottomrule
\end{tabular}
\caption{The three sharp month-over-month reversals in the live window, in percent change. The columns are the February durable goods orders release, the March industrial production release, and the March retail sales release. Agent rows are each agent's final pre-release nowcast. In every column all four agents err in the same direction, and on industrial production all four get the sign of the change wrong.}
\label{tab:err-shared}
\end{table}

\paragraph{Per-model patterns of error.}
Around that shared floor each agent fails in its own way, and reading the LiveMacro Score together with its directional variant separates the causes. We take the four agents in order of their aggregate score, with the middle two effectively tied. GPT-5 leads at $+0.004$, holds the highest directional score at $0.572$, and has the tightest cross-theme spread in the panel. Its one systematic weakness is under-predicting housing levels, seen most clearly on the January housing starts release, where it nowcast $1.26$M against a $1.49$M print while Claude-sonnet-4.5 nowcast $1.48$M. That error is cheap in the aggregate, because the housing series carry small surprise-to-return sensitivities. Housing starts has a $|\hat\beta_i|$ roughly one sixth of CPI's, and three of the four housing series fall in the bottom six of the sixteen indicators by $|\hat\beta_i|$, so even a large level miss barely moves the market-weighted score.

Claude-sonnet-4.5 follows at $-0.029$ and is the opposite case, a high-variance overshooter. Its advance Q4 GDP nowcast of $4.24\%$ against a $1.4\%$ release, with the consensus at $2.8\%$, overshoots by more than four standard deviations of the release surprise. Its LiveMacro standing therefore rests on cautious magnitude calibration rather than directional skill, which is why it falls below the coin-flip line at $0.430$. In February its CPI nowcast also sat one bucket above the realized print for most of the window, which gives it the worst CPI betting return in the panel (Appendix~\ref{app:livebetting-results-details}). Qwen3-235B sits alongside it at $-0.028$ but fails for a different reason, stale anchors that produce sign errors. Its unemployment nowcast is frozen at $3.9\%$ across both scored months while the realized rate ran at $4.3$ to $4.4\%$, and that single anchor is the direct cause of both its last-place Labor Market score and its last-place directional score of $0.388$. Qwen3-80B is last at $-0.122$ and is the clearest failure-to-update case, holding its March CPI nowcast between $2.60\%$ and $2.70\%$ through an energy-driven spike to $3.3\%$. Its directional score of $0.567$ nonetheless sits above the coin-flip line, the asymmetry noted in Appendix~\ref{app:livemacro-directional}, so what it lacks is magnitude calibration rather than sign.

\paragraph{Case study: divergent reactions to the same information.}
The hourly cadence lets us watch the agents split on a single public signal rather than infer the split from aggregate scores. On April~8, 2026 the EIA Petroleum Status Report and the FOMC minutes both pointed to upside inflation risk, and Section~\ref{sec:case-study} documents GPT-5 stepping its March CPI nowcast up from about $2.7\%$ to about $3\%$ inside that window. Table~\ref{tab:err-event} summarizes where each agent stood in the week before the release and where it finished. The release confirmed the signal. Headline CPI came in at $3.3\%$ over the year and $0.9\%$ over the month, with gasoline up $21.2\%$ and the energy index up $10.9\%$ on the month, while core inflation held at $2.6\%$ \citep{bls2026cpit01}. That gap between headline and core is what makes the outcome diagnostic. Qwen3-80B held the tightest distribution of the four, an interquartile range of $0.05$ percentage point centred on $2.65\%$, and still finished $0.60$ percentage point below the release. It was tracking the core rate and never revised, so stability here is a symptom of failing to update rather than of confidence. Qwen3-235B sat above the rest of the panel through the week, moved little, and finished $0.10$ percentage point below. Claude-sonnet-4.5 and GPT-5 were both dispersed over the same window, but only GPT-5 resolved that dispersion into an upward revision on April~8 and landed on the release, while Claude-sonnet-4.5 finished $0.12$ percentage point above it. All four agents run the same search-enabled protocol and the same prompt, so the April~8 releases were equally available to all of them. We do not record what an agent chose to search or retrieve in a given hour. The split therefore seems to turn on which public signals an agent treats as decision-relevant for the series it is nowcasting, rather than on unequal access to information.

\begin{table}[htbp]
\centering
\footnotesize
\setlength{\tabcolsep}{4pt}
\begin{tabular}{@{}lrrrr@{}}
\toprule
\textbf{Model} & \textbf{Median} & \textbf{IQR} & \textbf{Final} & \textbf{Deviation} \\
\midrule
GPT-5             & $2.50$ & $0.40$ & $3.30$ & $0.00$ \\
Claude-sonnet-4.5 & $2.65$ & $0.25$ & $3.42$ & $+0.12$ \\
Qwen3-235B        & $3.10$ & $0.10$ & $3.20$ & $-0.10$ \\
Qwen3-80B         & $2.65$ & $0.05$ & $2.70$ & $-0.60$ \\
\bottomrule
\end{tabular}
\caption{March~2026 CPI year-over-year nowcasts around the April~8 information event. ``Median'' and ``IQR'' are the median and interquartile range of the hourly nowcasts over April~2 to~7, 2026. ``Final'' is the last nowcast before the April~10 release and ``Deviation'' is its difference from the $3.3\%$ release, in percentage points.}
\label{tab:err-event}
\end{table}

\section{Case Study: Information Shocks and Nowcast Revisions}\label{app:case-study-details}

This appendix expands the case study summarized in Section~\ref{sec:case-study}.

\paragraph{Setup.}
High-frequency event windows are a standard tool in empirical macroeconomics for isolating the response of beliefs and asset prices to scheduled news shocks such as FOMC announcements and macro releases \citep{kuttner2001monetary,nakamura2018high,bauer2023alternative}. We use this framework to analyze the 10:30 to 14:00 ET window on April~8, 2026, during which the joint release of the EIA Petroleum Status Report and the FOMC meeting minutes coincides with the trend break in GPT-5's March-inflation nowcast cluster (Figure~\ref{fig:case-cpi-pce} in the main text).

\paragraph{Gasoline inventory release.}
The EIA Petroleum Status Report for the week ending April~3, 2026 \citep{eia2026wpsr} reported a 1.589~Mbbl decline in U.S.\ motor gasoline inventories against a Bloomberg consensus decline of 1.400~Mbbl, a 3.144~Mbbl decline in distillate-fuel inventories against a consensus decline of 1.500~Mbbl, and a 0.214~Mbbl/d week-over-week contraction in domestic motor-gasoline production against a consensus contraction of 0.152~Mbbl/d. In each of the three series the realized value fell below consensus, so the market had underestimated the prevailing tightness in U.S.\ refined-product inventories and supply. Let $I_t$ denote end-of-week U.S.\ gasoline inventories, $Q_t$ domestic refinery output, $M_t$ imports, $X_t$ exports, and $C_t$ domestic consumption. The accounting identity
\begin{equation}
\Delta I_t \;=\; Q_t + M_t - X_t - C_t
\label{eq:invid}
\end{equation}
implies that the inventory surprise reflects underlying surprises in production, net trade, and consumption. In this case, an unexpected stock decline alongside a production contraction characterizes a tighter-than-expected refined-product market. Inventory data are central to disentangling demand- from supply-driven movements in oil and refined-product prices \citep{kilian2009not,kilian2014quantitative}, and they act as a lagged indicator of the realized supply--demand gap in commodity markets \citep{pindyck1994inventories}. The release therefore provides new evidence that the late-March refined-product market was tighter than the agent had previously predicted. In the competitive storage framework of \citet{williams1991storage} and \citet{deaton1992behaviour}, lower inventories raise the shadow value of storage and the equilibrium spot price, so the appropriate sign of the inflation-nowcast revision is upward.

\paragraph{FOMC outlook signal.}
At 14:00 ET the Federal Reserve released the minutes of the March~17--18 FOMC meeting \citep{fomc2026marchminutes}. The minutes report that energy-driven price pressures had stalled disinflation, that the staff revised its near-term inflation outlook upward, and that inflation risks were viewed as skewed to the upside. Empirically, FOMC minutes move market expectations even when the policy rate is unchanged, by disclosing the Committee's assessment of incoming data and risks beyond what the post-meeting statement conveys, or known as the ``central-bank-information component'' of FOMC \citep{rosa2013financial,hansen2018transparency,jarocinski2020deconstructing}. The explicit upside-inflation language operates through the latter. This reinforces and emphasizes the signal delivered by the EIA Petroleum Status Report, so both April~8 releases push the March nowcast in the same direction.

\paragraph{Component-contribution calculation.}
The magnitude of the revision is also consistent with a simple component-contribution calculation. Let \(R^H_t\) denote the event-window revision in the agent's headline inflation nowcast for index \(H\). If the inventory release primarily changes the agent's assessment of the gasoline component, then, to a first order,
\begin{equation}
R^H_t
\equiv
\widehat{\pi}^{H,\mathrm{post}}_t
-
\widehat{\pi}^{H,\mathrm{pre}}_t
\approx
s^H_{g,t}
\left(
\widehat{\pi}^{g,\mathrm{post}}_t
-
\widehat{\pi}^{g,\mathrm{pre}}_t
\right),
\end{equation}
where \(s^H_{g,t}\) is the gasoline expenditure share. For CPI, the agent's March year-over-year nowcast increased by about \(0.3\) percentage point. The BLS relative importance of gasoline in CPI was approximately \(2.9\%\). The implied revision to the agent's latent March gasoline inflation forecast is therefore $\frac{0.3}{0.0289} \approx 10.4$ percentage points. The realized gasoline price increase was \(18.9\%\) year-over-year. The backed-out gasoline component adjustment of \(10.4\) percentage points is therefore of the same order of magnitude as the realized gasoline inflation rate.

\paragraph{Discussion.}
The inventory surprise supplies a theoretically signed, backward-looking demand signal that directly informs the month being nowcast. The FOMC minutes supply a complementary signal about the Committee's view of upside inflation risks. By pairing high-frequency forecasts with time-stamped public releases, LiveMacroEval supports interpretable, hypothesis-driven attribution of belief revisions to specific economic events, in the spirit of high-frequency event studies in economics and finance.

\section{Tool and Agent Design: Full Results}\label{app:agent-design-details}

This appendix expands Section~\ref{sec:tool-agent-design} with the full scoring tables, the paired tests, the behavioral evidence behind the two designs, and the caveats that bound the comparison.

\subsection{Design and scoring}

The two agent designs and the control share one base model, Claude-sonnet-4.5, and one live protocol. The control issues a plain single prompt. The tool-augmented agent is the same model equipped with the Anthropic Financial Services plug-in, for which we did not supply the paid data-subscription credentials that some of its data tools require, so our result bounds the free-tier configuration only. The multi-agent team replaces the single agent with an orchestrator that delegates to a researcher sub-agent, which gathers a first-pass value and sources for every indicator, and to an independent verifier sub-agent, which re-derives and cross-checks each number before the orchestrator reconciles the two into the final nowcast.

Scoring follows Appendix~\ref{app:capture-score-details}. The releases scored here run from 9~July to 7~August~2026, the span over which all three configurations were live at once, and the consensus is the Bloomberg consensus as published on the Investing.com economic calendar.

We report two scores. The LiveMacro Score of Section~\ref{sec:metric} takes the realized announcement-window return $r_g$ as its target. The consensus-relative score keeps the form of Appendix~\ref{app:capture-score-details} and replaces that target with the realized stock-equivalent shock $Q_g = \sum_{i \in \mathcal{I}} \beta_i S_{i,g}$, so it asks whether a nowcast anticipates the macro surprise itself rather than the market's response to it. Both give the consensus exactly zero, because the consensus carries $\hat Q_g \equiv 0$. They are reported as the ``LiveMacro'' and ``Cons.-rel.'' columns of Table~\ref{tab:agent-windows} and as the paired rows of Table~\ref{tab:agent-paired}.

\subsection{Scores and paired tests}

Table~\ref{tab:agent-windows} reports every configuration on that identical set of releases. Table~\ref{tab:agent-paired} reports paired event bootstraps. The paired form resamples the same release indices for both arms of a comparison, which is the appropriate test here because both arms face identical releases and their errors are correlated, so the two marginal intervals overlap far more than the difference warrants.

The tool-augmented agent leads the control on both scores, and every paired interval for that comparison contains zero. We therefore report no significant benefit from the tool rather than an advantage, and we regard the comparison as a tie.

The multi-agent team is the only configuration that clears the consensus baseline. Its advantage over both other rungs is significant on the consensus-relative score, with paired intervals excluding zero, and points the same way on the LiveMacro Score without reaching significance. We therefore do not claim a significant improvement in the LiveMacro Score.

\begin{table}[htbp]
\centering
\footnotesize
\setlength{\tabcolsep}{6pt}
\begin{tabular}{@{}lrrr@{}}
\toprule
\textbf{Configuration} & \textbf{Live-} & \textbf{Cons.-} & \textbf{Rel.} \\
 & \textbf{Macro} & \textbf{rel.} & \textbf{err.} \\
\midrule
plain prompt (control)     & $-0.080$ & $-0.404$ & $0.482$ \\
\;\;$+$ financial plug-in  & $-0.055$ & $-0.297$ & $0.365$ \\
\;\;$+$ multi-agent team   & $+0.020$ & $+0.019$ & $0.182$ \\
web-search agent (GPT-5)   & $-0.118$ & $-0.267$ & $0.349$ \\
auto-ARIMA                 & $-0.611$ & $-0.853$ & $0.578$ \\
\bottomrule
\end{tabular}
\caption{Scores on the identical set of releases from 9~July to 7~August~2026. ``Cons.-rel.'' is the consensus-relative score defined above, and ``Rel.\ err.'' is the mean absolute error normalized per field as in Appendix~\ref{app:nowcast-error}, where lower is better.}
\label{tab:agent-windows}
\end{table}

\begin{table}[htbp]
\centering
\footnotesize
\setlength{\tabcolsep}{2pt}
\begin{tabular}{@{}llrr@{}}
\toprule
\textbf{Comparison} & \textbf{Score} & $\boldsymbol{\Delta}$ & \textbf{95\% CI} \\
\midrule
plug-in $-$ control & LiveMacro & $+0.025$ & $[-0.026, +0.098]$ \\
                    & Cons.-rel. & $+0.107$ & $[-0.010, +0.490]$ \\
team $-$ plug-in    & LiveMacro & $+0.075$ & $[-0.109, +0.182]$ \\
                    & Cons.-rel. & $+0.315$ & $\mathbf{[+0.263, +1.213]}$ \\
team $-$ control    & LiveMacro & $+0.100$ & $[-0.093, +0.209]$ \\
                    & Cons.-rel. & $+0.422$ & $\mathbf{[+0.296, +1.548]}$ \\
\bottomrule
\end{tabular}
\caption{Paired event bootstraps, $10{,}000$ resamples of the release indices the two arms share. A positive $\Delta$ favors the first-named configuration, and intervals excluding zero are shown in bold.}
\label{tab:agent-paired}
\end{table}

\subsection{Position size and directional alignment}\label{app:kappa-phi}

Both scores share one form. Write $\hat Q_g$ for the model's predicted stock-equivalent shock of Appendix~\ref{app:capture-score-details} and $Y_g$ for the realized quantity it is scored against, either the return $r_g$ or the shock $Q_g$. Then
\begin{equation}\label{eq:kp-score}
\mathrm{Score} = \frac{\sum_g Y_g^2 - \sum_g (Y_g - \hat Q_g)^2}{\sum_g Y_g^2 + \sum_g (Y_g - \hat Q_g)^2},
\end{equation}
with the sums running over $g \in \mathcal{G}^{\mathrm{live}}$. Define the \emph{position size} $\kappa$, the magnitude of the shock a model predicts relative to the magnitude of what it is predicting, and the \emph{directional alignment} $\phi$, the uncentered correlation between the two,
\begin{equation}
\kappa = \sqrt{\frac{\sum_g \hat Q_g^2}{\sum_g Y_g^2}}, \qquad
\phi = \frac{\sum_g \hat Q_g Y_g}{\sqrt{\sum_g \hat Q_g^2 \sum_g Y_g^2}} .
\end{equation}
Both are dimensionless and $\phi \in [-1, 1]$. They are the two channels a nowcast can score through, how far it departs from the consensus and how well it points in the right direction, and the algebra below makes the trade-off between them exact. Expanding the squared error and dividing by $\sum_g Y_g^2$ gives $\sum_g (Y_g - \hat Q_g)^2 / \sum_g Y_g^2 = 1 + \kappa^2 - 2\kappa\phi$, so Equation~\ref{eq:kp-score} becomes
\begin{equation}\label{eq:kp-identity}
\mathrm{Score} = \frac{2\kappa\phi - \kappa^2}{2 + \kappa^2 - 2\kappa\phi} .
\end{equation}
The denominator is positive wherever the score is defined, so
\begin{equation}\label{eq:kp-condition}
\mathrm{Score} > 0 \iff \kappa < 2\phi .
\end{equation}
A model beats the consensus exactly when its position size is smaller than twice its directional alignment. This is the exact form of the trade-off between size and direction that Section~\ref{sec:tool-agent-design} states in words and that Appendix~\ref{app:livemacro-directional} describes qualitatively. Two cases follow. If $\phi \le 0$, no positive position size satisfies Equation~\ref{eq:kp-condition}, so a model with no directional alignment cannot be rescued by rescaling its predictions, and its only route to the reference value is to predict no surprise at all. If $\phi > 0$, the score is maximized at $\kappa^{\star} = \phi$, where it equals $\phi^2 / (2 - \phi^2)$, so a model is correctly sized when its position size equals its directional alignment and oversized whenever $\kappa > \phi$.

Table~\ref{tab:kappa-phi} reports both quantities for the two agent designs and the control. Only the multi-agent team satisfies Equation~\ref{eq:kp-condition}, and it does so on both terms at once, holding roughly a third of the position size of the other two and the only clearly positive directional alignment in the panel. The two lower rungs are oversized on both scores. Their directional alignment on the LiveMacro Score is small enough that we read it as no reliable alignment rather than as a wrong-way tilt, and on the consensus-relative score it is clearly negative. Neither could have reached the reference value by rescaling.

\begin{table}[htbp]
\centering
\footnotesize
\setlength{\tabcolsep}{4pt}
\begin{tabular}{@{}llrrr@{}}
\toprule
\textbf{Score} & \textbf{Configuration} & $\boldsymbol{\kappa}$ & $\boldsymbol{\phi}$ & \textbf{Score} \\
\midrule
LiveMacro & plain prompt        & $0.34$ & $-0.08$ & $-0.080$ \\
          & $+$ plug-in         & $0.31$ & $-0.03$ & $-0.055$ \\
          & $+$ team            & $0.10$ & $+0.25$ & $+0.020$ \\
Consensus- & plain prompt       & $0.72$ & $-0.59$ & $-0.404$ \\
relative  & $+$ plug-in         & $0.65$ & $-0.32$ & $-0.297$ \\
          & $+$ team            & $0.21$ & $+0.19$ & $+0.019$ \\
\bottomrule
\end{tabular}
\caption{Position size $\kappa$ and directional alignment $\phi$, on the same releases as Table~\ref{tab:agent-windows}. A configuration beats the consensus exactly when $\kappa < 2\phi$ (Equation~\ref{eq:kp-condition}). The rightmost column is recovered through Equation~\ref{eq:kp-identity}.}
\label{tab:kappa-phi}
\end{table}

\subsection{Why the multi-agent team clears the baseline}

Table~\ref{tab:agent-sizing} reports the relationship between how much a configuration stakes on a release and how wrong it is. For the plain prompt, the plug-in, the search agent and the ARIMA baseline, the correlation between predicted surprise magnitude and absolute error is strongly positive, so their largest predictions are also their largest misses. The multi-agent team is the only configuration for which that correlation is negative, and it carries the smallest predicted magnitude in the panel by a factor of three. Its sign accuracy is matched by the web-search agent, which stakes far more on each call and scores well below the baseline, so restraint and direction have to hold together for a configuration to satisfy Equation~\ref{eq:kp-condition}.

\begin{table}[htbp]
\centering
\footnotesize
\setlength{\tabcolsep}{6pt}
\begin{tabular}{@{}lrrr@{}}
\toprule
\textbf{Configuration} & \textbf{mean} $|\hat S|$ & \textbf{corr} & \textbf{sign} \\
\midrule
plain prompt (control)    & $0.97$ & $+0.70$ & $0.38$ \\
\;\;$+$ financial plug-in & $0.80$ & $+0.78$ & $0.31$ \\
\;\;$+$ multi-agent team  & $0.26$ & $-0.26$ & $0.54$ \\
web-search agent (GPT-5)  & $0.81$ & $+0.71$ & $0.54$ \\
auto-ARIMA                & $1.60$ & $+0.97$ & $0.38$ \\
\bottomrule
\end{tabular}
\caption{Sizing discipline, on the same releases as Table~\ref{tab:agent-windows}. ``corr'' is the correlation between a configuration's predicted surprise magnitude and its absolute error, and ``sign'' is the fraction of field-events on which the predicted surprise has the correct sign. Realized surprises average $0.62$.}
\label{tab:agent-sizing}
\end{table}

One further comparison locates the design step that produces them. Adding the plug-in leaves the predicted surprises correlated with those of the plain prompt at $0.60$ and the mean predicted surprise magnitude essentially unchanged. Adding the verifier instead lowers that correlation to $0.33$ and cuts the mean predicted surprise magnitude by a factor of $3.7$. These are descriptive associations on small samples, and they decompose the observed scores rather than establishing that the verifier caused the smaller positions.

Execution logs describe what the verifier does. We extracted statistics from the orchestrator's server logs for runs between August~22 and~25, 2026, which rotate after seven days, so the durable record is the extract released with our code. Most completed runs delegate to both sub-agents, and a completed run issues a mean of $173$ web searches. The verifier accounts for $49\%$ of all retrieval, issuing $1{,}754$ searches against the researcher's $1{,}823$, of which $1{,}676$ are distinct, so verification here is a second independent pass over the same evidence. This is the basis for the conjecture offered in Section~\ref{sec:tool-agent-design}. The orchestrator's contract is to reconcile the two sub-agents' estimates into one line, so where two independent passes disagree the reconciled value falls between them, which is a smaller departure than either estimate alone.

\subsection{Where the financial-analysis plug-in changes the error}

The tie between the tool-augmented agent and the control hides a redistribution of error rather than an absence of any effect. To see it, we split the field-events into two halves at the median market-impact weight $|\beta_i|$, so one half holds the releases that move equity prices most. On that half the plug-in roughly halves relative error, $0.206$ against the control's $0.397$. On the other half it is worse, $0.635$ against $0.525$. The tool helps where its coverage is deepest and adds noise elsewhere, and the two effects leave the aggregate score unchanged. The tails point the same way. The plug-in makes fewer moderate misses than the control, and owns the single worst miss at $4.94$ against the control's largest at $2.83$. It also stakes most where the control most clearly beats it, predicting a mean surprise of $1.35$ on the three inflation series against $0.49$ elsewhere, and calling the direction on those series correctly on $43\%$ of field-events against the control's $71\%$.

\subsection{Caveats}

All three configurations were live together for about one month, which bounds how much evidence this comparison can carry, so we report paired intervals and not just a bare point estimate. Each design was observed in a single live configuration over that period rather than in a controlled multi-seed experiment. The multi-agent team costs roughly fifteen minutes per run against a single prompt for the control, so its advantage, where present, has costs. Finally, the plug-in was run without the paid data subscriptions some of its tools require, so the tool result bounds the freely available configuration and does not speak to a fully credentialed one.

\section{Artifact Licenses and Terms of Use}\label{app:artifact-licenses}

This appendix documents the licenses and terms under which each external artifact is used in LiveMacroEval. We do not redistribute any of these artifacts. The supplementary release contains only our own code and the nowcast outputs we generated through provider APIs.

\paragraph{Federal Reserve regional-bank nowcasts.}
The Atlanta Fed GDPNow \citep{atlfed_gdpnow}, the New York Fed Staff Nowcast \citep{nyfed_nowcast}, the St.~Louis Fed Real GDP Nowcast \citep{stlfed_realgdp_nowcast}, the Cleveland Fed Inflation Nowcasting \citep{clevelandfed_inflation_nowcast}, and the Chicago Fed CHURN \citep{chicagofed_churn} are publicly released by their respective Federal Reserve regional banks as works of the U.S.\ federal government and are in the public domain. We retrieve each series from the issuing bank's public release pages at the published vintage timestamps and use them solely as comparison nowcasts within this benchmark.

\paragraph{FRED-MD and FRED-QD.}
The FRED-MD and FRED-QD vintage databases \citep{mccracken2016fredmd,mccracken2021fredqd} are maintained by the Federal Reserve Bank of St.~Louis and distributed publicly for research use. We download the monthly and quarterly vintages from the official FRED-MD/FRED-QD pages and use them for the canonical transformation codes and historical training samples described in Appendix~\ref{app:arima-baseline}.

\paragraph{Bloomberg ECOS and Bloomberg Economic Calendar.}
The Bloomberg Economist Estimates (ECOS) panel \citep{bloomberg2024ecos} and the Bloomberg Economic Calendar \citep{bloomberg_econ_cal} are accessed through the authors' institutional Bloomberg Terminal subscription. Usage is restricted to academic research consistent with Bloomberg's Terminal terms of service, and we report only derived quantities (the cross-sectional median consensus and release timestamps) in the paper. No raw Bloomberg data are redistributed.

\paragraph{E-mini S\&P~500 futures intraday data.}
Intraday quotes on front-month E-mini S\&P~500 futures (ticker ES, CME Globex), used to compute the announcement-window equity returns $r_g$ in the LiveMacro Score (Appendix~\ref{app:capture-score-details}), are purchased from FirstRate Data under their standard end-user research license. The license permits internal research use and the reporting of derived quantities. We report only the resulting announcement-window log returns and the estimated $\hat\beta_i$ coefficients in Table~\ref{tab:hist-beta}; the raw intraday quote data are not redistributed.

\paragraph{LLM agents.}
GPT-5 \citep{openai2025gpt5}, Claude-sonnet-4.5 \citep{anthropic2025sonnet45}, Qwen3-235B, and Qwen3-80B \citep{yang2025qwen3} are queried through their providers' public APIs under the respective providers' terms of service for research use. We report only the agents' nowcast outputs; no model weights are redistributed.

\paragraph{Polymarket prediction-market prices.}
Polymarket bucket prices used in the LiveBetting Score (Appendix~\ref{app:polymarket-return-details}) are retrieved from the public Polymarket API consistent with Polymarket's published API terms. We report only derived per-bucket prices at the hourly timestamps used for the simulated betting return.

\paragraph{Intended use of LiveMacroEval.}
LiveMacroEval is released as a non-commercial academic research benchmark for evaluating LLM agents on real-time U.S.\ macroeconomic nowcasting. The released materials consist of (i) the LLM-agent nowcast outputs we generated through provider APIs, (ii) the LiveMacro and LiveBetting scoring code, (iii) the auto-ARIMA baseline code, and (iv) per-indicator metadata such as titles, units, official-source labels, and scale guardrails. No raw Bloomberg, FirstRate Data, or Polymarket data are included, and the released LLM nowcasts are derivative outputs of the underlying provider APIs whose terms permit reporting derived research results. This intended use is consistent with the access conditions of every source artifact documented above: each upstream artifact is used either under a public-domain or research-use license (Fed nowcasts, FRED-MD/QD), or under a license that permits internal research use and the reporting of derived quantities (Bloomberg ECOS, FirstRate Data ES futures), or under the providers' standard API terms (LLM agents, Polymarket).

\end{document}